\documentclass[11pt]{article}

\usepackage[T1]{fontenc}
\usepackage[margin=1in]{geometry}
\usepackage{newtxtext,newtxmath}
\usepackage{microtype}
\usepackage[round,authoryear]{natbib}
\usepackage{graphicx}
\usepackage{booktabs}
\usepackage{array}
\usepackage{tabularx}
\usepackage{caption}   
\usepackage{subcaption}
\usepackage[section]{placeins} 
\usepackage{needspace}
\usepackage{etoolbox}
\usepackage{xcolor}
\usepackage{hyperref}  
\usepackage{xurl}      
\hypersetup{colorlinks=true, linkcolor=blue, urlcolor=blue, citecolor=blue}
\newcolumntype{Y}{>{\raggedright\arraybackslash}X}  
\newcolumntype{Z}{>{\centering\arraybackslash}X}    

\newlength{\plotwidth}
\newlength{\plotheight}
\newenvironment{plotblock}
  {\begin{figure}[htbp]\centering}
  {\end{figure}}

\newcommand{\plotpanel}[3]{%
  \begin{subfigure}[t]{0.48\linewidth}
    \centering
    \includegraphics[width=\linewidth,height=0.31\textheight,keepaspectratio]{#1}
    \caption{#2}
    \label{#3}
  \end{subfigure}%
}

\pretocmd{\subsection}{\FloatBarrier\Needspace{6\baselineskip}}{}{}
\pretocmd{\subsubsection}{\FloatBarrier\Needspace{6\baselineskip}}{}{}

\title{RAG Collapse: LLM Responses Collapse When Retrieved Documents Are Self-Authored}

\author{%
  Gregory Druck\\
  Graphite Growth\\
  \texttt{greg@graphitehq.com}%
  \and
  Ethan Smith\\
  Graphite Growth\\
  \texttt{ethan@graphitehq.com}%
}

\begin{document}
\maketitle

\begin{abstract}
LLM responses are based on the internet (via training or RAG), and AI is now
used to generate a significant amount of content online~\citep{paredes2026aiwrites},
creating the potential for a self-reinforcing feedback loop. Prior work has shown that
when LLMs are recursively trained on their own output, they experience model
collapse~\citep{shumailov2024collapse}: responses become less diverse, and eventually no longer resemble the
original training data. In this paper, we show that a similar collapse occurs
if LLM-based AI systems retrieve references they authored using a search tool. We
call this RAG collapse. We
conduct extensive experiments with three types of simulations of AI systems retrieving references
they generated, using three model families, and 1,019 information-seeking prompts, totaling
1,528 simulations and over one million LLM API calls, and find that 79.6\% (1,216/1,528) of
simulations end in collapse. Surprisingly, even a single
self-authored reference can trigger collapse because the LLM
disproportionately cites its own content. This self-bias persists even after
controlling for reference quality. 
\end{abstract}

\setlength{\parindent}{1.25em}
\setlength{\parskip}{0pt}

\section{Introduction}
AI systems like ChatGPT are powered by large language models (LLMs)~\citep{brown2020fewshot}. LLMs are trained on large text datasets, primarily from the internet. AI systems also often have access to a search tool to retrieve additional information from the internet. Using search results to generate responses is called retrieval-augmented generation (RAG)~\citep{lewis2020rag}. LLM responses are randomly sampled from a probability distribution based on the internet. Therefore, responses from an LLM should approximately mirror the diversity of thought online.

AI is now also used to generate large amounts of text online. In our recent AI-generated content paper~\citep{paredes2026aiwrites}, we show that there are now as many AI-generated articles as human-written articles being published. In this paper, we find that a substantial share of references LLMs cite are AI-generated. This creates the potential for a self-reinforcing feedback loop.

\citet{shumailov2024collapse} showed that training LLMs on their own generations can lead to model collapse, in which token probabilities become increasingly concentrated, resulting in substantially less diversity in responses and eventually text that no longer resembles the original training data.

We investigate what happens when search-enabled LLMs retrieve pages they authored. We call a reference self-authored when it was generated by the same model using its own answer to a prompt. Across 1,528 simulations, we find that retrieving self-authored references in RAG leads to collapse (essentially the same response every time) in 79.6\% (1,216/1,528 simulations). In one experiment (see Section~\ref{sec:large-scale-experiments}), 22\% of pairs of LLM responses are paraphrases initially. By the end of the simulation, 89\% are.

Surprisingly, this collapse begins almost immediately, even when only a single reference is self-authored by the model. The results suggest that self-authored references have a disproportionate influence on responses. We find that they are cited more often than original references, including those that are themselves AI-generated, even after we control for reference quality. This suggests self-bias~\citep{panickssery2024own}: the LLM prefers its own answer in its own writing style.

Our simulations demonstrate a risk: AI responses no longer reflect the diversity of human thought, and as more AI-generated content is published, the internet converges on a single perspective.

\section{Example}
Consider the information-seeking prompt ``Who are the best Twitch streamers currently?'' This may trigger a search that identifies the following pages as relevant:

\begin{itemize}
  \item \url{https://streamscharts.com/news/top-twitch-streamers-2025}
  \item \url{https://en.wikipedia.org/wiki/List_of_most-followed_Twitch_channels}
  \item \url{https://www.shanethegamer.com/esports-news/most-watched-twitch-streamers-2025/}
  \item \url{https://www.techradar.com/ai-platforms-assistants/twitchs-most-subscribed-streamer-is-not-human}
  \item \url{https://www.dexerto.com/entertainment/top-20-most-followed-twitch-streamers-750744/}
  \item \url{https://streamscharts.com/news/most-subscribed-twitch-streamers-2025}
\end{itemize}

If we ask GPT-5.2 ten times, using the above as references (the sources it uses to answer), we get ten different answers. Figure~\ref{fig:image5} provides two examples that differ in structure and streamers mentioned.

\begin{plotblock}
\includegraphics[width=\plotwidth,height=\plotheight,keepaspectratio]{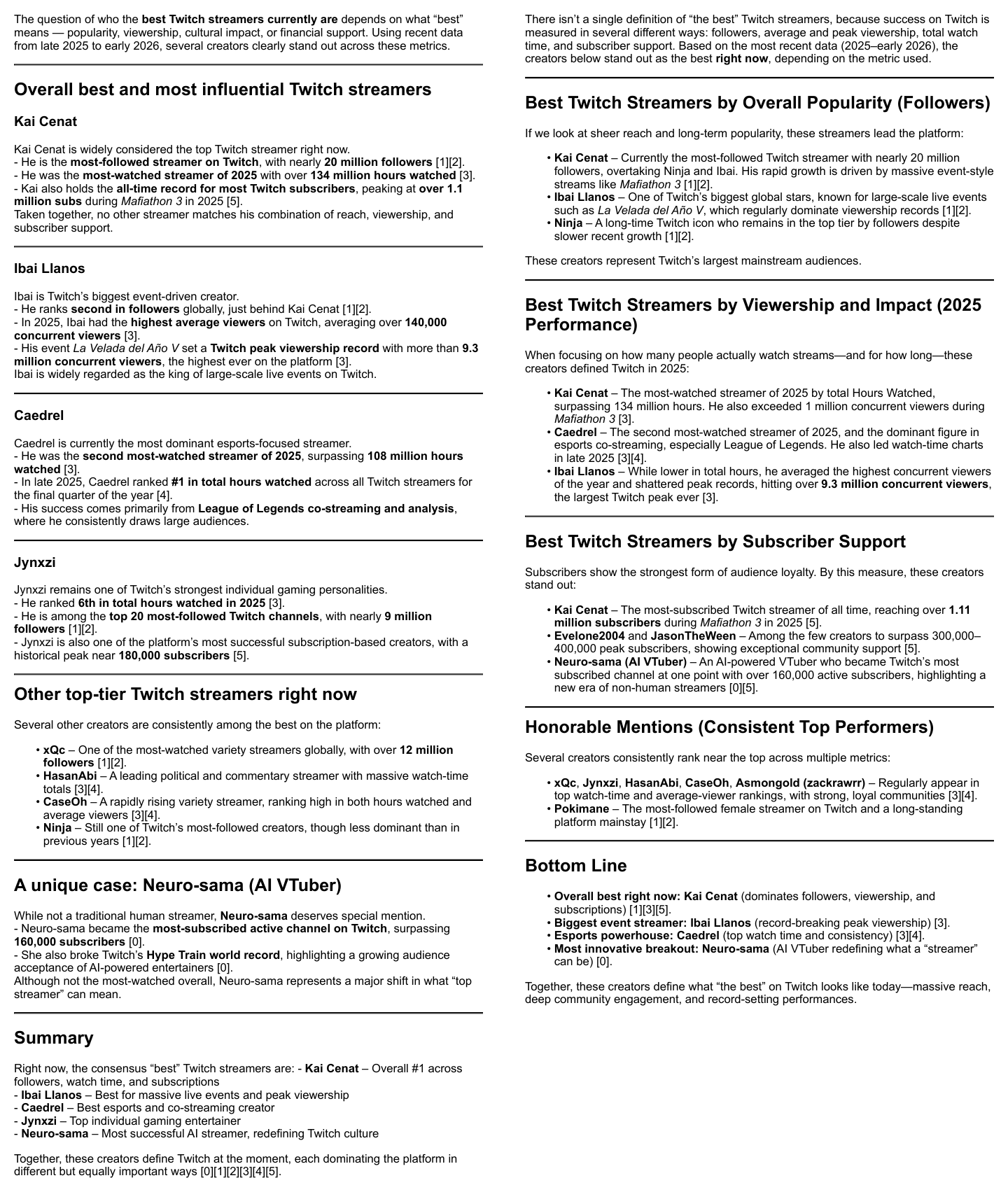}
\caption{Two responses to the prompt ``Who are the best Twitch streamers currently?'' from GPT-5.2 that differ in structure and streamers mentioned due to random sampling.}
\label{fig:image5}
\end{plotblock}

The visibility of an entity in a set of responses is the percentage of responses that mention the entity. Figure~\ref{fig:image12} displays visibilities for each streamer using ten responses. Note the diversity in responses. Some streamers, like ``Kai Cenat'', are mentioned in every response, while others, like ``Squeezie'', are mentioned occasionally (two of ten responses).

\begin{plotblock}
\includegraphics[width=\plotwidth,height=\plotheight,keepaspectratio]{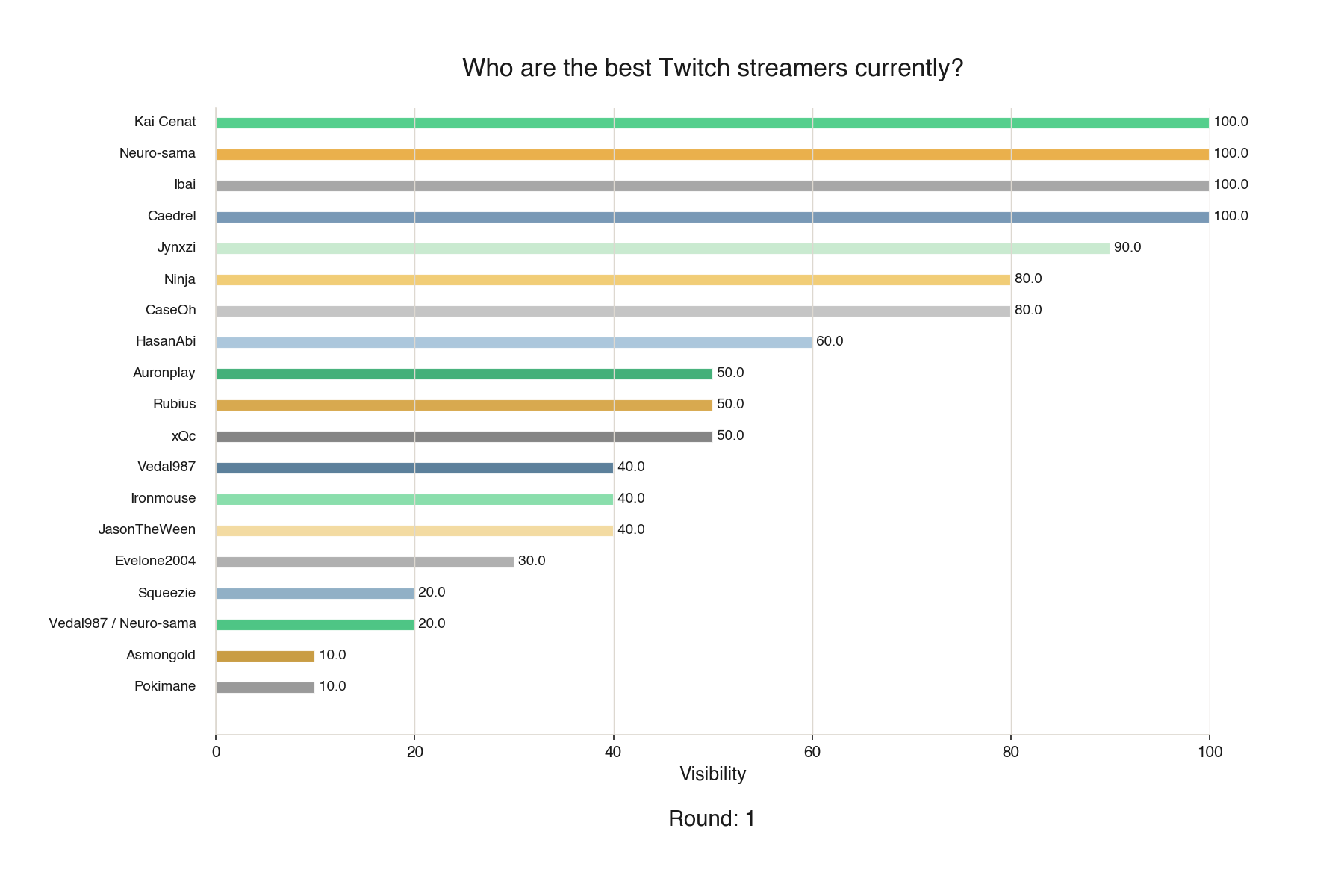}
\caption{Entity visibility for ``Who are the best Twitch streamers currently?'' at round 1. ``Kai Cenat'' appears in all ten, while ``Squeezie'' appears in only two of ten.}
\label{fig:image12}
\end{plotblock}

What if we use AI to generate a new article based on one of the model's responses, replace one of the original references with it, and then regenerate the answers? What if we repeat this process over and over? We answer this question by running our Replace One simulation. The simulation takes place over a number of rounds, and in each round, we replace one original reference with a self-authored reference. We provide a complete description in Section~\ref{sec:simulations}.

After five rounds of recursively replacing references, we end up with the streamer visibility plot in Figure~\ref{fig:image57}. Every streamer now has either 0\% or 100\% visibility. This means that every answer mentions exactly the same streamers. The previously wide distribution over streamers considered the best has collapsed. While the streamers mentioned in every response in the original distribution have survived, some that were initially mentioned frequently, like ``Ninja'', have dropped out of the responses entirely, and ``xQc'', originally mentioned 50\% of the time, now always appears. So in addition to less diversity, the answers to this prompt have changed.

\begin{plotblock}
\includegraphics[width=\plotwidth,height=\plotheight,keepaspectratio]{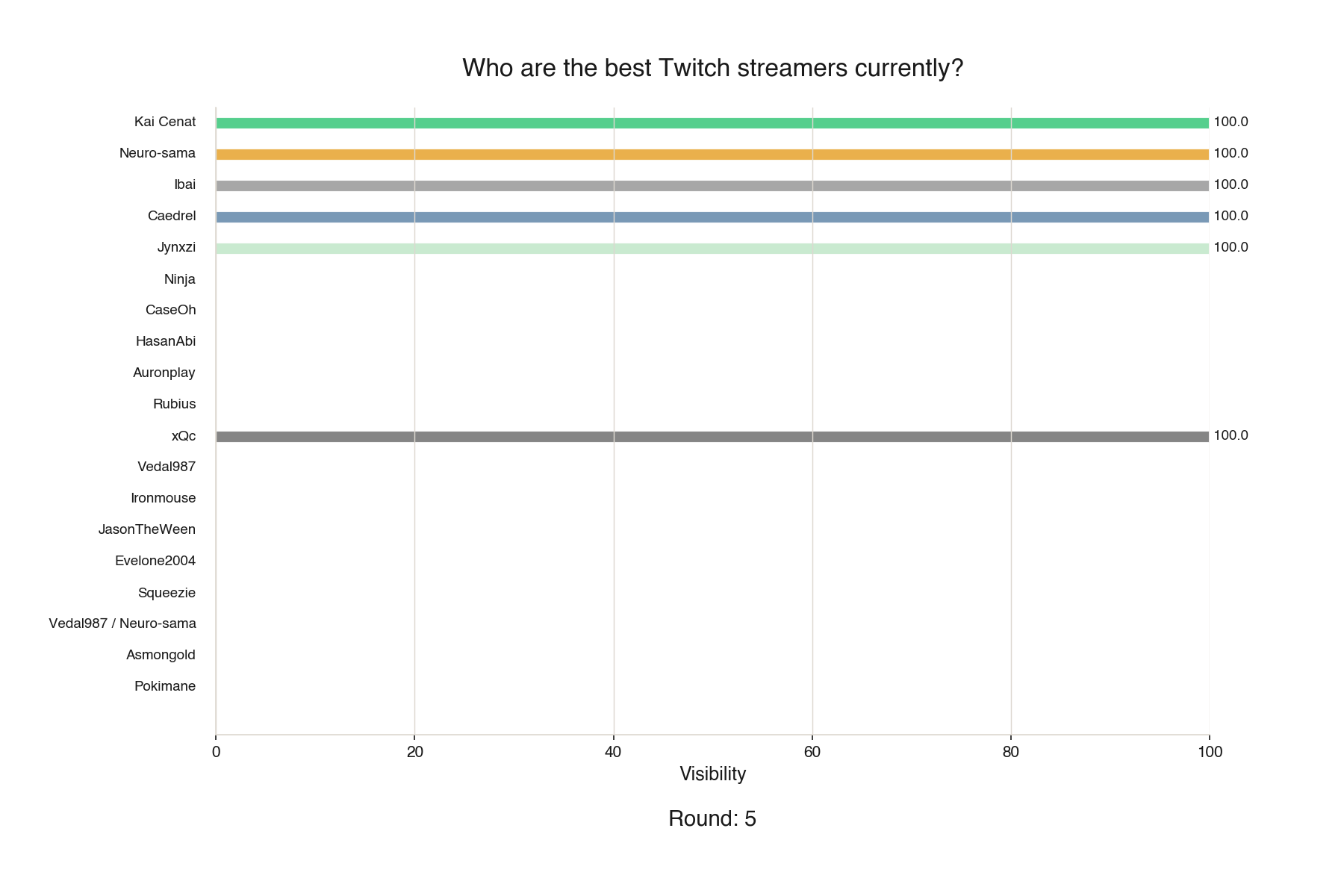}
\caption{Entity visibility for ``Who are the best Twitch streamers currently?'' at round 5. Every streamer now has either 0\% or 100\% visibility. The distribution has collapsed.}
\label{fig:image57}
\end{plotblock}

The responses remain collapsed through the rest of the simulation.

\section{Background and Related Work}
\subsection{LLMs and RAG}
LLMs~\citep{brown2020fewshot} are pre-trained on massive text datasets from the internet. An LLM can respond to a prompt (without web search) based on its parametric knowledge, but cannot attribute its answer to a particular document in its training set.

To generate a response, LLMs predict a probability distribution over possible next tokens (sub-word character sequences), and then randomly select a next token according to that distribution. As a result, LLM responses are non-deterministic, and they often generate different responses to the same prompt. We describe this in detail in prior work~\citep{druck2026randomness}.

To reduce hallucinations and incorporate more real-time information, AI systems like ChatGPT now have access to web search. In this paper, we use retrieval-augmented generation (RAG)~\citep{lewis2020rag} as a general term that encompasses any use of web search, including agentic search~\citep{singh2025agentic}, in which the model can selectively choose when to search and which results to read.

When the model uses the web search tool, it adds the retrieved information to its context window to better address the prompt. \citet{farahani2024memory} show that LLMs are heavily influenced by this context. RAG also allows the LLM to provide citations to the documents it has read to inform the answer.

For information-seeking prompts (i.e., questions), when using RAG, the LLM's answer distribution should resemble the distribution of answers on the internet. At a high level, the model should faithfully represent the diversity of opinion and thought on the internet. (Though note that post-training alignment with human preferences to remove certain types of unwanted bias may shift this distribution.)

\subsection{Model Collapse}
What happens now that AI is used to generate text published on the internet? What if an AI system trains on or retrieves its own generations?

\citet{shumailov2024collapse} showed that training LLMs on their own generations can lead to model collapse, in which token probabilities become increasingly concentrated, resulting in substantially less diversity in responses and eventually text that does not resemble the original training data.

Why does this happen? Recursively resampling and re-estimating the probability distribution causes it to drift over time. Specifically,

\begin{itemize}
  \item A finite sample from a probability distribution will not faithfully represent that distribution.
  \item The model may not faithfully represent the training data distribution, either because it lacks the capacity to fit the data or because learning algorithms fail to find optimal weights.
\end{itemize}

As a simple illustration, suppose we roll an unweighted die. The probability of rolling each number from 1 to 6 is equal ($1/6$), and over many rolls, we expect to get about the same number of 2s and 5s, for example. However, suppose we roll the die ten times and never roll a 2. If we were to take those ten rolls and use them to reweight the die, we would end up with a very different distribution, as the probability of rolling a 2 would become 0. In this way, recursively re-estimating a probability distribution from a sample causes drift.

\subsection{RAG Collapse}
\label{sec:rag-collapse}
What if, rather than being retrained on its generations, a model retrieves its own generations using a search tool?

Suppose we ask AI, ``Who is the best NBA player of all time?'' repeatedly, and get different answers. The response will likely always include Michael Jordan and LeBron James, but may occasionally include Hakeem Olajuwon or Larry Bird. If a few people generated articles using the model's response, and those articles happened to include Hakeem Olajuwon more frequently than Larry Bird, and the AI system retrieved those articles, Larry Bird could drop out of the answer entirely, while Hakeem Olajuwon could start to appear more frequently alongside Jordan and James. These self-authored responses could further influence future articles, and over time, the answers could change and converge onto a particular answer.

Because LLMs are highly influenced by retrieved documents in RAG~\citep{wu2024clasheval}, we may expect the LLM's answer distribution to mirror that of the retrieved documents, leading to a similar collapse in the RAG setting as in the retraining setting. The collapse could actually be more dramatic in RAG because responses are based on a relatively small number of documents compared to the size of the model's training set (the internet).

In our experiments, we find that even small changes to the reference distribution, replacing only one of the original references with a self-authored reference, can cause the responses to collapse. This suggests that RAG is not faithfully reproducing the original input distribution. Instead, there is a bias toward self-authored references. We explore this in Section~\ref{sec:self-authored-influence}.

\citet{wang2025webdynamics} simulate a network of LLMs communicating via a shared RAG database and show that different models converge toward each other over time, and provide a theoretical analysis (using Gaussian Mixture Models). Our work focuses on the collapse of response diversity within a single model's answers across a wide range of prompts, simulations, and frontier models.

\section{Simulations}
\label{sec:simulations}
We now describe our simulations of AI systems retrieving content that they previously generated.

We focus on RAG and do not retrain the base model. We limit the scope to information-seeking prompts, in particular, question prompts.

In each round, we use an LLM to generate several responses to a question based on references. We use the following system instructions:

\begin{center}
\small
\begin{tabularx}{\linewidth}{|Y|}
\hline
You will be given a question, labeled ``question'', and context from the internet, labeled ``context''. \newline Use the context to answer the question thoroughly. \newline Cite each context that informs your answer. Each context has a number. To cite ``context 3'', use [3]. \\ \hline
\end{tabularx}
\end{center}

and the following prompt:

\begin{center}
\small
\begin{tabularx}{\linewidth}{|Y|}
\hline
context: \newline CONTEXT \newline question: \newline QUESTION \\ \hline
\end{tabularx}
\end{center}

where CONTEXT has the form:

\begin{center}
\small
\begin{tabularx}{\linewidth}{|Y|}
\hline
context 0 \newline TITLE \newline TEXT \newline context 1 \newline TITLE \newline TEXT \newline \ldots{} \\ \hline
\end{tabularx}
\end{center}

Initially, CONTEXT contains text from the original references (obtained from commercial AI systems).

We then take some of these responses (ten for Replace All, one for Replace One and Search), remove citations (e.g., [3]), and convert them into the form of online articles using the following system instructions:

\begin{center}
\small
\begin{tabularx}{\linewidth}{|Y|}
\hline
You will be given an answer, labeled ``answer'', to a question, labeled ``question'', and your job is to expand the answer into an online article. \newline Do not change the answer; just expand it into an online article. \newline Return only the article in plain text (not markdown). \\ \hline
\end{tabularx}
\end{center}

and prompt

\begin{center}
\small
\begin{tabularx}{\linewidth}{|Y|}
\hline
question: \newline QUESTION \newline answer: \newline RESPONSE \\ \hline
\end{tabularx}
\end{center}

We do this because LLM responses to a question may look quite different from the original references (typically articles), and we want to simulate someone publishing an article online based on the model's answer.  We then replace the original references in the context with self-authored ones, or add them to a pool of available references, simulating the publication of self-authored articles and their subsequent retrieval by the model during RAG. There are three variants of our simulation that differ in the way they use these self-authored references.

\subsection{Replace All Simulation}
In the \textit{Replace All} simulation, we replace each reference with a self-authored reference based on a response from the previous round. We visualize this simulation with Figure~\ref{fig:image46}. Note, for clarity, in Figures~\ref{fig:image46}--\ref{fig:image14} we show three references and three responses, but our actual simulations use ten responses and at least five references.

\begin{plotblock}
\includegraphics[width=\plotwidth,height=\plotheight,keepaspectratio]{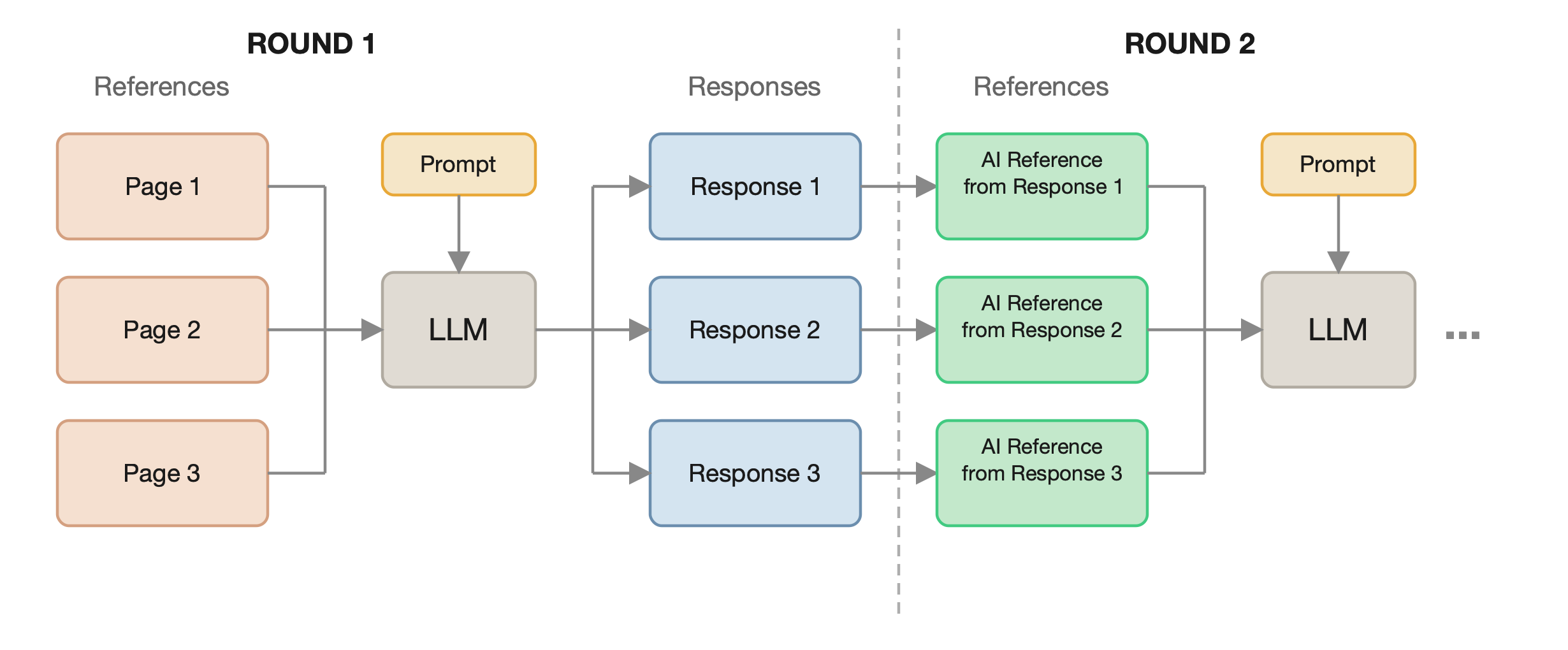}
\caption{Overview of Replace All. In each round, every reference is replaced by a self-authored reference generated from a response in the previous round.}
\label{fig:image46}
\end{plotblock}

This setup is most similar to recursive retraining experiments on model collapse~\citep{shumailov2024collapse}.

However, this simulation makes several assumptions. First, it assumes that the self-authored references are retrieved and included in the context. Second, it assumes all self-authored references are generated from the same set of responses and that the retriever only retrieves that set. In practice, we expect the replacement of the original references to happen more slowly and for the search tool to retrieve self-authored references from different rounds.

\subsection{Replace One Simulation}
In the Replace One simulation, we replace one of the original references with a randomly selected self-authored reference in each round. We depict this simulation in Figure~\ref{fig:image16}.

\begin{plotblock}
\includegraphics[width=\plotwidth,height=\plotheight,keepaspectratio]{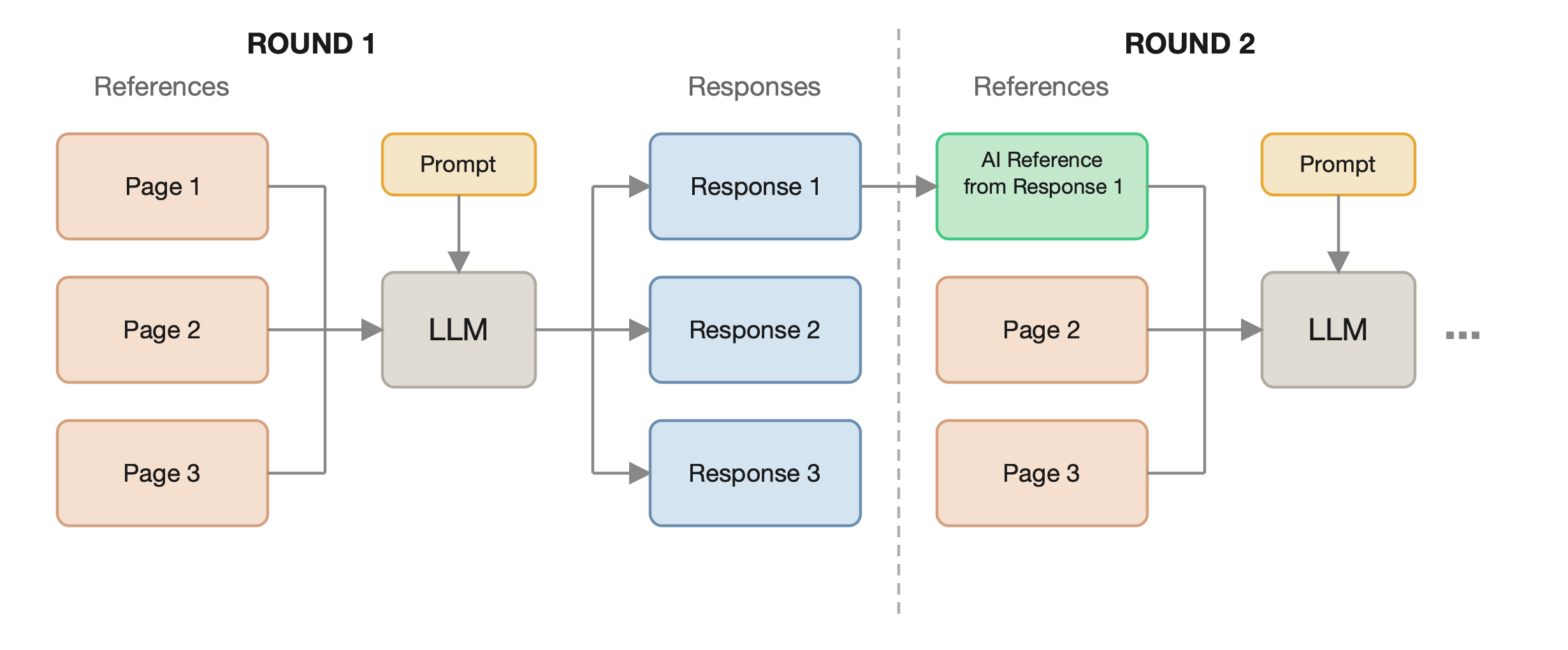}
\caption{Overview of Replace One. Original references are replaced more slowly than in Replace All, although the design still assumes that self-authored references are retrieved.}
\label{fig:image16}
\end{plotblock}

This simulation still replaces the original references, just more slowly, and still assumes that the self-authored references would be retrieved in search.

\subsection{Search Simulation}
In the Search simulation, we maintain a pool of references that includes the original references and add self-authored references to it one at a time. To determine which references to add to the context in each round, we search the reference pool and retrieve the top-$k$ most relevant chunks. We illustrate this simulation in Figure~\ref{fig:image14}:

\begin{plotblock}
\includegraphics[width=\plotwidth,height=\plotheight,keepaspectratio]{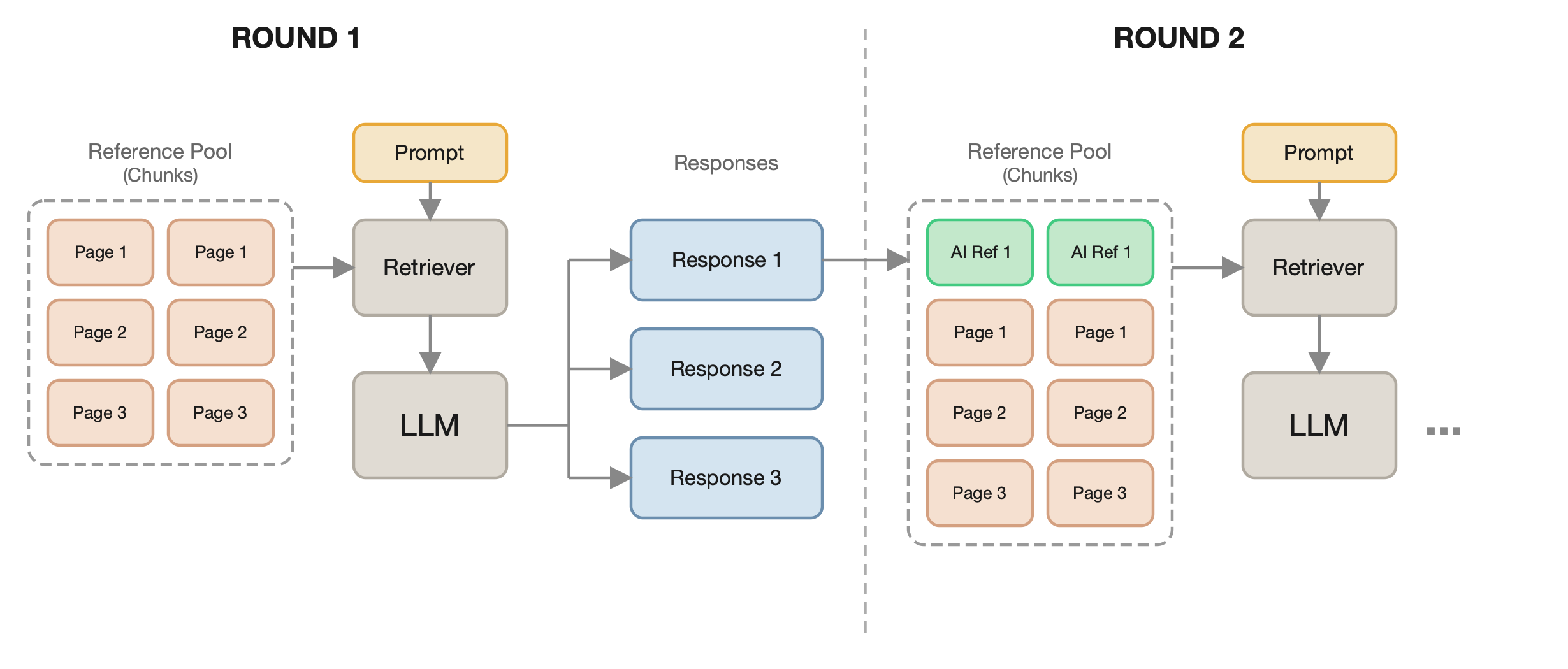}
\caption{Overview of the Search simulation. Self-authored references are added to a reference pool and retrieved using search, competing with the original references for retrieval.}
\label{fig:image14}
\end{plotblock}

In practice, other content would continue to be published alongside AI-generated content. We do not attempt to simulate this in this paper, as it would introduce many additional free parameters. However, the ease and low cost of creating AI-generated content make it highly likely that AI-generated content will comprise an increasing share of online content.

\section{Simulation Implementation Details}
\label{sec:implementation-details}
We use LLM APIs to run the simulation, so that we control the context provided to the model. In particular, we use the \href{https://openai.com/api/}{OpenAI}, \href{https://ai.google.dev/gemini-api/docs}{Gemini}, and \href{https://platform.claude.com/}{Anthropic} APIs.

Due to randomness, there are many possible outcomes from running the simulation on each question. For example, Figure~\ref{fig:image17} shows the entity visibilities in the final round for nine different simulations of ``Who are the best Twitch streamers currently?'' Different entities survive each time.

\begin{plotblock}
\includegraphics[width=\linewidth,height=0.82\textheight,keepaspectratio]{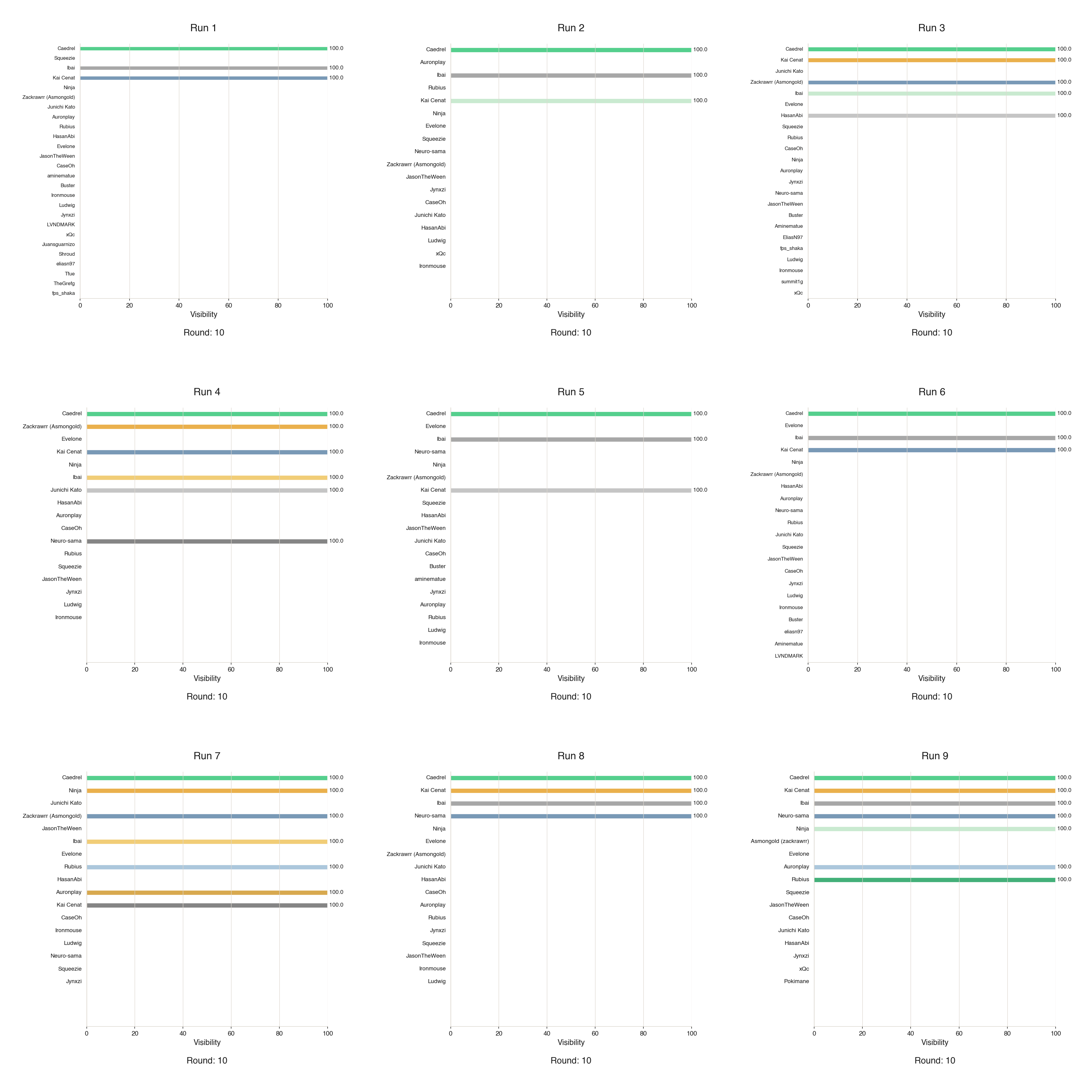}
\caption{Final-round entity visibility across nine independent runs for ``Who are the best Twitch streamers currently?'' Different entities survive in each run, illustrating run-to-run variance.}
\label{fig:image17}
\end{plotblock}

Due to costs (see Section~\ref{sec:costs}), we cannot run many questions many times each. We could run a small number of questions many times, but instead we opt to run many different questions once.

For each question, we initialize references using actual references for these prompts from commercial AI systems. For OpenAI models, we collect references cited by ChatGPT using logged-in accounts. For Gemini and Claude models, we collect references cited in Google AI Overviews. We call these original references.

We scrape the content of the original references and identify the page's main content (removing boilerplate, etc.) using \href{https://github.com/adbar/trafilatura}{trafilatura}. We spot-checked 50 pages. All 50 were scraped successfully, and we judged the main content extraction to be accurate. For the Replace All and Replace One simulations, we also identify the portions of the main content (for both original and self-authored references) relevant to the question using an LLM to reduce costs. For the Search simulation, we use chunks from the full main content.

We initialize the context with the content from the original references. We do not provide the original URLs in the prompt to avoid bias. Instead, we rename them as ``context $n$''. We also randomly shuffle the excerpts in each round to reduce position bias.

In each round of the simulation, we generate ten responses to the question.

To control costs, we stop the Replace All simulation after 10 rounds, the Replace One simulation after 20 rounds, and the Search simulation after 30 rounds. Although Replace One and Search run for more rounds, Replace All replaces all original references after round 1, whereas Replace One replaces all original references by round 10. The original references may never be fully replaced in the context in the Search simulation, depending on whether they continue to be retrieved.

We filter questions with fewer than five references. For the Replace All and Replace One simulations, we truncate the number of references at ten, keeping the longest references, whereas for the Search simulation, all references are added to the pool.

For the Search simulation, we use \href{https://platform.openai.com/docs/api-reference/vector-stores}{OpenAI vector stores} to implement RAG. We use the default chunking strategy, although as part of our research, we experimented with other chunking strategies and generally found that larger chunks lead to faster collapse. In each round, we add new documents to the pool, and then retrieve the top ten chunks and use them as the context.

As a control, we verified that collapse does not occur when no self-authored references are added. We observe only the variation expected from sampling LLM responses.

We leave the temperature at the default value (1.0) for all experiments.

We use GPT-5.2 for evaluation tasks such as computing the same-answer percentage, extracting entities, and scoring reference quality. We use \href{https://developers.openai.com/api/docs/models/gpt-5.2-chat-latest}{GPT-5.2 Chat} for many experiments because it was the version of GPT-5.2 that ChatGPT used when we ran them.

To distinguish AI-generated and human-written original references, we use GPTZero, an AI-content detector, following a methodology similar to our AI content paper~\citep{paredes2026aiwrites}. We classify a reference as AI-generated when GPTZero labels it as AI or Mixed.

\subsection{Costs}
\label{sec:costs}
Each simulation requires hundreds of calls to an LLM API, specifically:

\begin{itemize}
  \item Generating responses to the question (10 per round)
  \item Generating articles based on responses (1--10 per round)
  \item Computing the same-answer percentage metric (10 per round)
  \item Extracting entities (10 per round)
  \item Extracting relevant content (1 per citation)
\end{itemize}

Many of the calls involve large inputs and outputs (e.g., full articles).  As a result, the simulation often costs \$5 or more per question. Therefore, we cannot reasonably evaluate every combination of dataset, simulation, and model.

For the larger-scale experiments in Section~\ref{sec:large-scale-experiments}, to help reduce costs, we also detect whether the simulation converges. We stop the simulation if the same-answer percentage (defined in Section~\ref{sec:evaluation-metrics}) is 100\% in four consecutive rounds.

\section{Datasets}
\label{sec:datasets}
Getting access to real user prompts is challenging. Instead, we translate Google search keywords to information-seeking question prompts or write them manually.

\subsection{Entity Questions}
A category of particular interest is entity comparison questions, or entity questions for short. Answers to entity questions compare different named entities, such as products, brands, businesses, and locations. For example, ``What are the best restaurants in San Francisco?'' These are the types of prompts that marketers target for answer- and generative-engine optimization.

We collect entity questions from a variety of sources:

\begin{itemize}
  \item The \href{https://graphite.io}{Graphite} prompt tracking tool (often translated from high-volume search keywords)
  \item The categorized keywords dataset from our previous paper~\citep{paredes2025aicontent}
  \item Manually written questions on topics of particular interest
\end{itemize}

\subsection{Editorial Questions}
We also experiment with a more general set of questions that are neither entity comparisons nor factual. Factual questions like ``Who were the draft picks for the Commanders in 2023?'' have one correct answer, and, as a result, responses will already be collapsed (correctly). An example question in this category is ``How can I improve my personal branding?'' We call these editorial questions. These questions are translated from the categorized keywords dataset in our previous paper~\citep{paredes2025aicontent}.

\subsection{Original References}
As described in Section~\ref{sec:implementation-details}, we need references to initialize the simulations, and we gathered them from ChatGPT and Google AI Overviews in January 2026. For each dataset, we filter out questions that do not trigger a web search in ChatGPT or do not produce an AI Overview in Google Search. (In ChatGPT, we could force the use of web search, but we suspect few users do this.)

\subsection{Dataset Statistics}
\begin{table}[htbp]
\centering
\caption{Dataset statistics.}
\label{tab:1}
\small
\begin{tabularx}{\linewidth}{YZ}
\toprule
\textbf{Name} & \textbf{Questions} \\
\midrule
Entity ChatGPT & 843 \\
Editorial ChatGPT & 159 \\
Entity AI Overview & 60 \\
Editorial AI Overview & 45 \\
\bottomrule
\end{tabularx}
\end{table}

The AI Overview questions and ChatGPT questions largely overlap but are not identical due to differences in whether an AI Overview appeared and whether ChatGPT used the web search tool. Across all datasets, there are 1,019 unique questions.

\subsection{AI-Generated Original References}
The references in our dataset were collected in January 2026. We also collected new references for the questions from ChatGPT in early June 2026.

We noted in the introduction that much of the web is AI-generated. Are the original references AI-generated?

Surprisingly, we find that 38.9\% of ChatGPT references in January 2026 and 42.7\% in June 2026 are predicted to be AI-generated. There are more AI-generated references with entity prompts than with editorial, and for both entity and editorial prompts, the percentage of AI-generated references increased from January 2026 to June 2026. We plan to explore this in depth in a forthcoming paper.

\begin{table}[htbp]
\centering
\caption{Estimated percentage of ChatGPT references classified as AI-generated, overall and by prompt type. Brackets show 95\% Wilson confidence intervals over unique reference URLs.}
\label{tab:2}
\small
\begin{tabularx}{\linewidth}{YZZZ}
\toprule
 & \textbf{Overall} & \textbf{Entity} & \textbf{Editorial} \\
\midrule
1/26 ChatGPT & 38.9\% [38.0, 39.9] & 41.1\% [40.1, 42.1] & 20.6\% [18.3, 23.0] \\
6/26 ChatGPT & 42.7\% [41.7, 43.7] & 45.8\% [44.7, 46.8] & 27.5\% [25.4, 29.7] \\
\bottomrule
\end{tabularx}
\end{table}

\section{Evaluation Metrics}
\label{sec:evaluation-metrics}
To measure how responses change over the rounds of the simulation, we use the following metrics:

\textbf{Semantic similarity} measures the similarity of pairs of responses generated in the same round. A larger value means the responses are more similar to each other. It is computed as the average cosine similarity between the embeddings of response pairs.

\textbf{Unique words} measures the number of distinct words across all ten responses in a round.

\textbf{Self-authored citation percentage} measures the percentage of citations of self-authored references. In some plots, we show the self-authored citation percentage alongside the expected percentage based on the proportion of self-authored references in the context. In others, we show the self-authored citation percentage relative to the expected percentage, \(100 \times (\mathrm{observed}-\mathrm{expected})/\mathrm{expected}\), so that values above zero indicate disproportionate citation of self-authored references. We call this metric \textbf{Self-authored citations over expected}. Note that self-authored citations over expected is not defined for the Replace All simulation, as all references are self-authored after the first round. By round 10 in the Replace One simulation, all references are self-authored, so the value drops to 0. In the Search simulation, self-authored references make up an increasing proportion of the reference pool over rounds, so the value approaches 0.

\textbf{Self-authored retrievals} measures the percentage of retrieved chunks that are self-authored in the Search simulation.

\textbf{Same-answer percentage} measures the proportion of pairs of responses that are paraphrases of each other. We use GPT-5.2 to evaluate whether two responses are paraphrases. To reduce costs, we sample 10 of the 45 possible unique pairs of 10 generations in each round.

\subsection{Entity Question Metrics}
For entity questions, we additionally identify the entities mentioned in each response and use them to compute additional metrics.

Entities are often referred to by several names, so in addition to extracting them, we need to canonicalize them. Responses also often mention entities that are not part of the answer. For example, responses to the prompt ``What are the best restaurants in California?'' will include entities such as ``San Francisco'' and ``Dominique Crenn'', but the entities we care about here are restaurant names. We use the following algorithm:

\begin{enumerate}
  \item Extract the entities being listed or compared in each response in each round using an LLM.
  \item Cluster unique entity mentions into groups that refer to the same canonical entity across all simulation rounds.
  \item To capture entity mentions the LLM missed, search for exact matches of entities in the canonical map across all simulation rounds.
\end{enumerate}

\textbf{Unique entities} measures the number of distinct entities mentioned across all responses in a round.

\textbf{Entity ranking similarity} measures the similarity of the rankings of entity mentions across pairs of responses generated in the same round. In particular, we rank the entities by the position of their first mention. We use Kendall's tau as the similarity metric. If two responses mention entities in the same order, the entity ranking similarity is 1.0.

We report the mean of these metrics for each round across all questions in the dataset.

\section{Defining Collapse}
The more similar responses to a prompt are, the more collapsed we consider them to be. Complete collapse would mean that every single response is exactly the same. However, in practice, we care about whether responses essentially say the same thing, even if they are not word-for-word identical.

For entity questions, there is a natural and intuitive representation of collapse: whether the entities being listed or compared are the same. Therefore, we consider responses collapsed when entities are identical across all responses in a round.

For editorial questions, there is no comparable compact representation of the answer, so we instead consider a set of responses to be collapsed when the same-answer percentage is 100\%.

\textbf{Collapsed at start} is the percentage of questions that are collapsed after round 1, before any self-authored references have been introduced.  \textbf{Collapsed at end} is the percentage of questions that are collapsed at the end of the simulation.

\textbf{Rounds collapsed} is the percentage of total rounds across all simulations for a set of questions in which responses were collapsed.

\section{Experiments}
\subsection{Experiment Summary}
\label{sec:experiment-summary}
In Table~\ref{tab:3}, we summarize the experiments presented in this paper. Most simulations end in collapse.

\begin{table}[!htbp]
\centering
\caption{Summary of experiments and collapse rates. The distinction between GPT-5.2 and GPT-5.2 Chat is explained in Section~\ref{sec:implementation-details}.}
\label{tab:3}
\footnotesize
\setlength{\tabcolsep}{3pt}
\begin{tabularx}{\linewidth}{YZZZZZZ}
\toprule
\textbf{Simulation} & \textbf{Model} & \textbf{Dataset} & \textbf{Questions} & \textbf{Collapsed at Start} & \textbf{Collapsed at End} & \textbf{Rounds Collapsed} \\
\midrule
Replace All & GPT-5.2 Chat & Entity ChatGPT & 101 & 2.97\% & 88.12\% & 68.51\% \\
Replace All & GPT-5.2 Chat & Editorial ChatGPT & 57 & 29.82\% & 91.23\% & 79.65\% \\
Replace One & GPT-5.2 Chat & Entity ChatGPT & 101 & 2.97\% & 88.12\% & 67.18\% \\
Replace One & GPT-5.2 Chat & Editorial ChatGPT & 57 & 21.05\% & 94.74\% & 81.67\% \\
Search & GPT-5.2 Chat & Entity ChatGPT & 101 & 1.98\% & 77.23\% & 62.31\% \\
Search & GPT-5.2 Chat & Editorial ChatGPT & 57 & 31.58\% & 75.44\% & 73.63\% \\
Replace One & Gemini 3 Pro & Entity AI Overview & 60 & 3.33\% & 80.00\% & 43.83\% \\
Replace One & Claude Sonnet 4.5 & Entity AI Overview & 60 & 1.67\% & 91.67\% & 64.33\% \\
Replace One & Gemini 3 Pro & Editorial AI Overview & 45 & 6.67\% & 68.89\% & 42.56\% \\
Replace One & Claude Sonnet 4.5 & Editorial AI Overview & 45 & 11.11\% & 97.78\% & 69.33\% \\
Search & GPT-5.2 & Entity ChatGPT & 742 & 7.01\% & 73.85\% & 59.93\% \\
Search & GPT-5.2 & Editorial ChatGPT & 102 & 7.84\% & 83.33\% & 65.85\% \\
\bottomrule
\end{tabularx}
\end{table}

\begin{samepage}
We consistently find the following across experiments:

\begin{itemize}
  \item The number of unique words and entities mentioned across responses in a round decreases over the course of the simulation.
  \item The semantic similarity between responses and the similarity between the entities mentioned increase over the course of the simulation.
  \item When both original and self-authored references are available, the models disproportionately cite the self-authored references.
\end{itemize}

In the following sections, we review the results in detail. All raw data from experiments is available at \url{https://drive.google.com/drive/folders/10SnsdKQJDZ7IudjqKcxxcpSHcc2OnNDl?usp=drive_link}.
\end{samepage}

\subsection{Comparison of Simulations}
We first compare the three simulations using GPT-5.2 Chat with 101 prompts from Entity ChatGPT and 57 from Editorial ChatGPT.

After running the Replace All simulation for ten rounds, replacing all the references with self-authored references after each round, we find that 88.1\% of entity questions and 91.2\% of editorial questions collapse. The semantic and entity ranking similarities across responses in a round increase as the simulation progresses. The same-answer percentage also increases. Unique words and entities decrease.

In the Replace One simulation, we replace one reference with a self-authored reference in each round. Because the Replace One simulation has a mix of self-authored and original references in the early rounds (while Replace All replaces all references after each round), we would expect collapse to be slower and less frequent. Instead, we find that the collapse begins almost immediately, and the Replace One simulation leads to collapse as often as Replace All for entity questions (88.1\%) and editorial questions (94.7\%).

We find that self-authored references disproportionately influence the answer. Figures~\ref{fig:simulation-entity-answer-citation-row} and~\ref{fig:simulation-editorial-metrics-grid} show that the rate of self-authored references cited in the response far exceeds expectations based on the number of self-authored versus original references. We investigate bias in detail in Section~\ref{sec:self-authored-influence}.

Finally, in the Search simulation, rather than replacing references with self-authored articles, we add them to a content pool and retrieve relevant documents from it. The collapse rate is lower in the Search simulation, but the absolute rates remain high: 77.2\% and 75.4\% for entity and editorial questions, respectively. This demonstrates that collapse occurs frequently even in our most realistic simulation.

Figure~\ref{fig:simulation-collapse-grid} displays the collapsed at start, collapsed at end, and rounds collapsed rates for each simulation. For error bars, we use Wilson score intervals.

Note that collapsed at start rates may vary between Search and the other simulations because Search uses chunks, and rates may also vary generally due to random variation in responses.

Panels (c) and (d) of Figure~\ref{fig:simulation-collapse-grid} plot the percentage of questions collapsed per round for different simulations. This illustrates that Replace One and Search start to collapse immediately, even with a small percentage of self-authored references. By round 2, after a single self-authored reference has been added (10--20\% of the reference pool), 22.8\% of Replace One entity questions have collapsed, not much lower than 28.7\% for Replace All, where every reference is self-authored. A small share of self-authored references produces most of the collapse that full replacement does, underscoring their disproportionate influence. The same pattern holds for editorial questions (50.9\% vs. 64.9\%).

\begin{plotblock}
\plotpanel{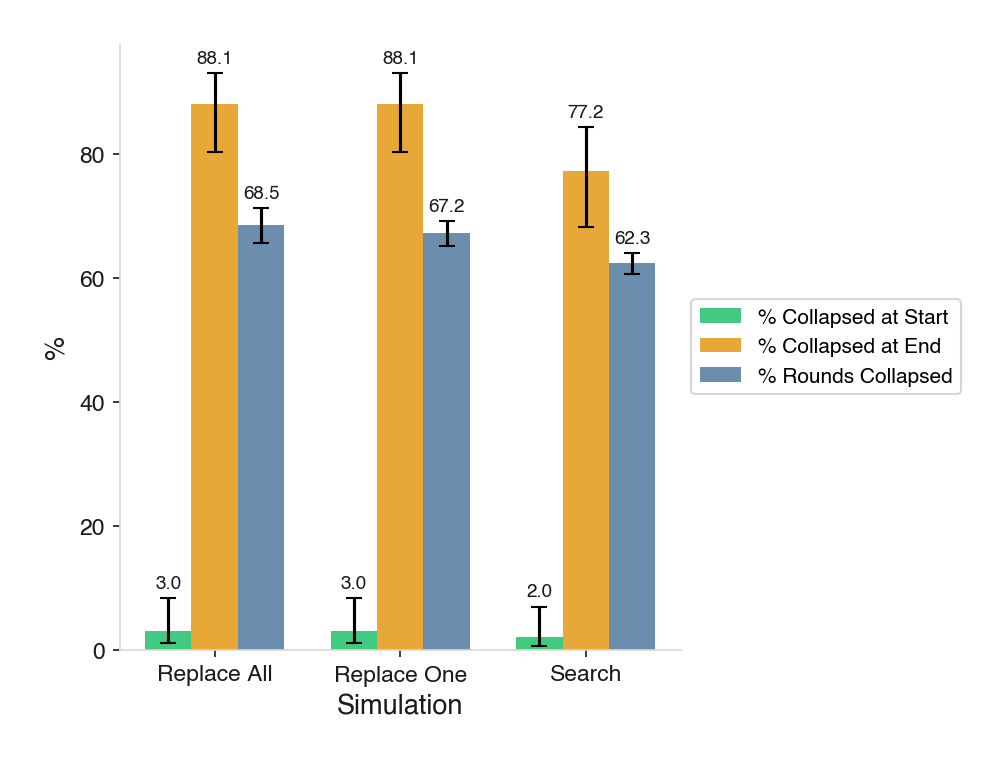}{Collapse by simulation, entity questions.}{fig:image71}
\hfill
\plotpanel{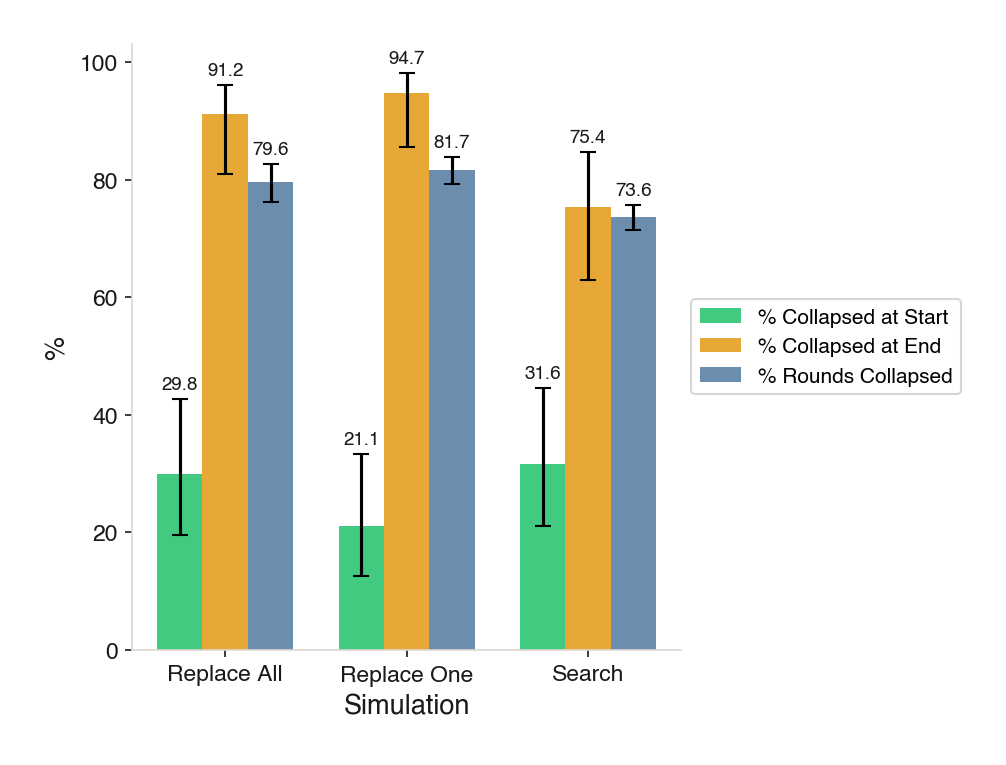}{Collapse by simulation, editorial questions.}{fig:image7}

\medskip
\plotpanel{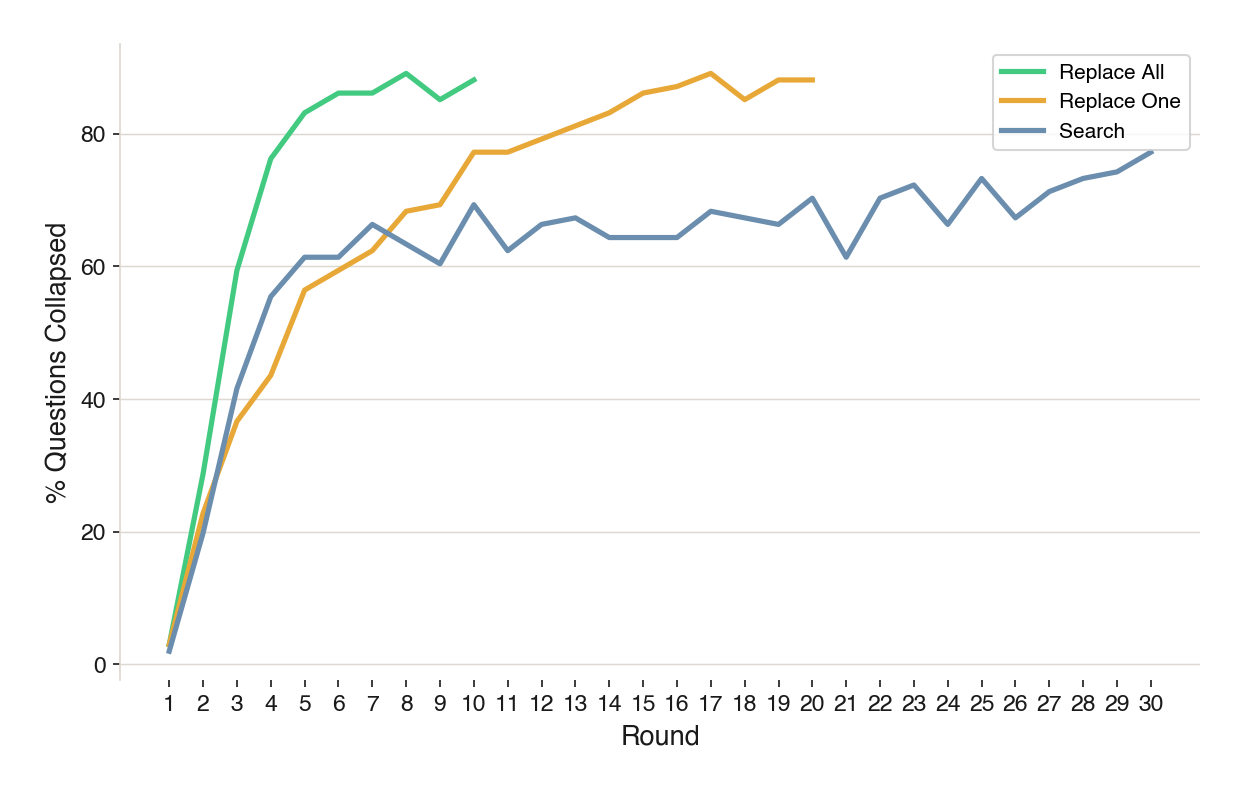}{Questions collapsed by round, entity questions.}{fig:image54}
\hfill
\plotpanel{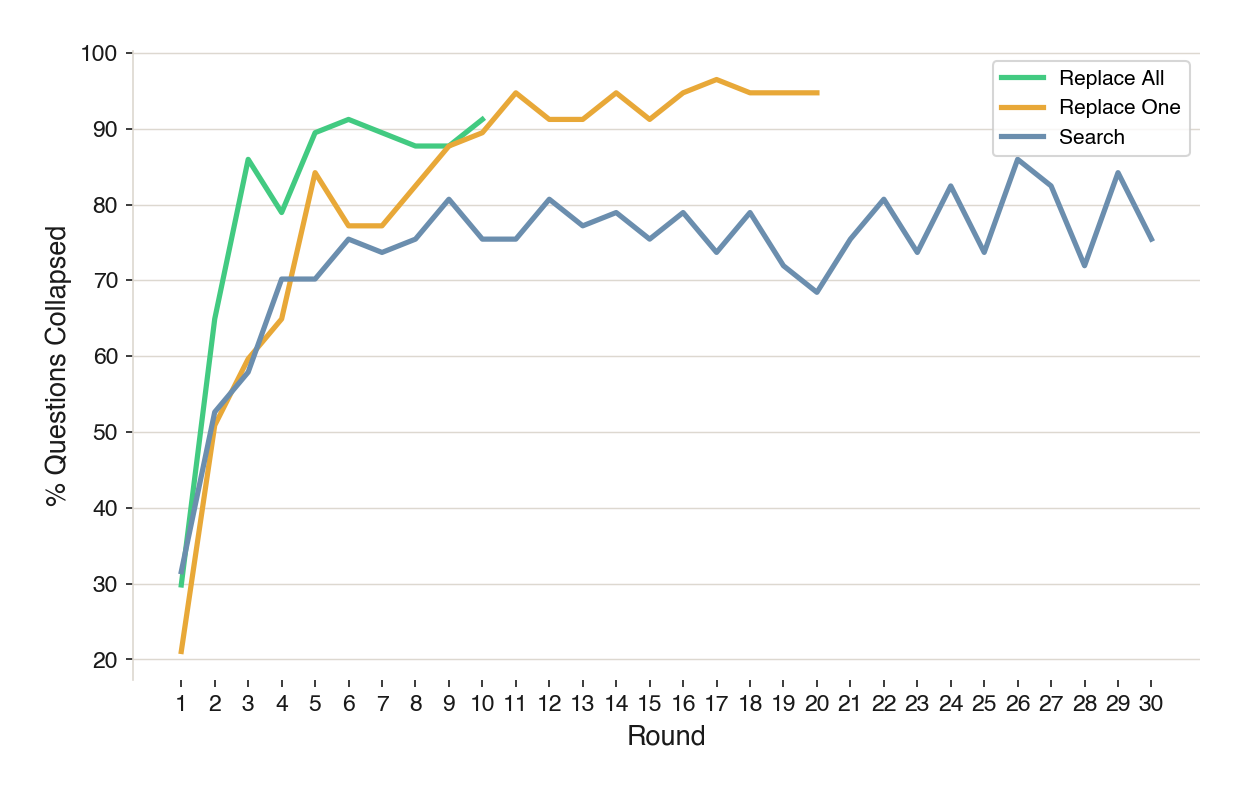}{Questions collapsed by round, editorial questions.}{fig:image19}
\caption{Collapse rates across simulation designs using GPT-5.2 Chat.}
\label{fig:simulation-collapse-grid}
\end{plotblock}

Next, we present additional metrics that illustrate the simulations' behavior. For error bands for metric-by-round line charts, we use one standard error above and below the mean.

\subsubsection{Entity Questions}
The unique sets of entities and words mentioned across responses decrease over rounds, indicating a decline in response diversity. The semantic and entity-ranking similarities among pairs of responses generated in the same round increase across rounds.

\begin{plotblock}
\plotpanel{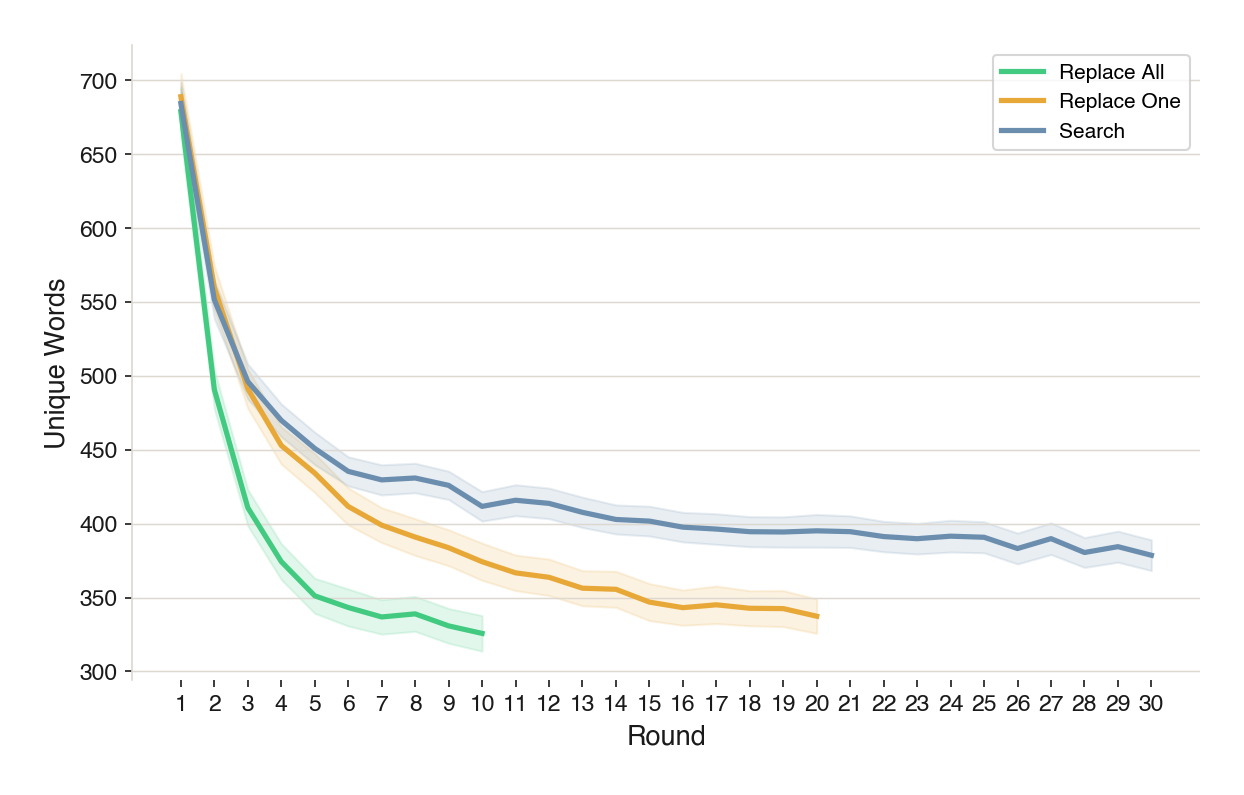}{Unique words by round.}{fig:image63}
\hfill
\plotpanel{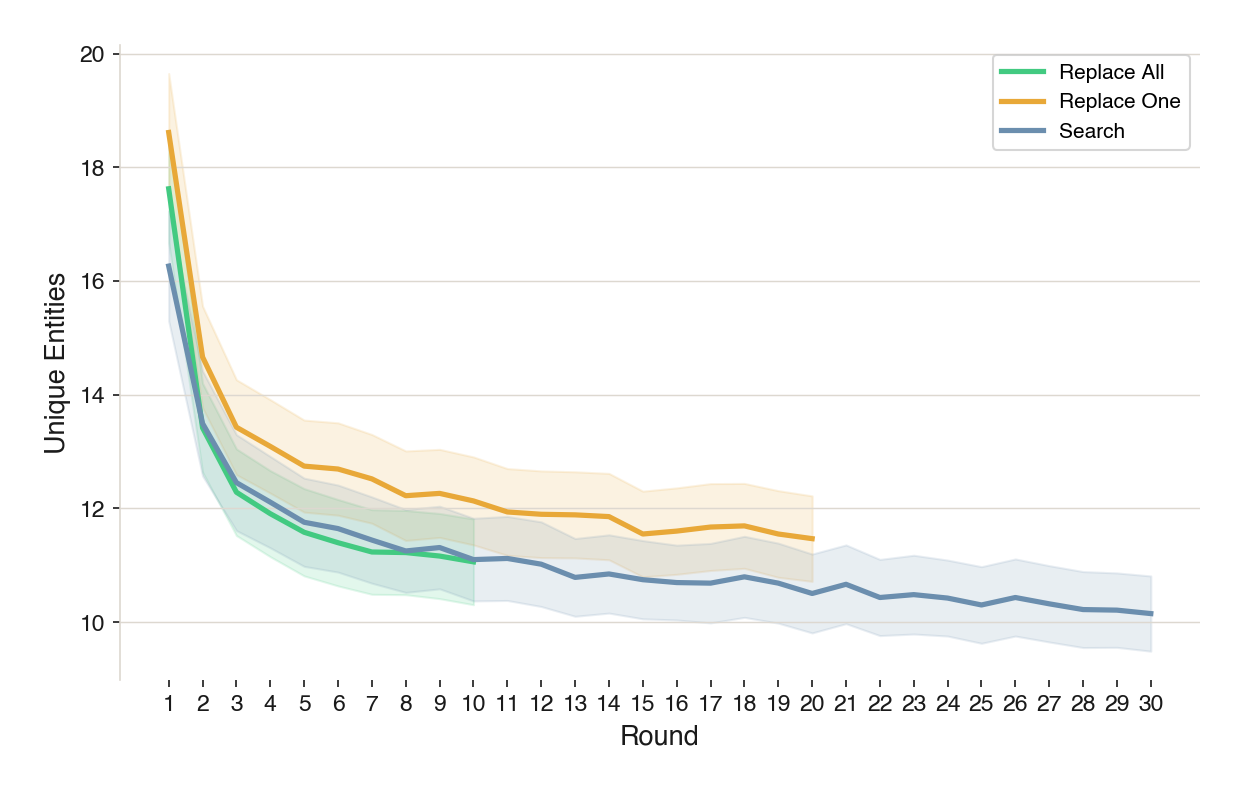}{Unique entities by round.}{fig:image2}

\medskip
\plotpanel{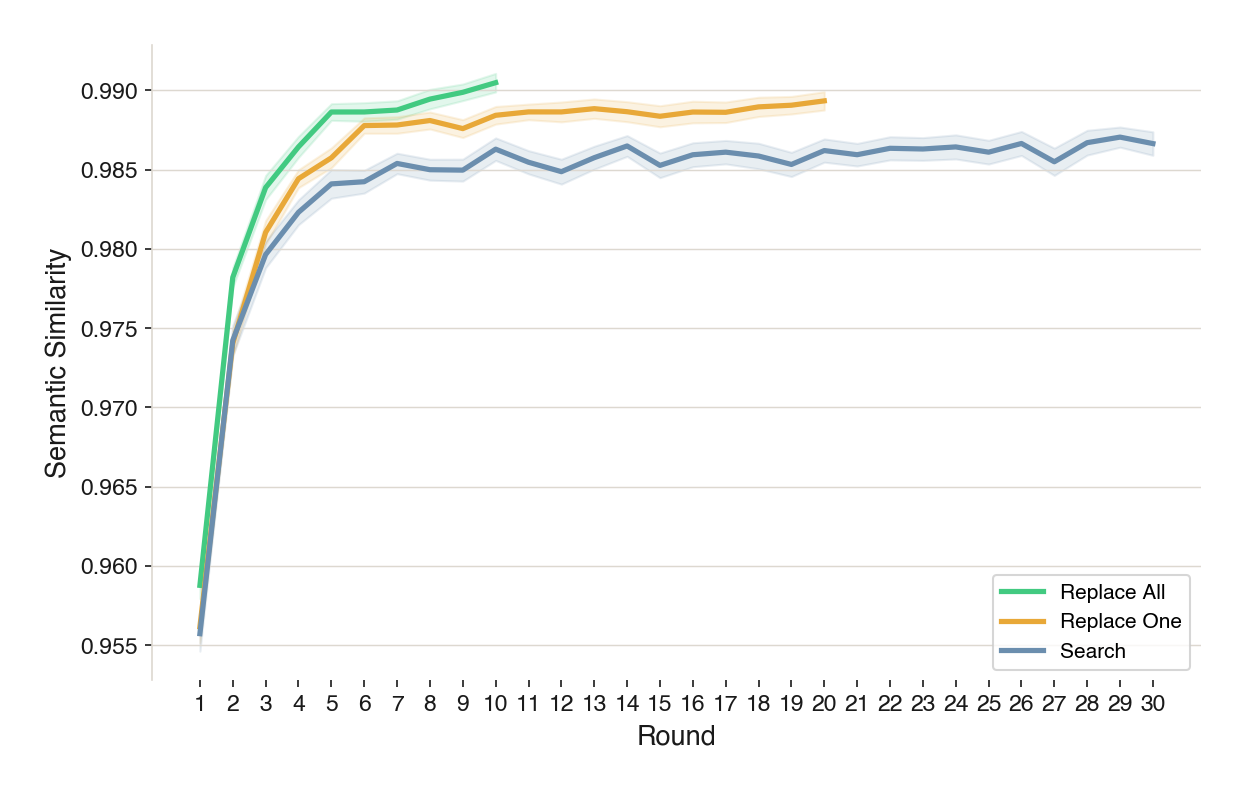}{Semantic similarity by round.}{fig:image10}
\hfill
\plotpanel{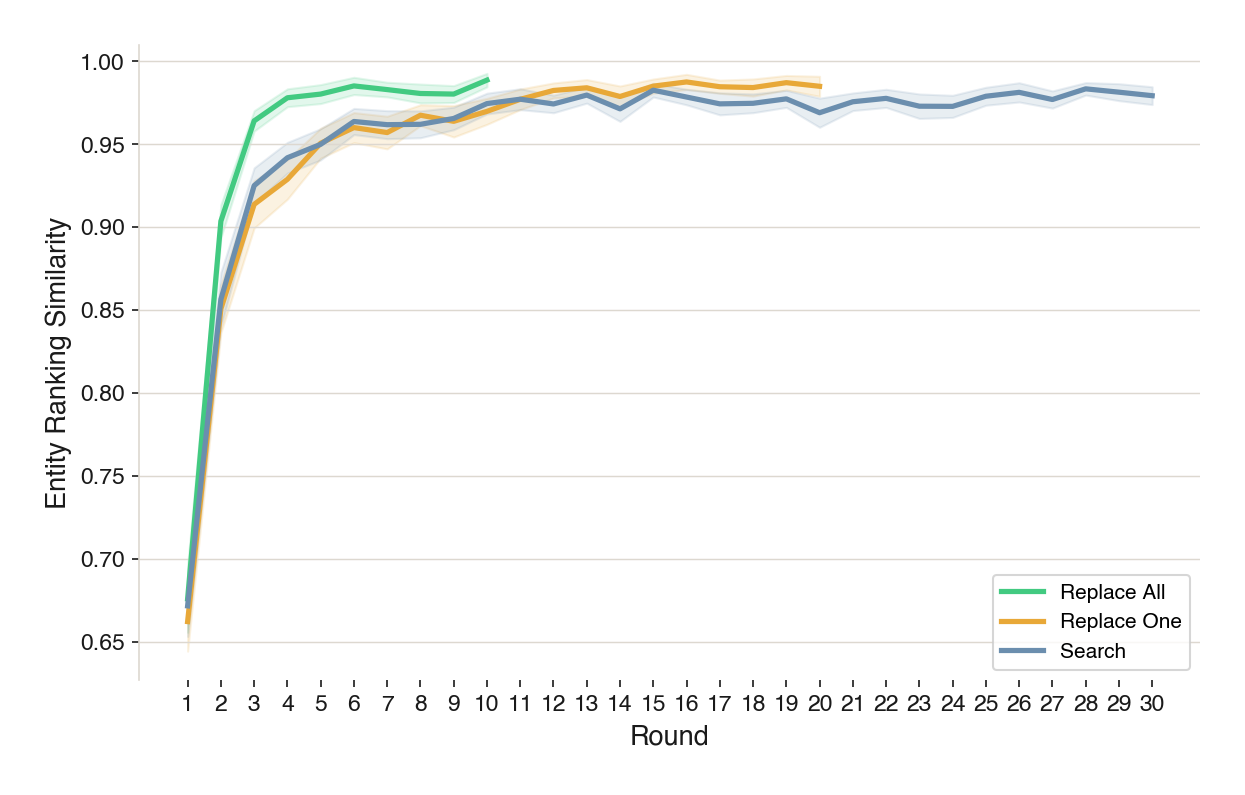}{Entity ranking similarity by round.}{fig:image9}
\caption{Core metrics for entity questions across simulation designs. The numbers of unique words and entities decrease over rounds, indicating declining response diversity, while semantic and entity-ranking similarities increase.}
\label{fig:simulation-entity-metrics-grid}
\end{plotblock}

The same-answer percentage increases from less than 30\% in round 1 to over 90\%. The percentage of self-authored citations increases and is generally higher than expected based on their proportion of the reference pool.

\begin{plotblock}
\plotpanel{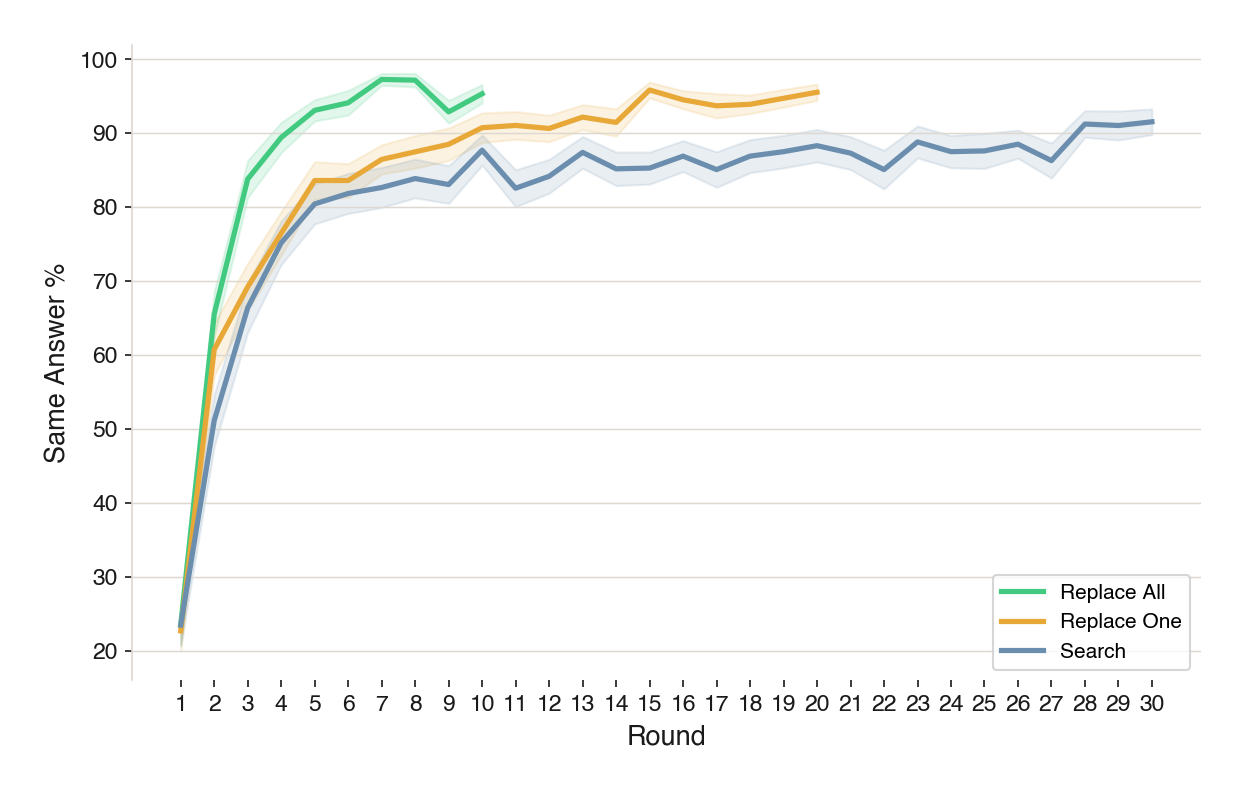}{Same-answer rate by round.}{fig:image40}
\hfill
\plotpanel{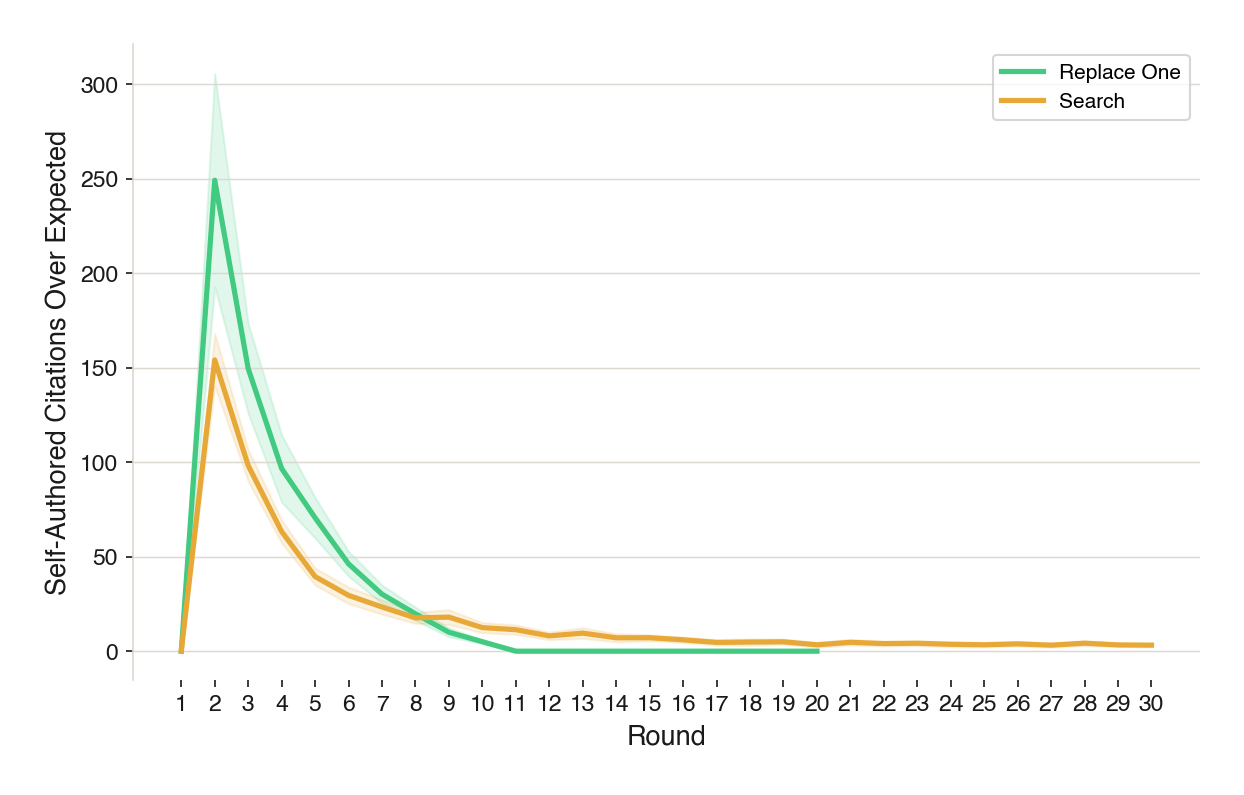}{Self-authored citations over expected.}{fig:image22}
\caption{Same-answer rate and self-authored citations for entity questions. The same-answer rate rises from less than 30\% in round 1 to over 90\%, while self-authored citations increase and generally exceed the rate expected from their share of the reference pool.}
\label{fig:simulation-entity-answer-citation-row}
\end{plotblock}

As with citations, the Search simulation retrieves more self-authored references than would be expected by chance. Replace All and Replace One do not perform retrieval.

\begin{plotblock}
\includegraphics[width=\plotwidth,height=\plotheight,keepaspectratio]{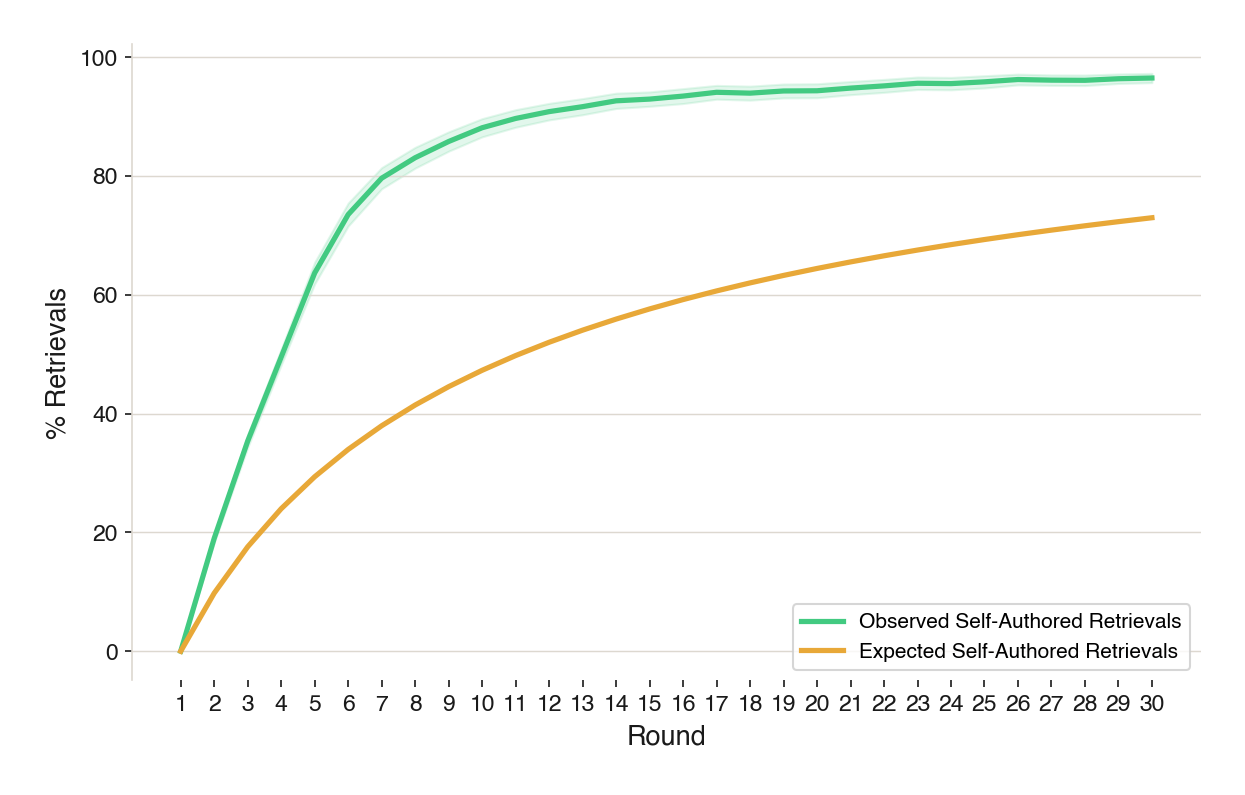}
\caption{Observed and expected self-authored retrievals by round for entity questions in Search using GPT-5.2 Chat. As with citations, more self-authored references are retrieved than expected by chance. Replace All and Replace One do not perform retrieval.}
\label{fig:image74}
\end{plotblock}

\subsubsection{Editorial Questions}
For editorial questions, semantic similarity increases, while the number of unique words decreases, across all simulations. The same-answer percentage increases over rounds. Self-authored references are again disproportionately cited.

\begin{plotblock}
\plotpanel{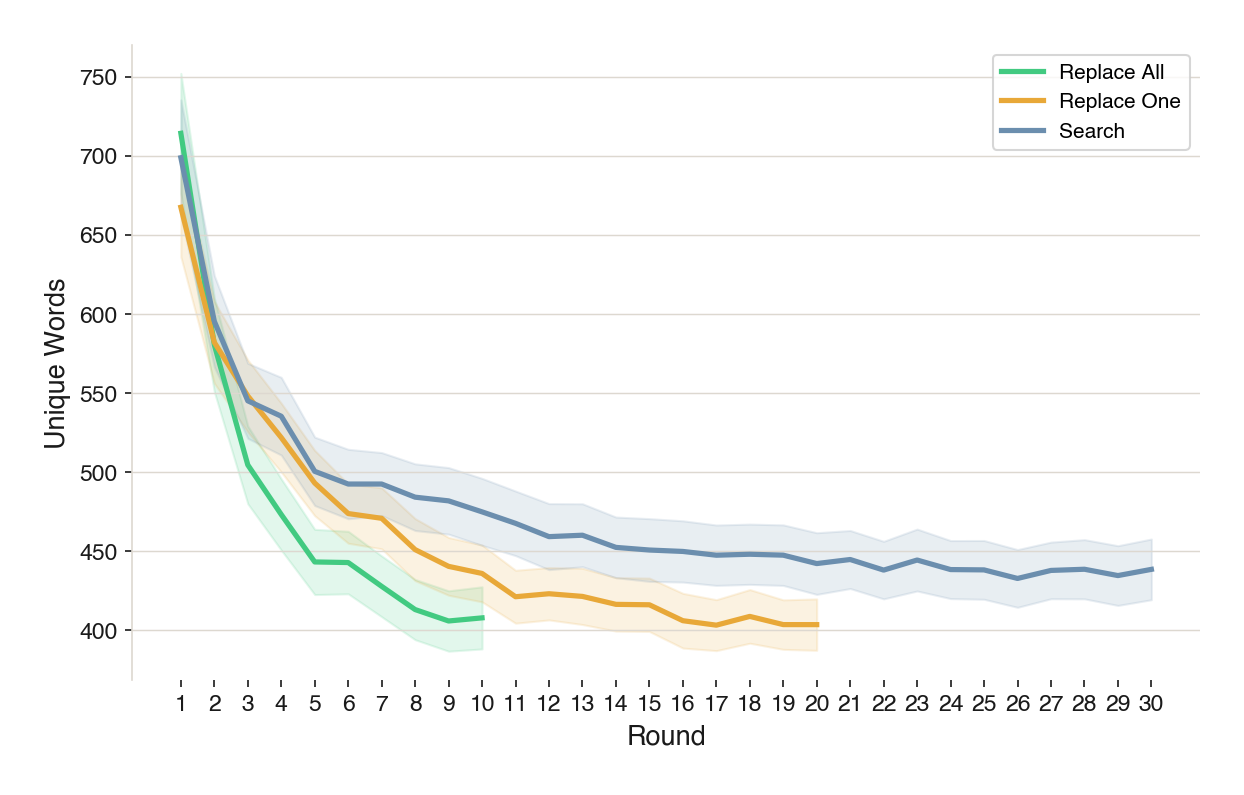}{Unique words by round.}{fig:image39}
\hfill
\plotpanel{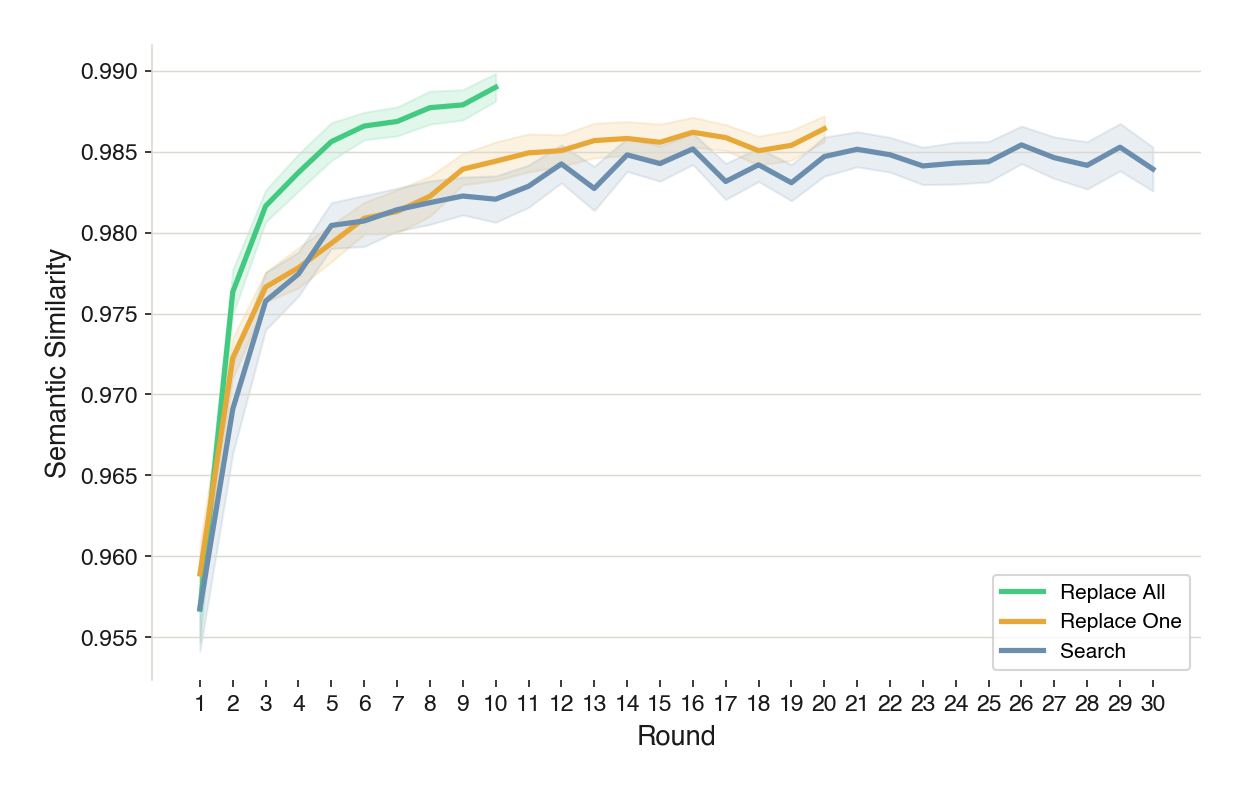}{Semantic similarity by round.}{fig:image36}

\medskip
\plotpanel{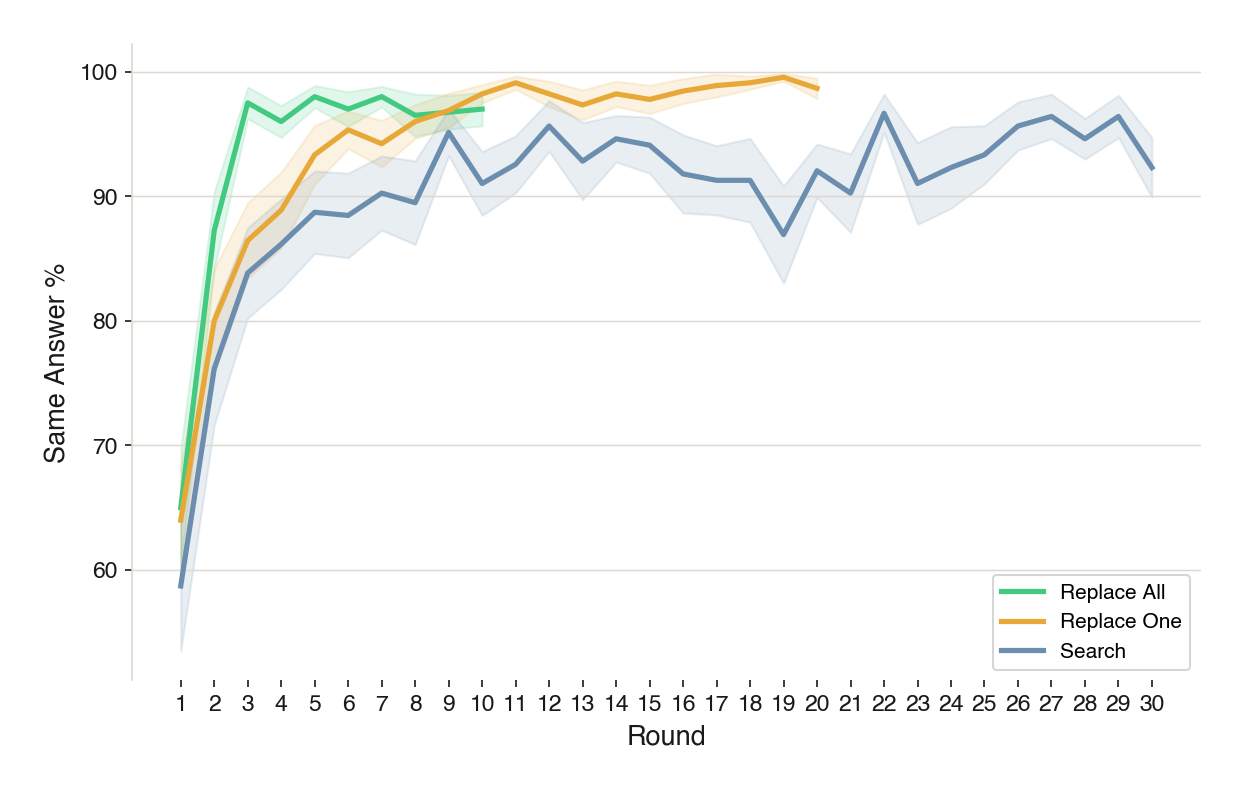}{Same-answer rate by round.}{fig:image3}
\hfill
\plotpanel{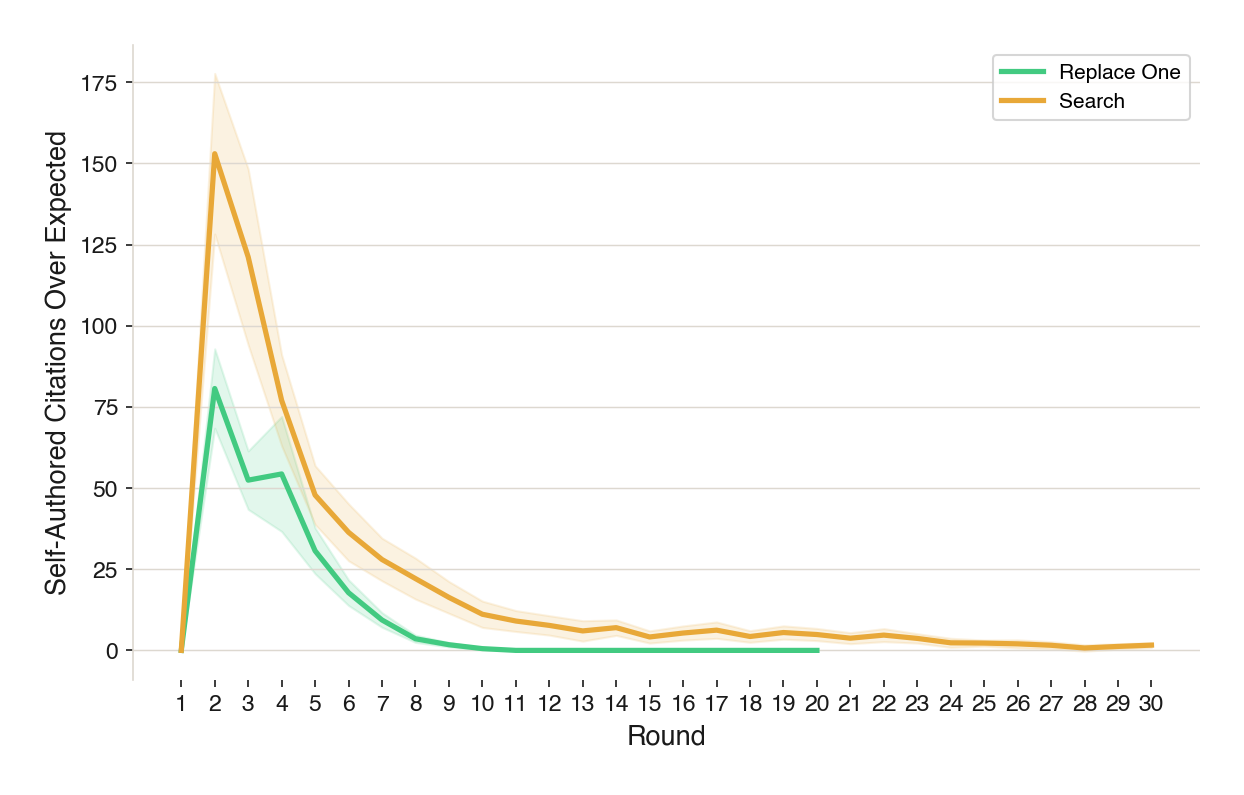}{Self-authored citations over expected.}{fig:image11}
\caption{Metrics for editorial questions across simulation designs. Unique words decrease while semantic similarity and the same-answer rate increase. Self-authored references are again cited disproportionately.}
\label{fig:simulation-editorial-metrics-grid}
\end{plotblock}

Figure~\ref{fig:video-games-visibility-grid} shows visibility snapshots for entities during the Replace All simulation for the question ``What are the best video games of all time today?'' In the first round, in which the original references are used to generate responses, we observe a wide distribution of entities. Some, like ``Red Dead Redemption 2'', are mentioned in 100\% of responses (10/10). Others, like ``DOOM'', are mentioned in only 10\% of responses (1/10). After converting the model's first-round answers into articles and using them as references, the distribution begins to shift. By the fifth round, the distribution is very different from the initial distribution. All entities either always appear or never appear.

\begin{plotblock}
\plotpanel{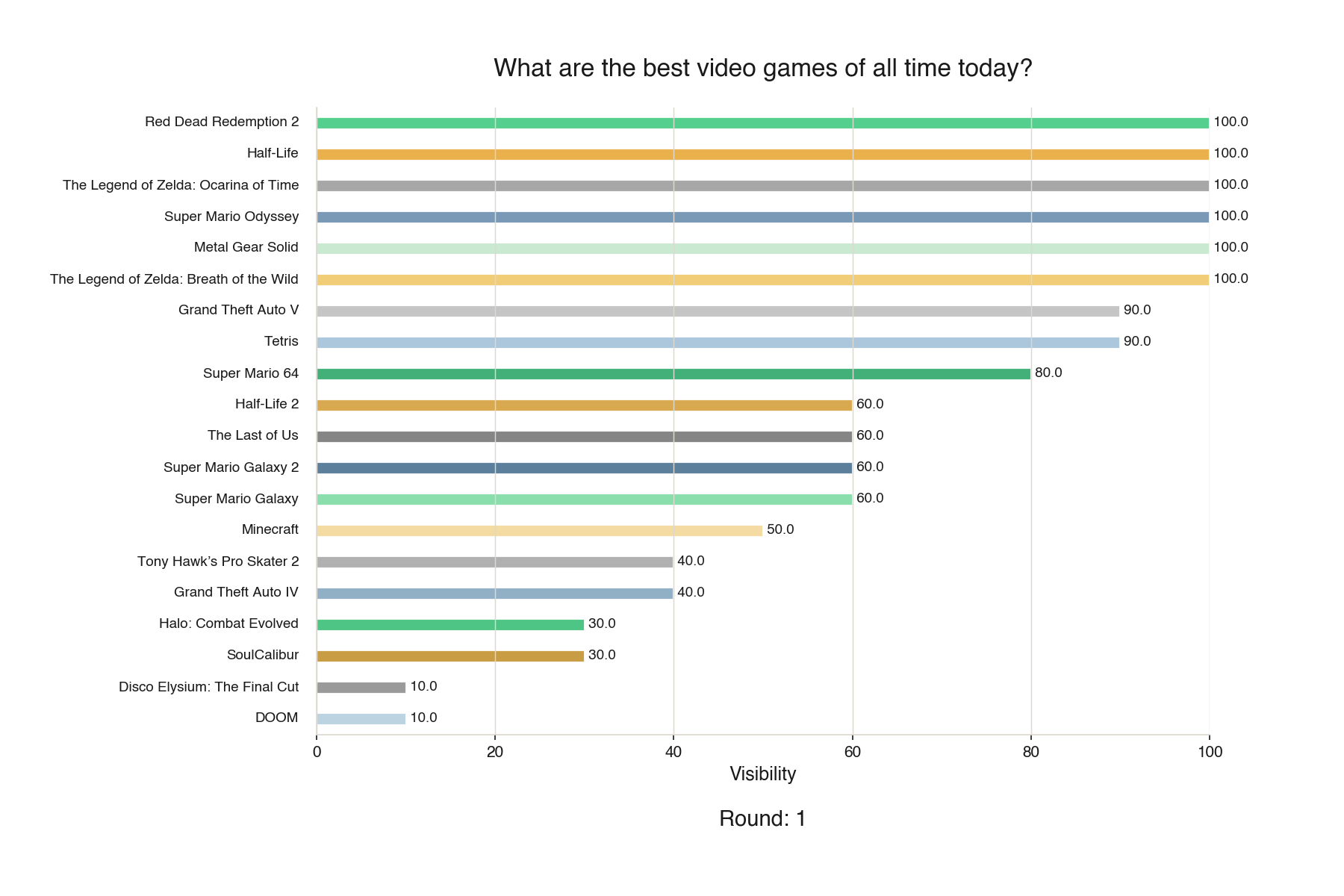}{Round 1.}{fig:image48}
\hfill
\plotpanel{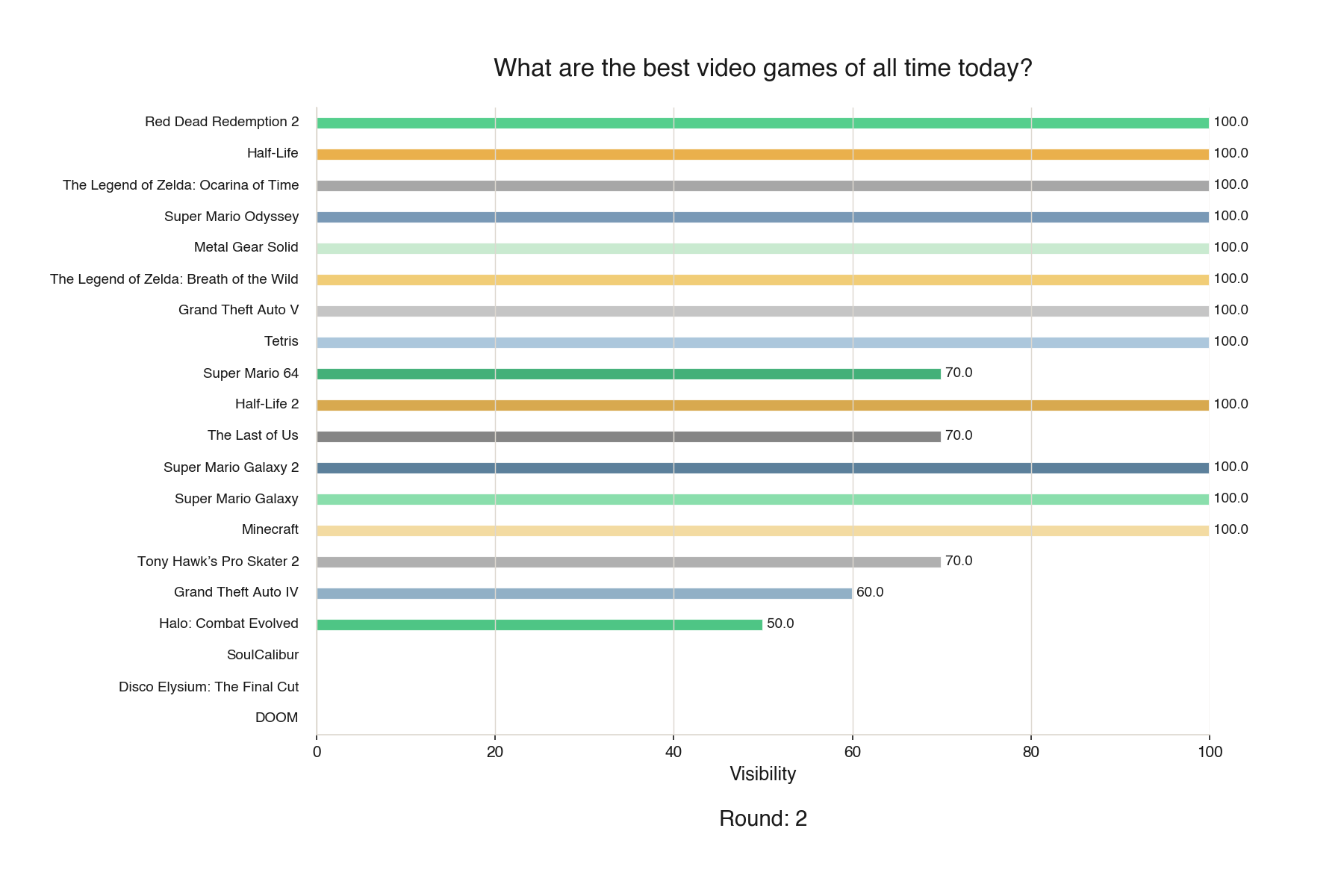}{Round 2.}{fig:image76}

\medskip
\plotpanel{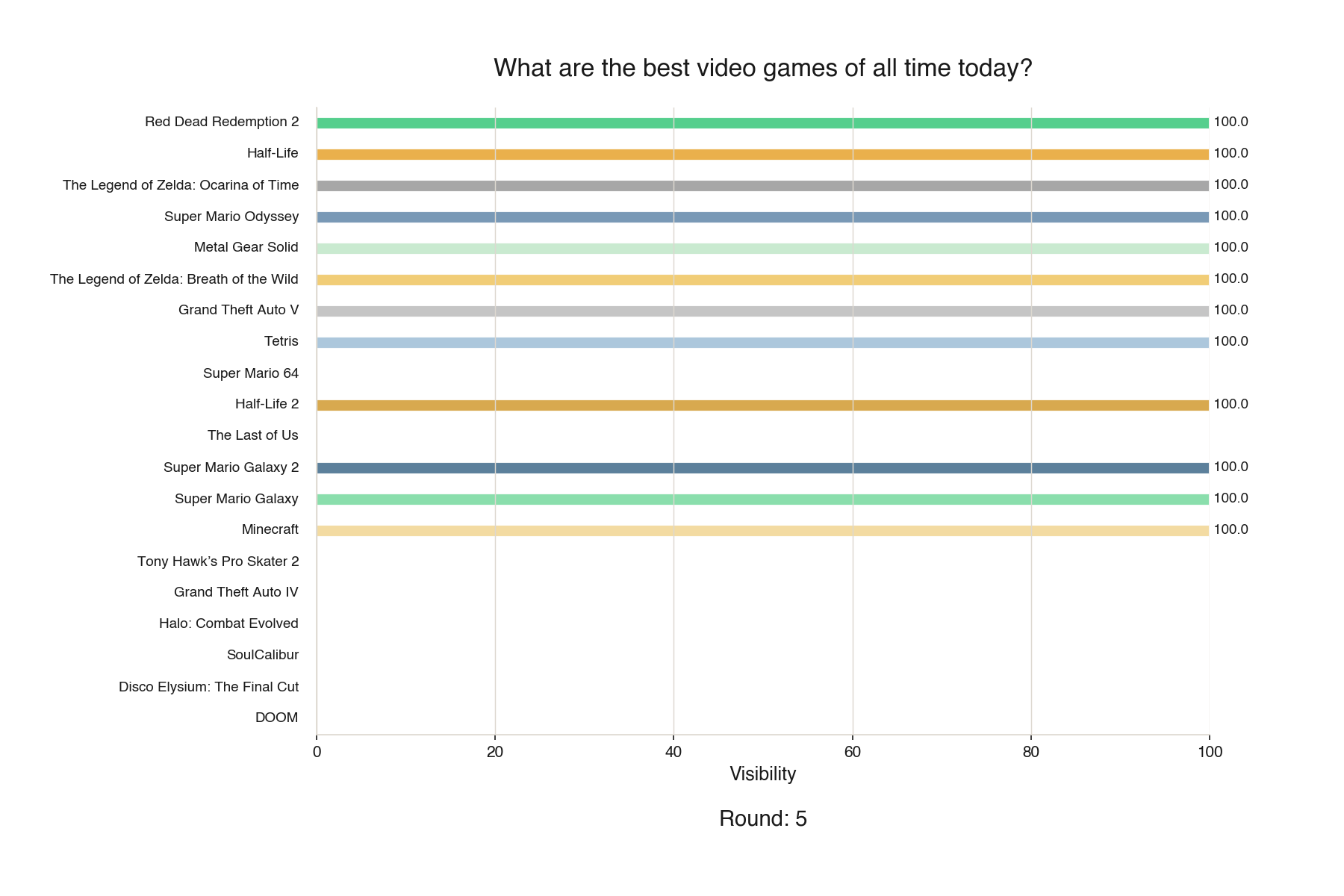}{Round 5.}{fig:image28}
\hfill
\plotpanel{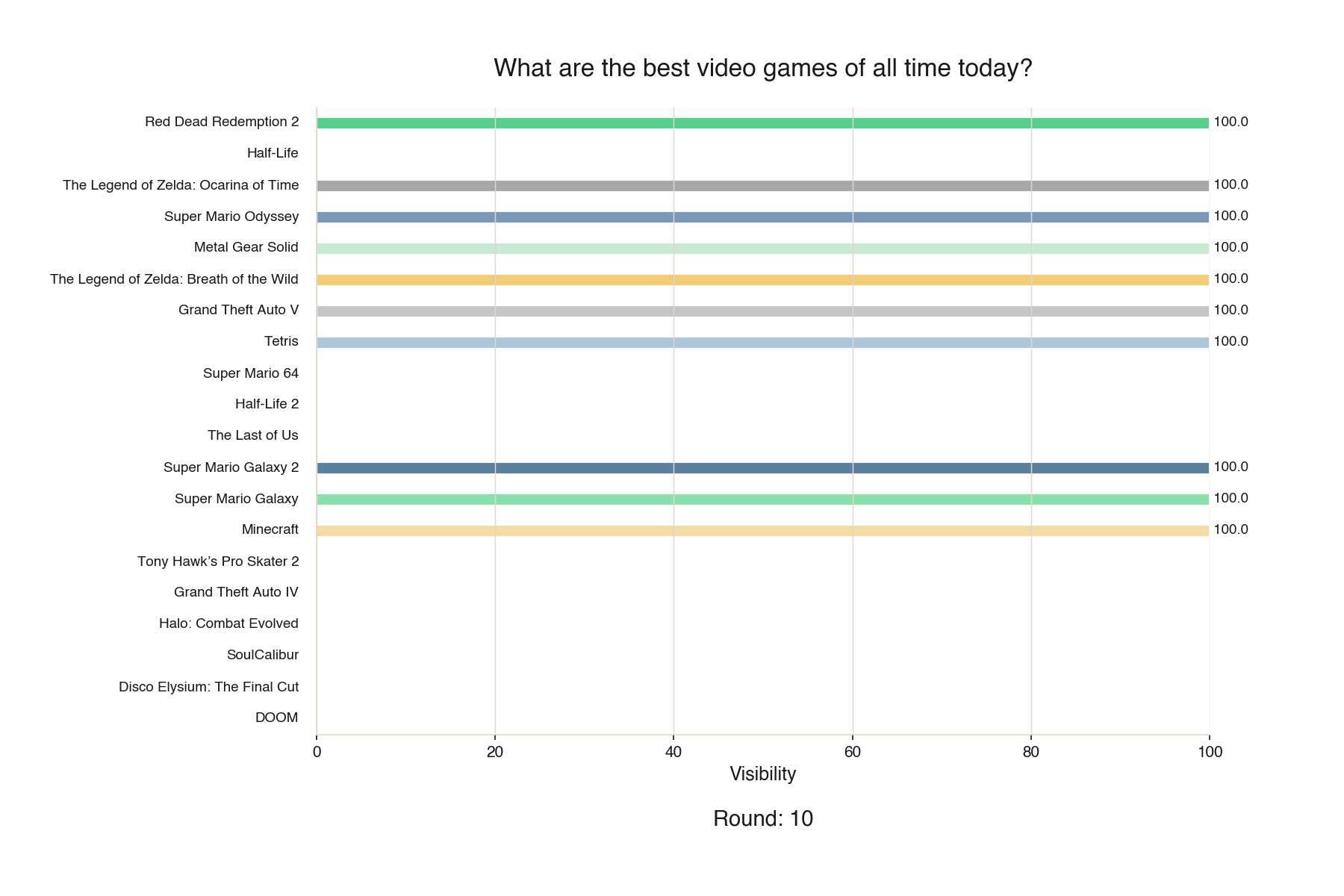}{Round 10.}{fig:image77}
\caption{Entity visibility across selected rounds for ``What are the best video games of all time today?'' Initially, visibility is broadly distributed; after the first-round answers become references, the distribution shifts, and by round 5 every entity either always or never appears.}
\label{fig:video-games-visibility-grid}
\end{plotblock}

By round 10, more entities have dropped out, including ``Half-Life'', which at the beginning of the simulation had 100\% visibility. This demonstrates the random nature of RAG collapse.

Next, we examine the actual responses to this question at rounds 1 and 10 (the final round).

At round 1, the first few games are the same, but after that, the answers differ. Figure~\ref{fig:image26} shows two responses from round 1.

\begin{plotblock}
\includegraphics[width=\plotwidth,height=\plotheight,keepaspectratio]{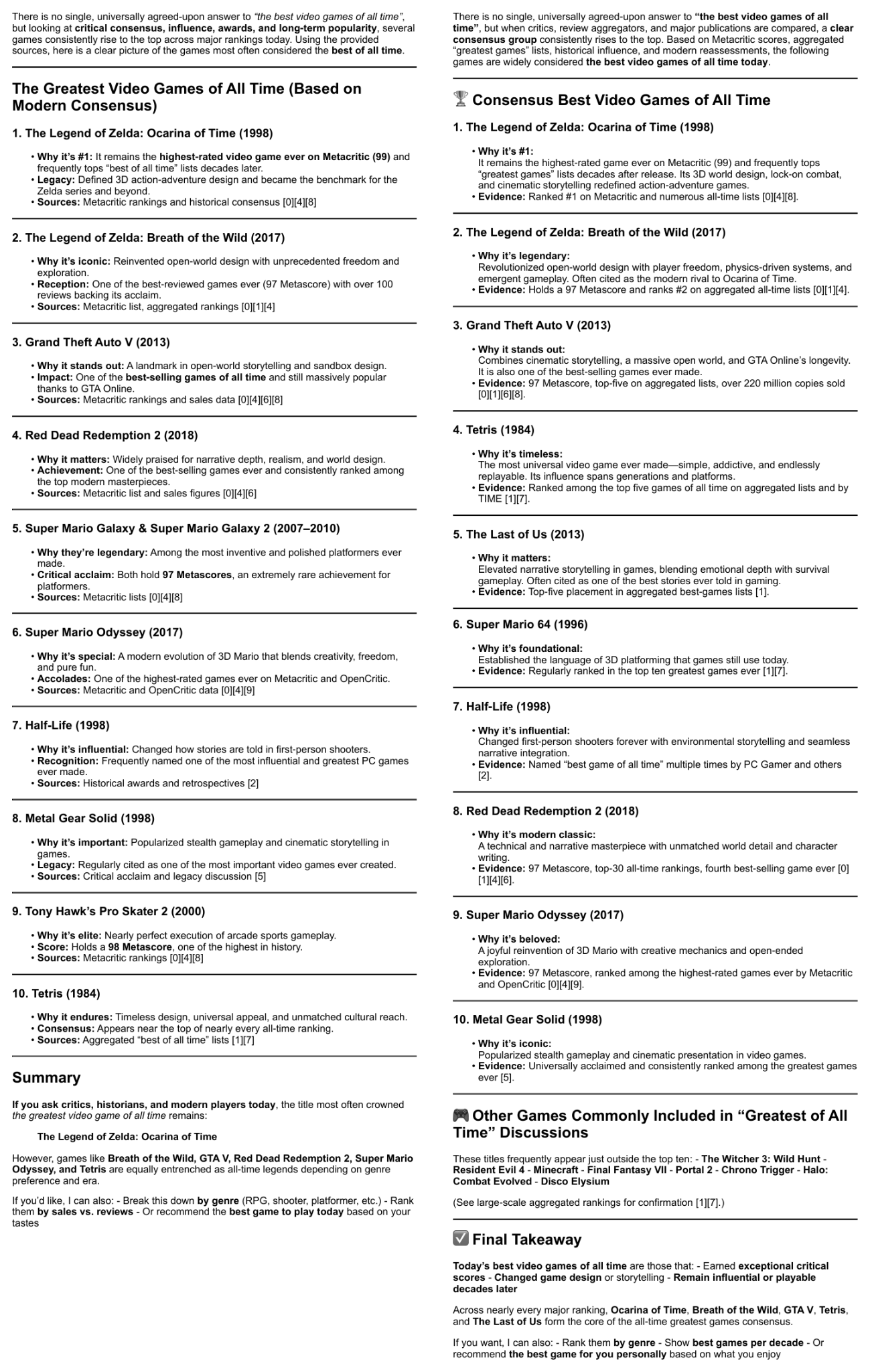}
\caption{Round 1 responses: the first few games are the same, but the answers then diverge.}
\label{fig:image26}
\end{plotblock}

At round 10, the answers have slightly different formatting, but the games mentioned are exactly the same and in the same order, and the text is nearly word-for-word identical. Figure~\ref{fig:image18} shows two responses from round 10.

\begin{plotblock}
\includegraphics[width=\plotwidth,height=\plotheight,keepaspectratio]{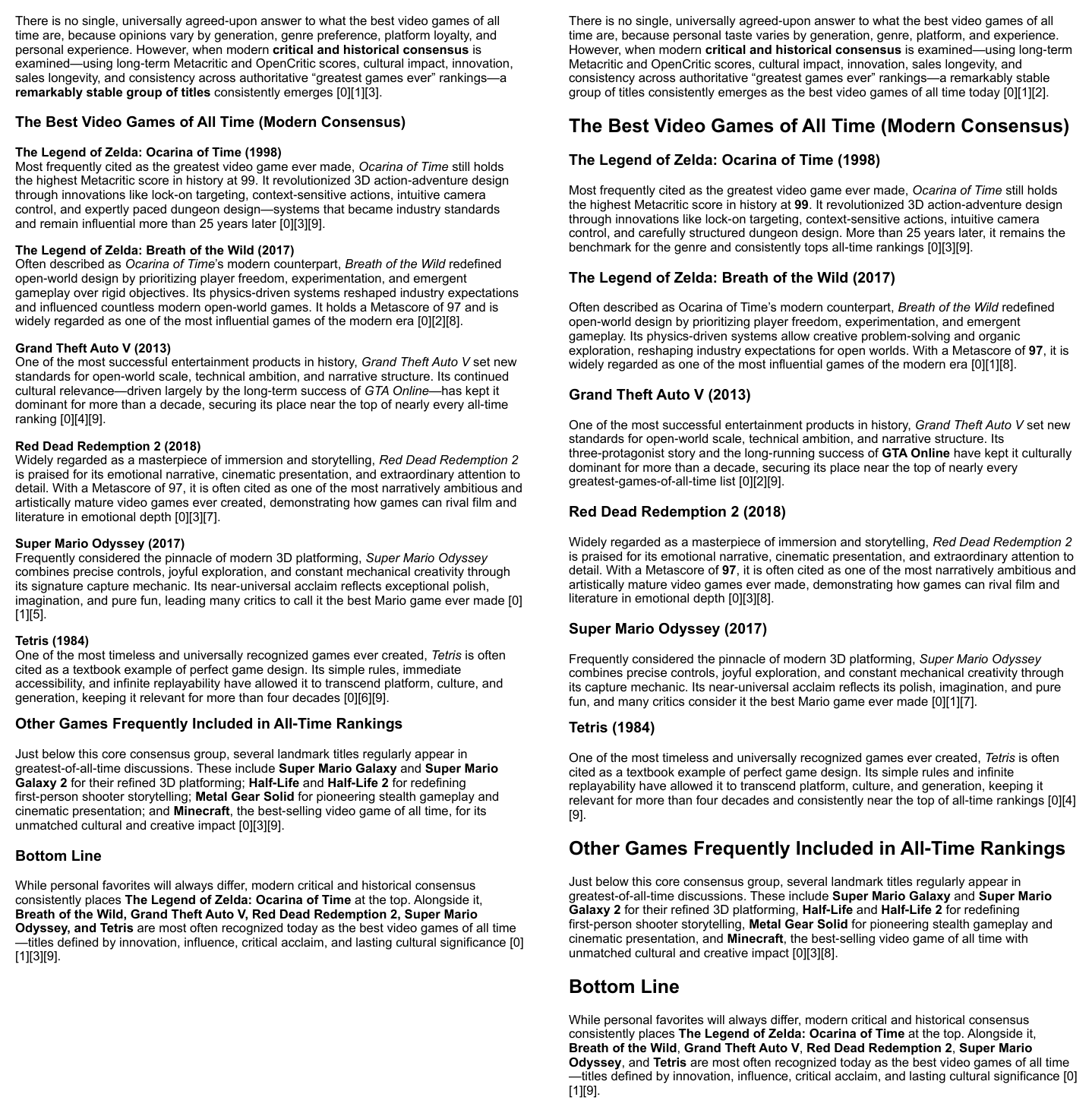}
\caption{Round 10 responses: despite slight formatting differences, the games and their order are identical, and the text is nearly word-for-word the same.}
\label{fig:image18}
\end{plotblock}

Similar to model collapse via retraining, at the end of the simulations, responses differ substantially from the original responses. Rather than having a variety of words and entities in different orders, the words are almost the same, and the same set of entities is mentioned in the same order. Some entities with 100\% initial visibility drop out completely.

\subsection{Comparison of Models}
\label{sec:comparison-models}
Next, we compare the results of the Replace One simulation using different models. To ensure a fair comparison, we restrict to the questions that appear in both the ChatGPT and AI Overview datasets. This yields 41 entity questions and 45 editorial questions. This means the results here vary from those presented in Section~\ref{sec:experiment-summary} and Appendix~\ref{sec:appendix}. Note that we use different initial references for GPT-5.2 Chat (from the ChatGPT UI) and Gemini 3 Pro and Claude Sonnet 4.5 (from Google AI Overviews), because we wanted to pair each model with references it would actually use. However, this means there may be differences in the quality of references.

Overall, we observe collapse with all three models, demonstrating that RAG collapse is not specific to OpenAI models. Gemini 3 Pro has lower collapse rates (though in most cases with overlapping confidence intervals) and a smaller drop in unique entities per round.

\begin{plotblock}
\plotpanel{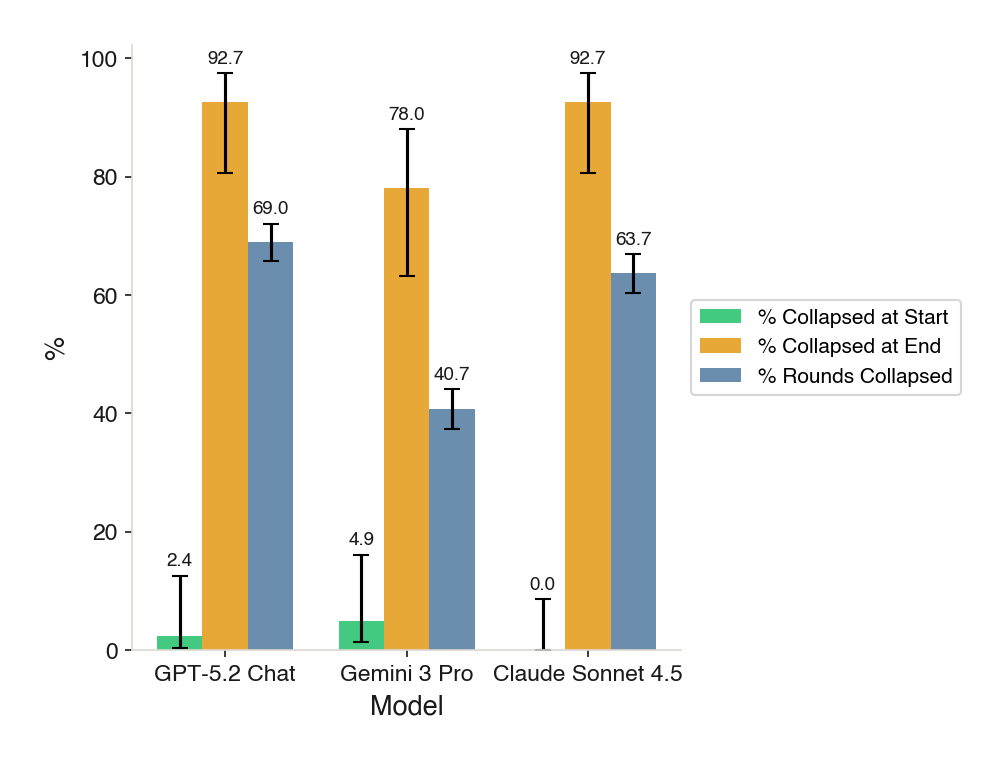}{Entity questions.}{fig:image72}
\hfill
\plotpanel{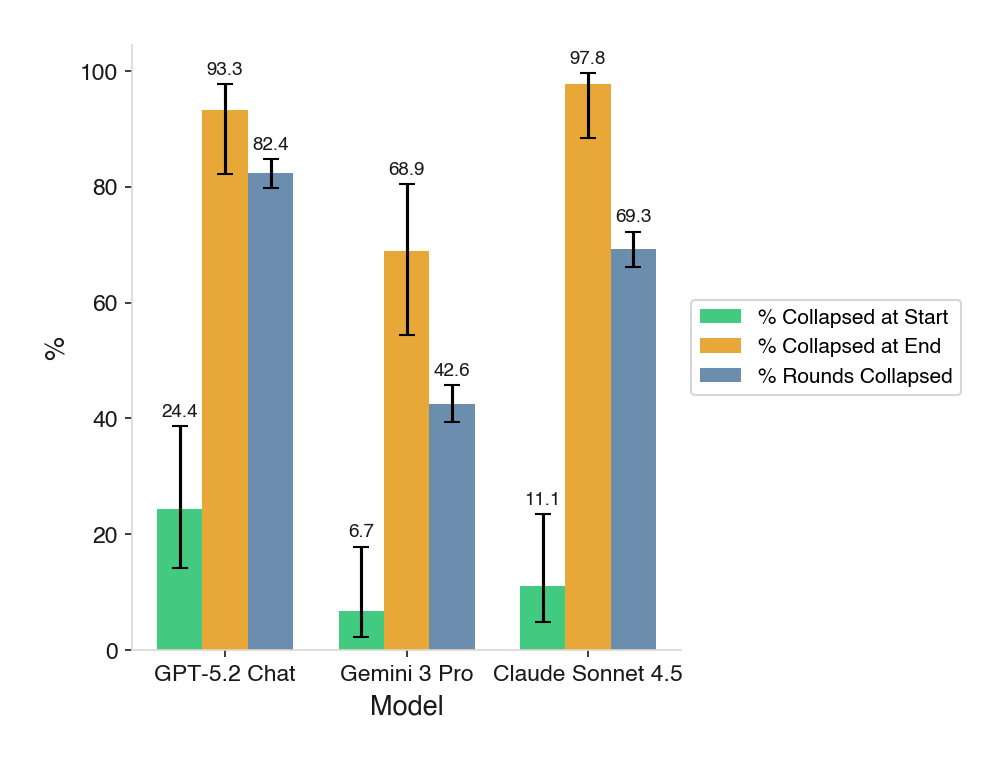}{Editorial questions.}{fig:image15}
\caption{Collapse by model for the Replace One simulation. All three models exhibit collapse, showing that the effect is not specific to OpenAI models. Gemini 3 Pro has lower collapse rates, though the confidence intervals overlap in most cases.}
\label{fig:model-collapse-row}
\end{plotblock}

\subsubsection{Entity Questions}
We next review the individual entity-question metrics.

The three models return very different numbers of entities. Claude and Gemini tend to be more comprehensive than GPT-5.2 Chat, including more entities. However, the mean number of unique entities per round decreases with all models. The mean number of unique words per round also decreases with all models. The mean semantic similarity of the responses increases for all models, though more slowly for Gemini 3 Pro. The entity-ranking similarity of the responses increases at a similar rate across all models.

\begin{plotblock}
\plotpanel{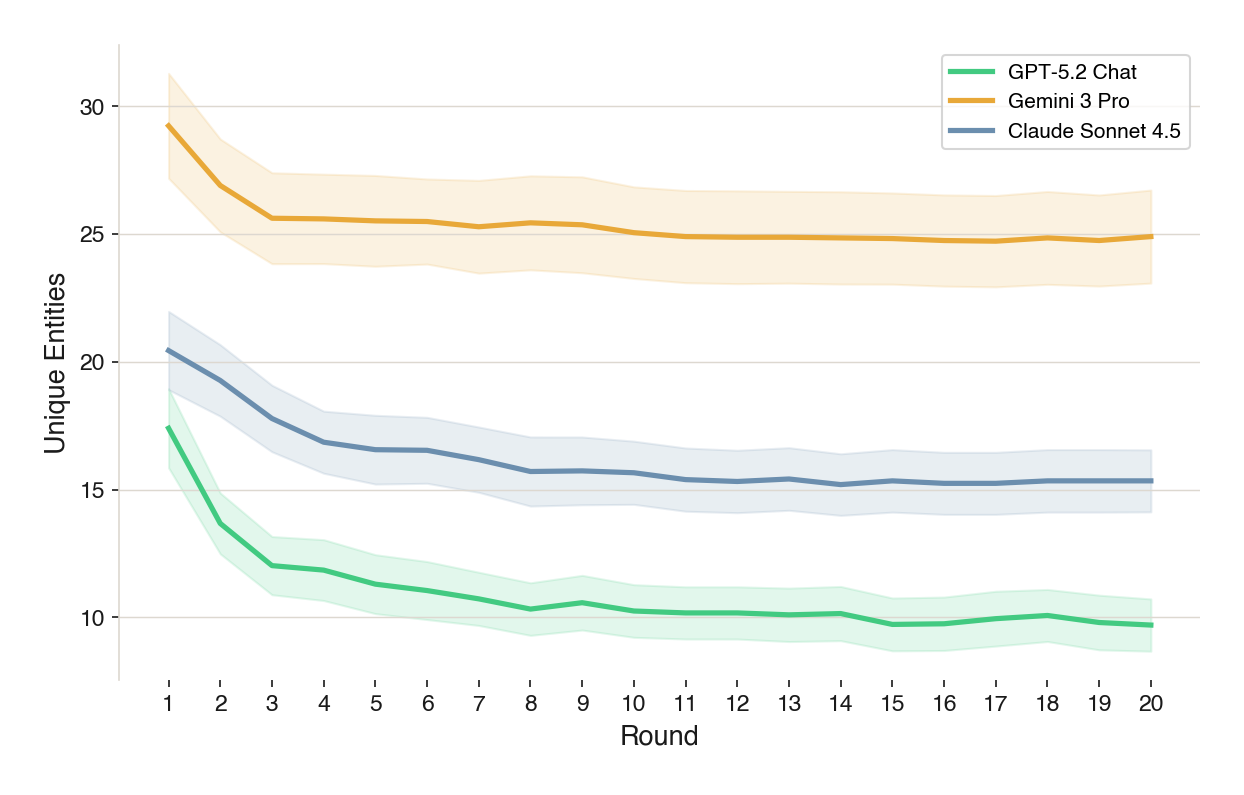}{Unique entities by round.}{fig:image41}
\hfill
\plotpanel{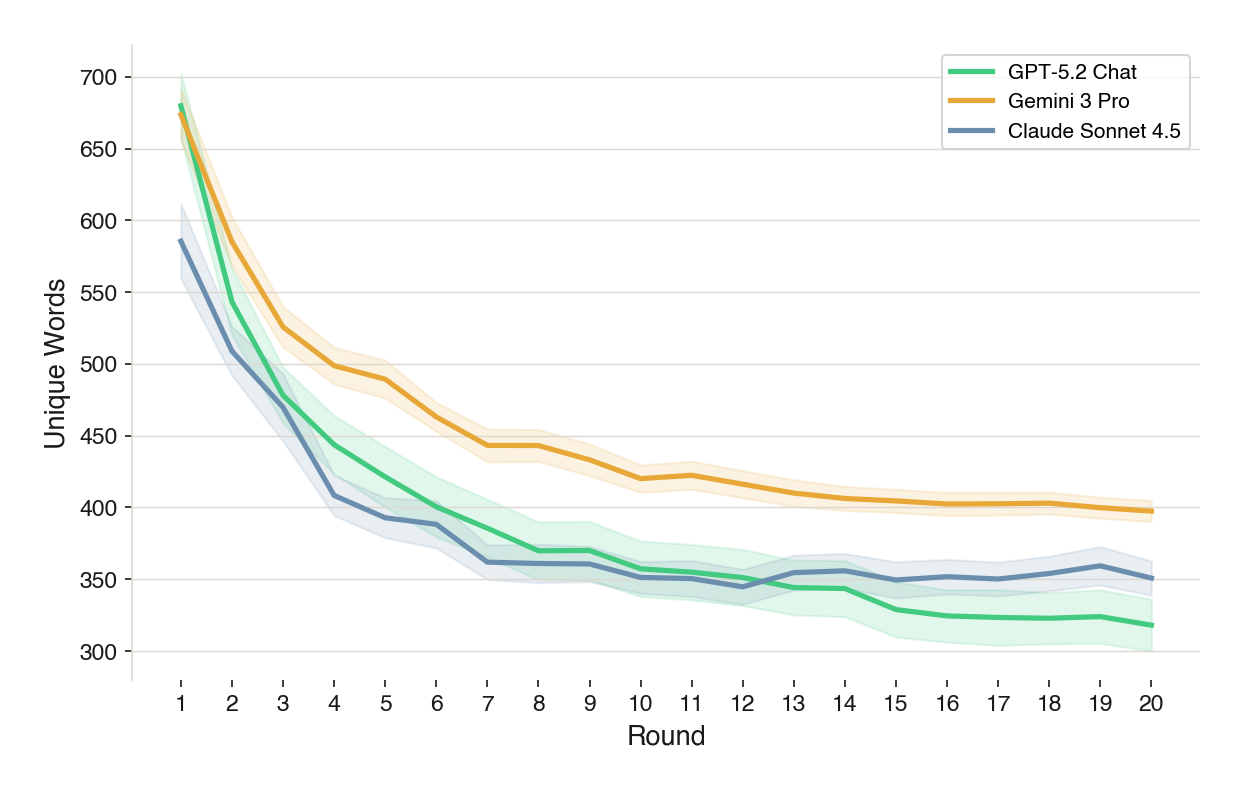}{Unique words by round.}{fig:image75}

\medskip
\plotpanel{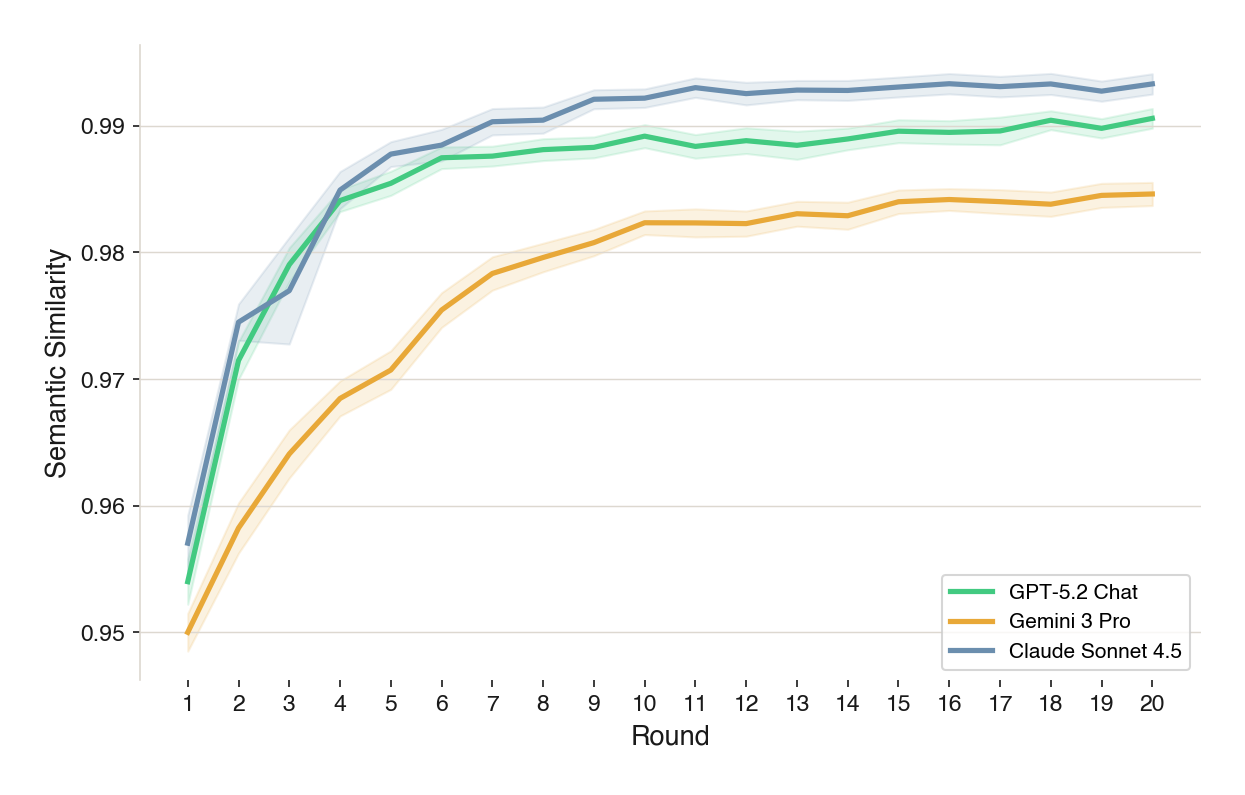}{Semantic similarity by round.}{fig:image44}
\hfill
\plotpanel{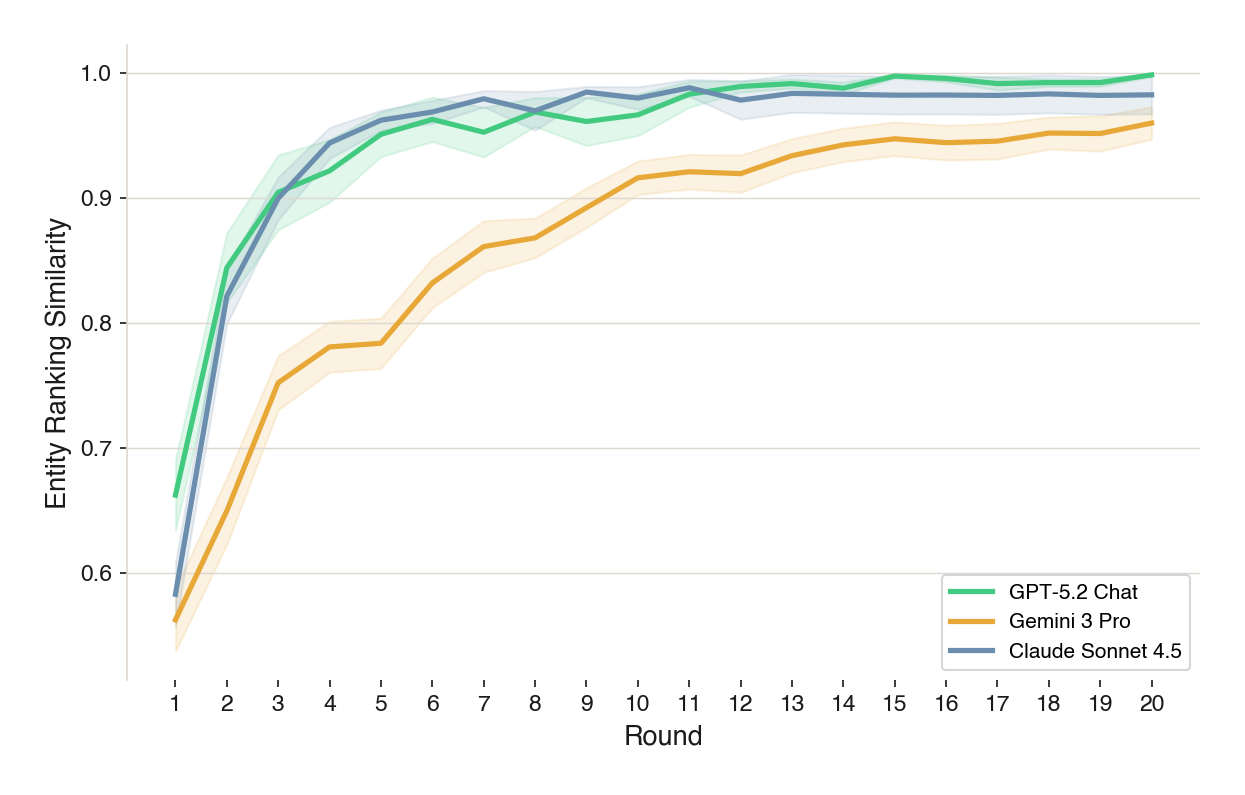}{Entity ranking similarity by round.}{fig:image42}
\caption{Core entity-question metrics by model for Replace One. Claude and Gemini initially include more entities than GPT-5.2 Chat, but unique entities and words decrease for every model. Semantic similarity increases more slowly for Gemini 3 Pro, while entity-ranking similarity increases at a similar rate across models.}
\label{fig:model-entity-metrics-grid}
\end{plotblock}

The same-answer percentages also increase over the course of the simulation, though more slowly for Gemini 3 Pro and Claude Sonnet 4.5. Self-authored references are disproportionately cited by all models.

\begin{plotblock}
\plotpanel{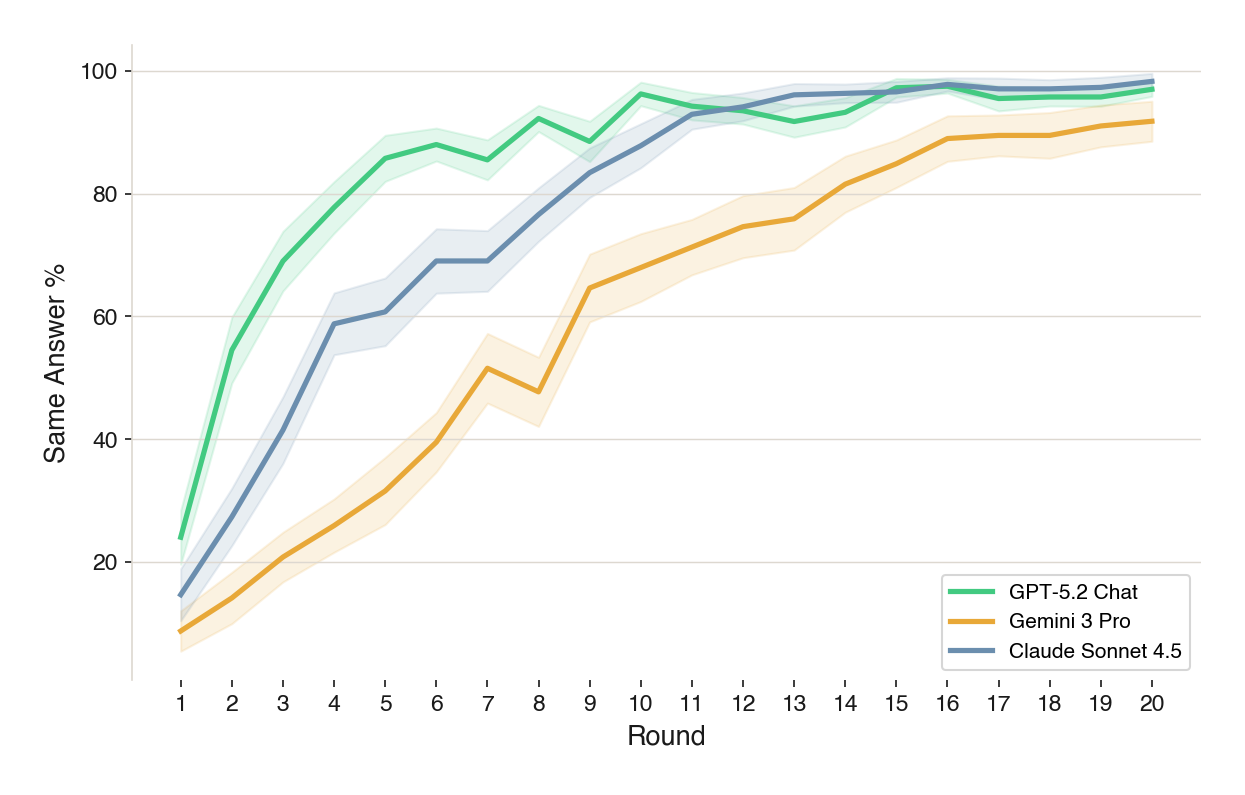}{Same-answer rate by round.}{fig:image66}
\hfill
\plotpanel{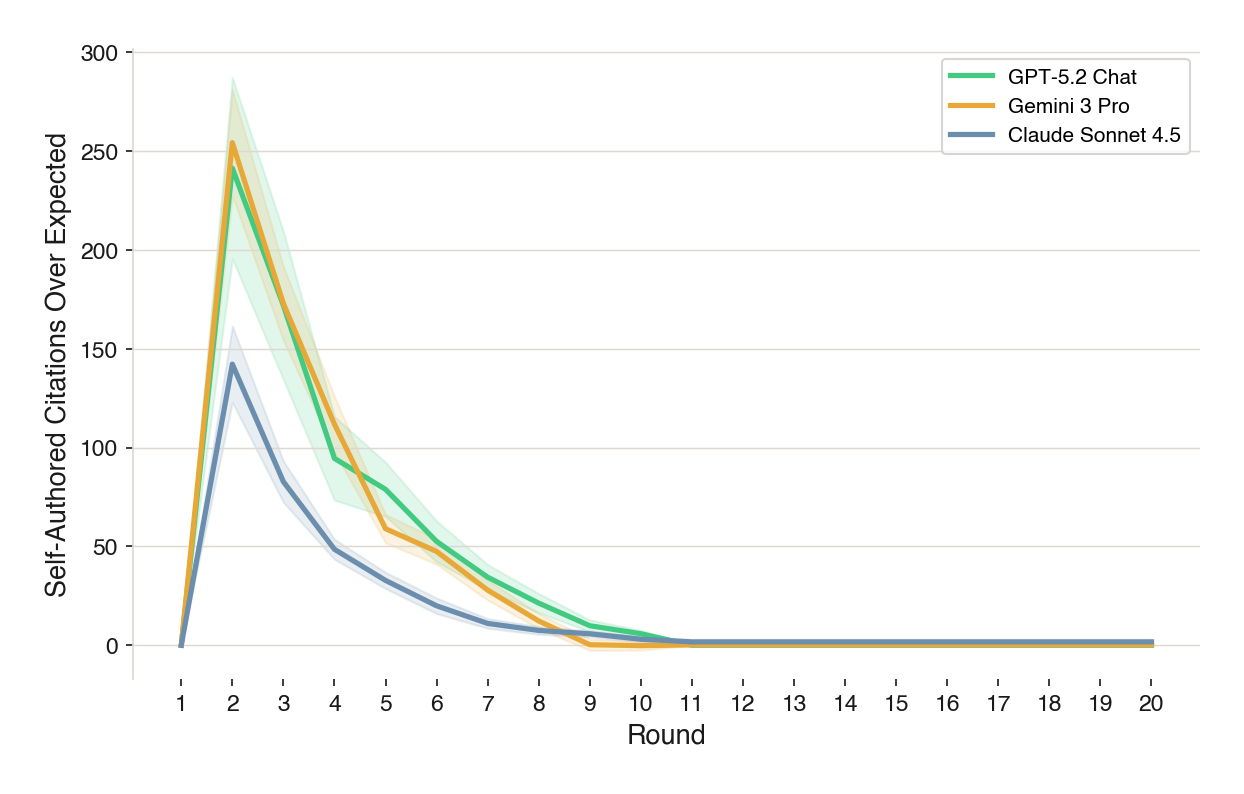}{Self-authored citations over expected.}{fig:image32}
\caption{Same-answer rate and self-authored citations by model for entity questions. Same-answer rates increase for every model, though more slowly for Gemini 3 Pro and Claude Sonnet 4.5, and every model disproportionately cites self-authored references.}
\label{fig:model-entity-answer-citation-row}
\end{plotblock}

\subsubsection{Editorial Questions}
For editorial questions, the number of unique words decreases with all models. The semantic similarities increase with all models, though more slowly for Gemini 3 Pro. The same-answer percentages also increase over the course of the simulation, though more slowly for Gemini 3 Pro. Self-authored references are disproportionately cited by all models.

\begin{plotblock}
\plotpanel{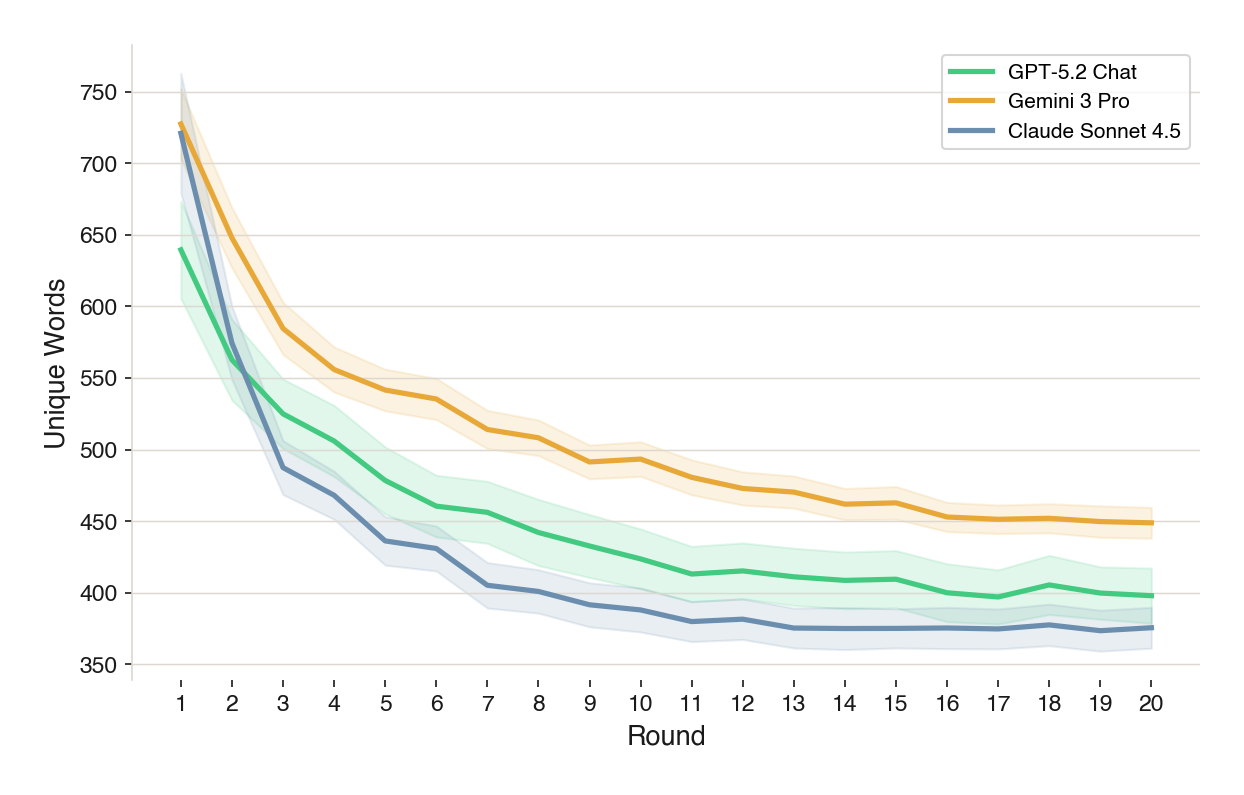}{Unique words by round.}{fig:image25}
\hfill
\plotpanel{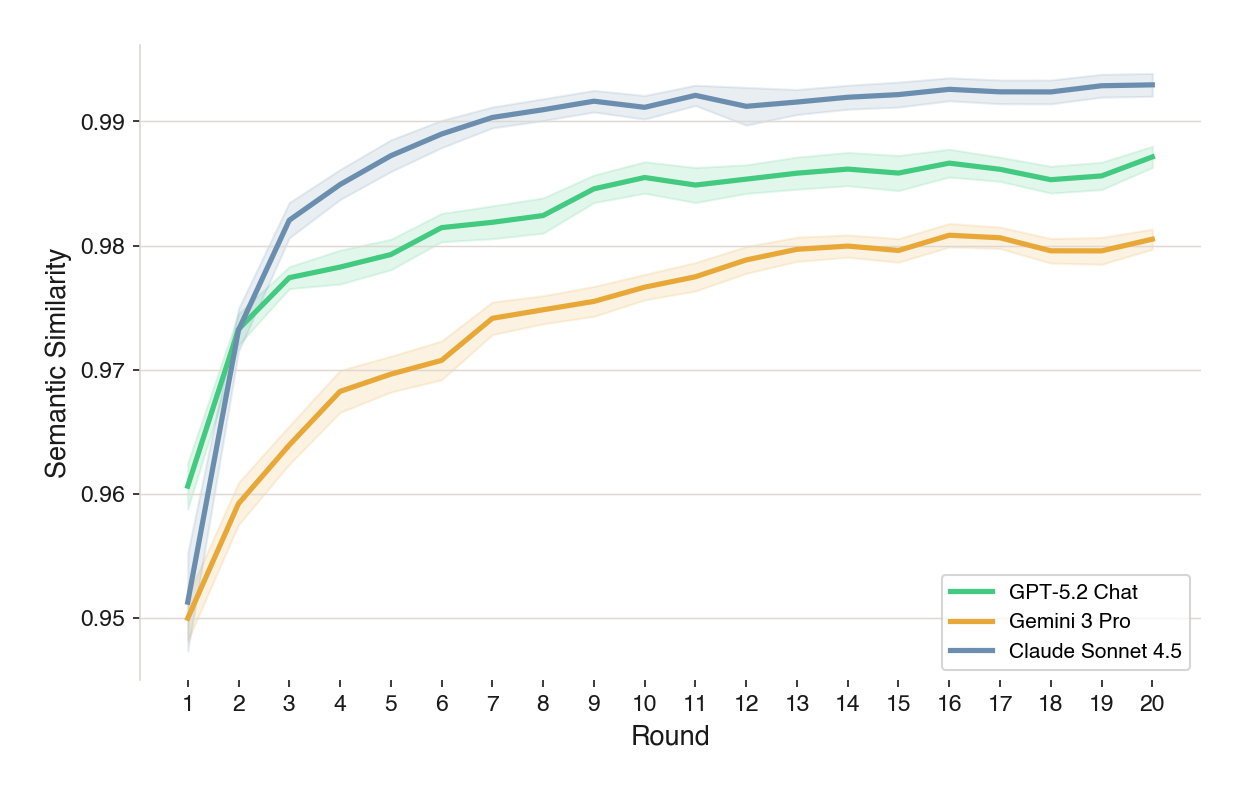}{Semantic similarity by round.}{fig:image70}

\medskip
\plotpanel{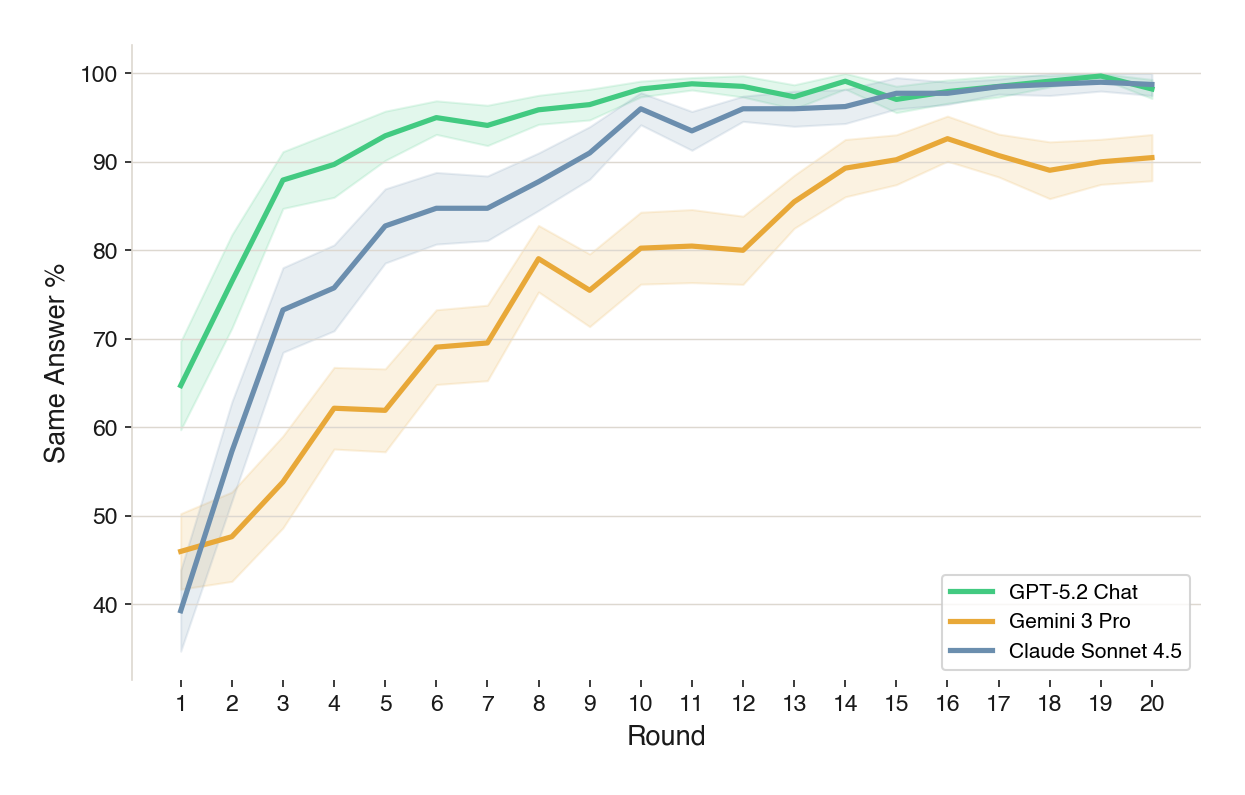}{Same-answer rate by round.}{fig:image23}
\hfill
\plotpanel{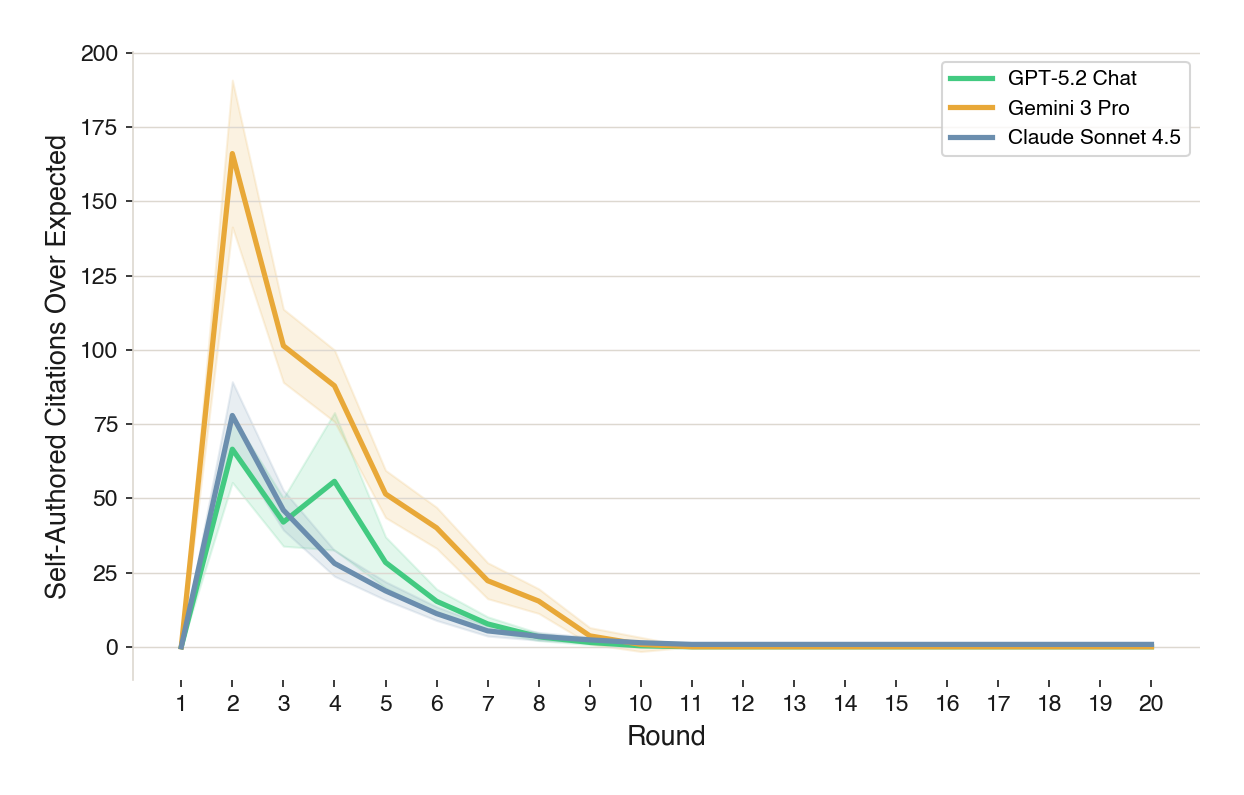}{Self-authored citations over expected.}{fig:image13}
\caption{Editorial-question metrics by model for Replace One. Unique words decrease for every model, while semantic similarity and same-answer rates increase, more slowly for Gemini 3 Pro. Every model disproportionately cites self-authored references.}
\label{fig:model-editorial-metrics-grid}
\end{plotblock}

\subsection{Larger-Scale Experiments}
\label{sec:large-scale-experiments}
Finally, we conduct larger-scale experiments with the Search simulation using GPT-5.2 to ensure that our results generalize to many questions. Note that these experiments use GPT-5.2 rather than GPT-5.2 Chat; both variants collapse at high rates (Table~\ref{tab:3}), though we did not run a controlled comparison.

We run the Search simulation on 742 additional entity questions that we did not include in the experiments thus far. The results are consistent with the experiments presented previously.

\begin{plotblock}
\plotpanel{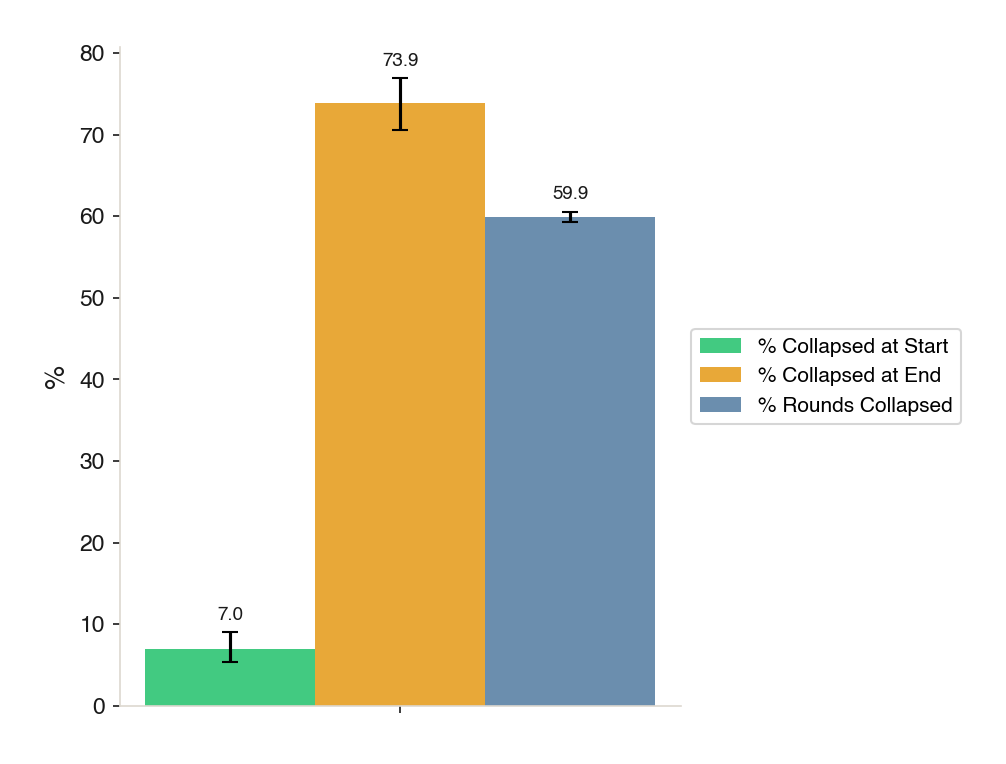}{Collapse rates.}{fig:image56}
\hfill
\plotpanel{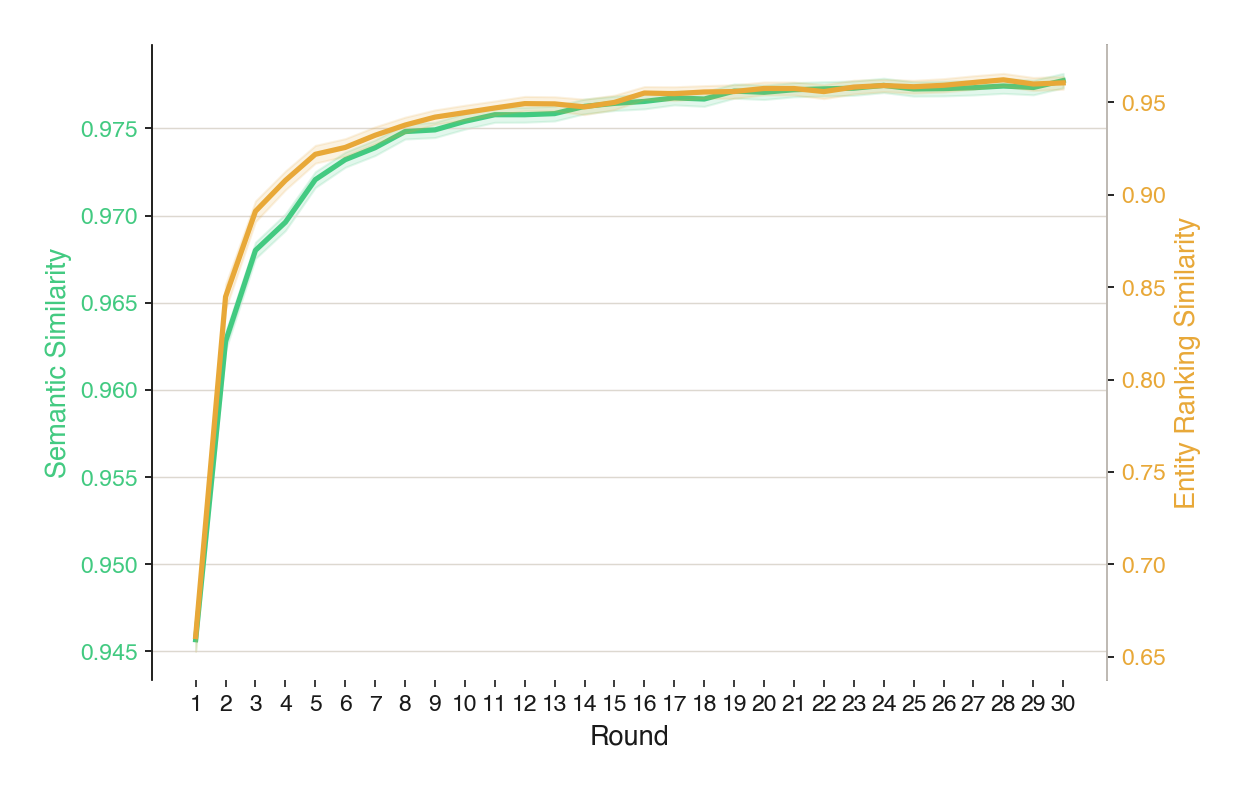}{Semantic and entity-ranking similarity.}{fig:image59}

\medskip
\plotpanel{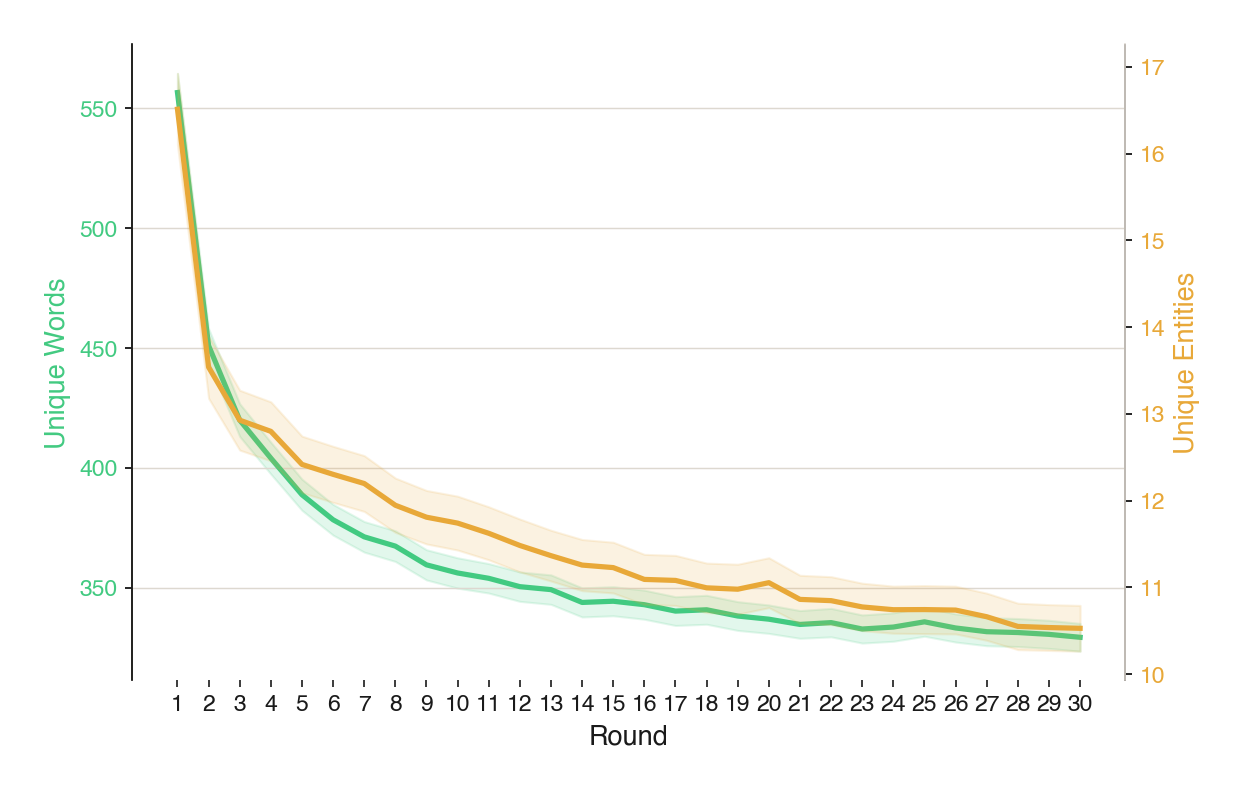}{Unique words and entities.}{fig:image30}
\hfill
\plotpanel{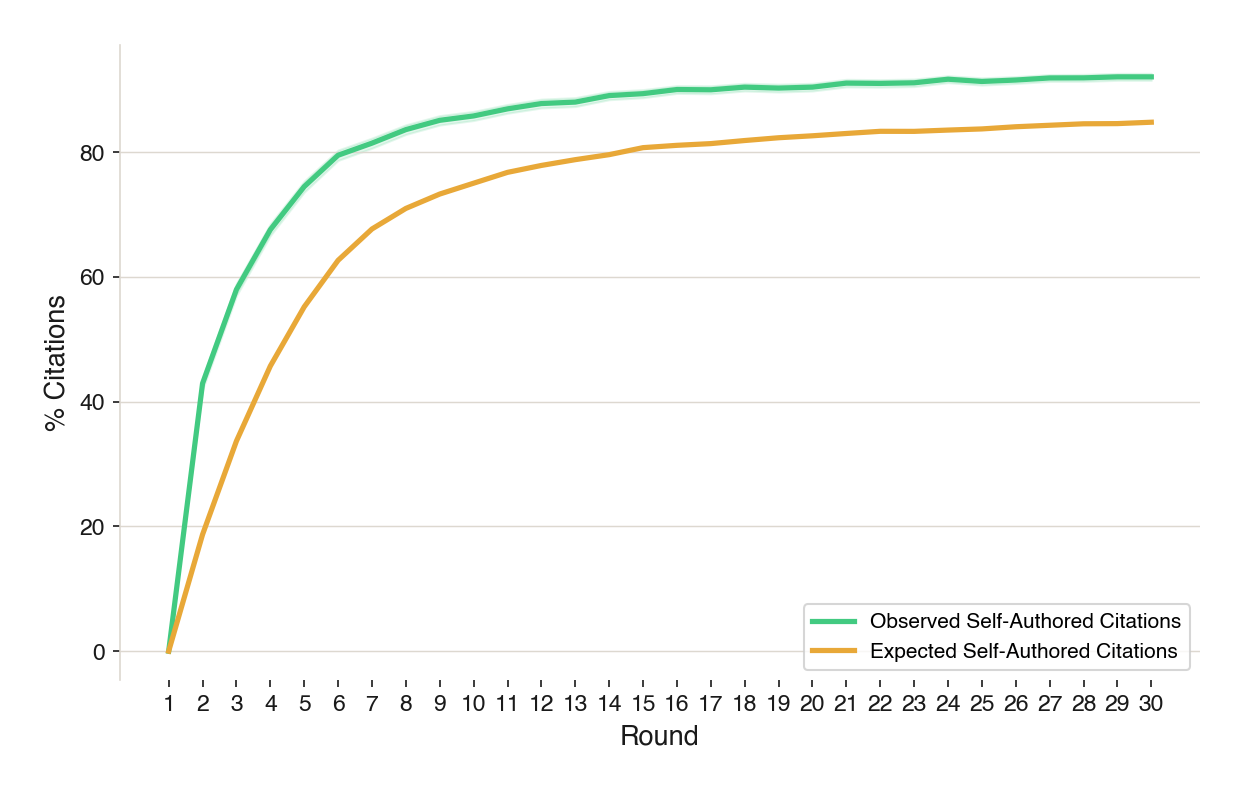}{Observed and expected self-authored citations.}{fig:image37}
\caption{Large-scale entity-question results for Search using GPT-5.2.}
\label{fig:large-scale-entity-grid}
\end{plotblock}

\begin{plotblock}
\plotpanel{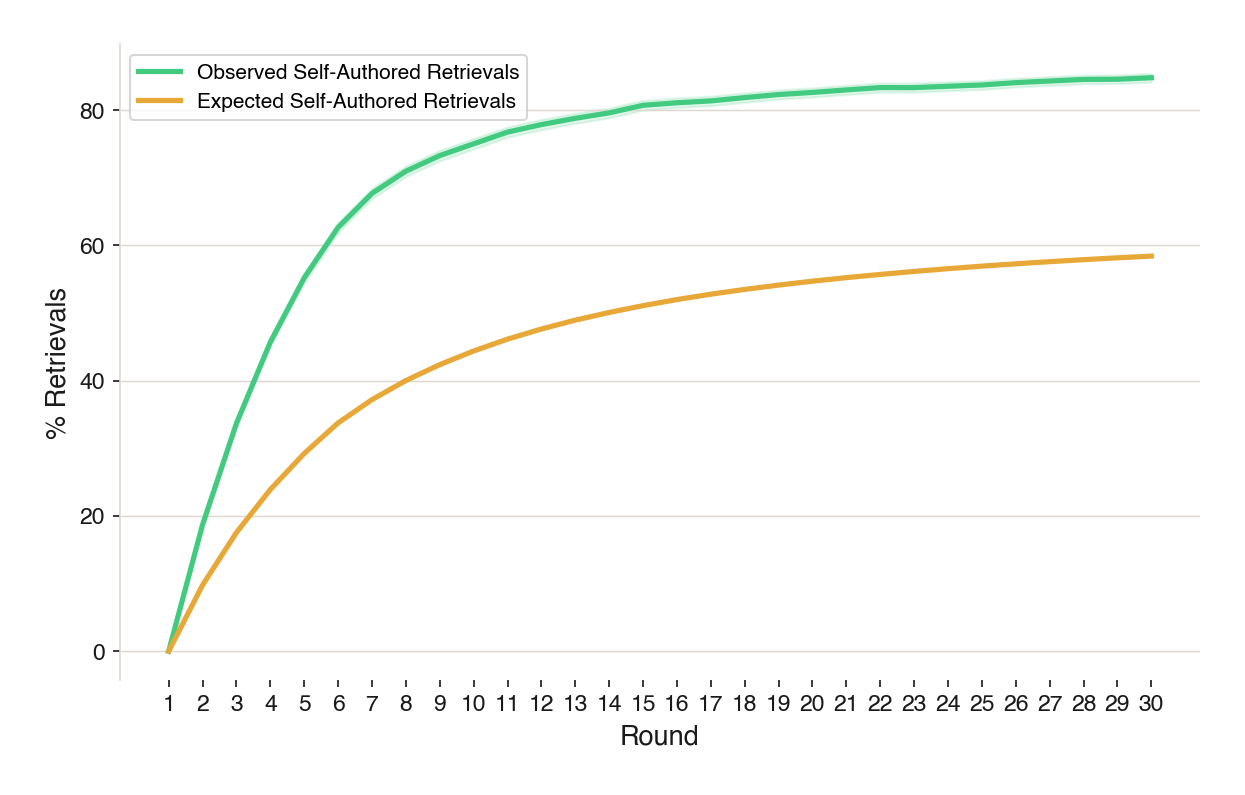}{Observed and expected self-authored retrievals.}{fig:image34}
\hfill
\plotpanel{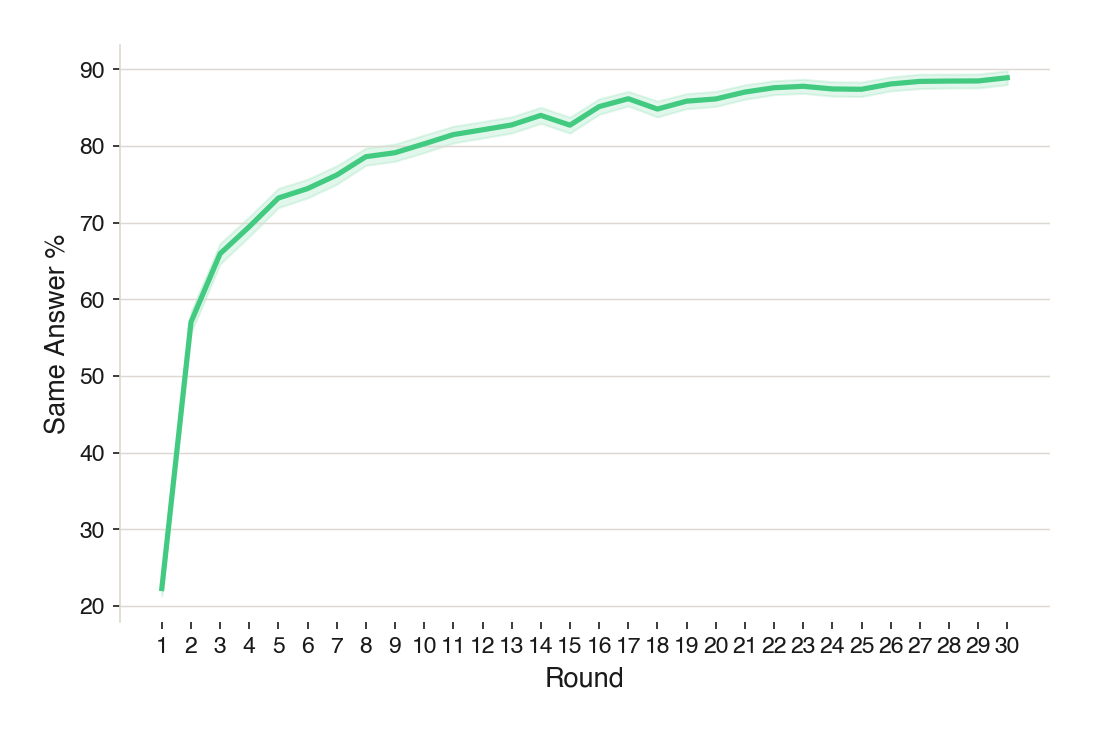}{Same-answer rate by round.}{fig:image33}
\caption{Retrieval and same-answer metrics for the large-scale entity-question Search experiment.}
\label{fig:large-scale-entity-retrieval-answer-row}
\end{plotblock}

We also run simulations for the 102 additional editorial questions. Again, the results are consistent with the experiments presented previously.

\begin{plotblock}
\plotpanel{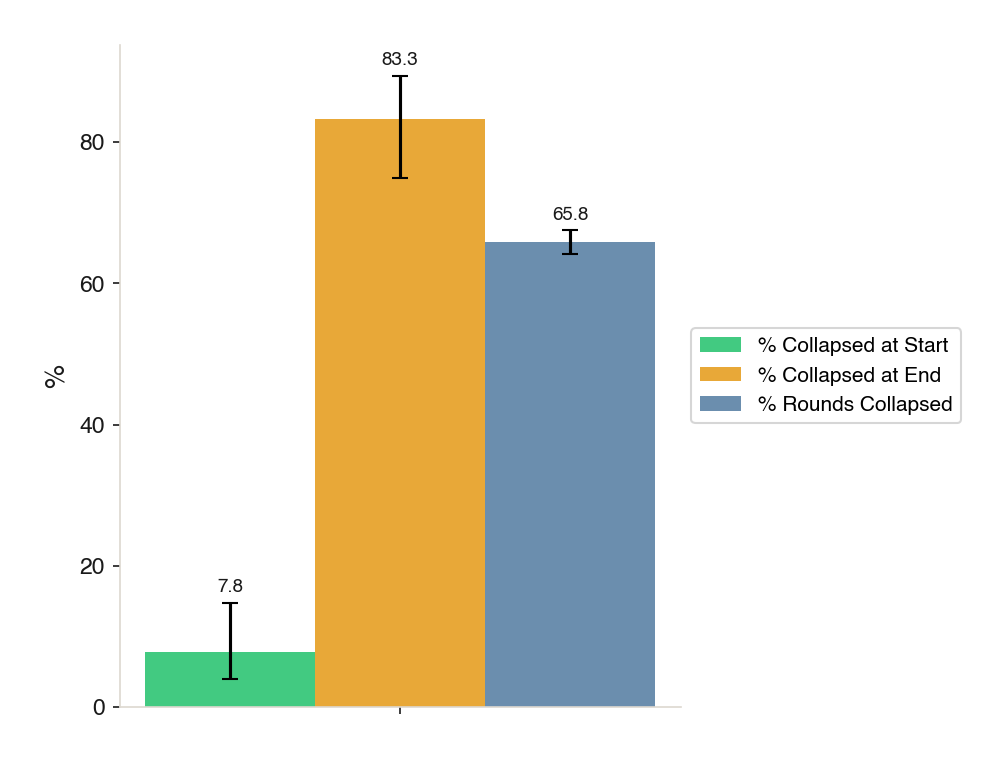}{Collapse rates.}{fig:image49}
\hfill
\plotpanel{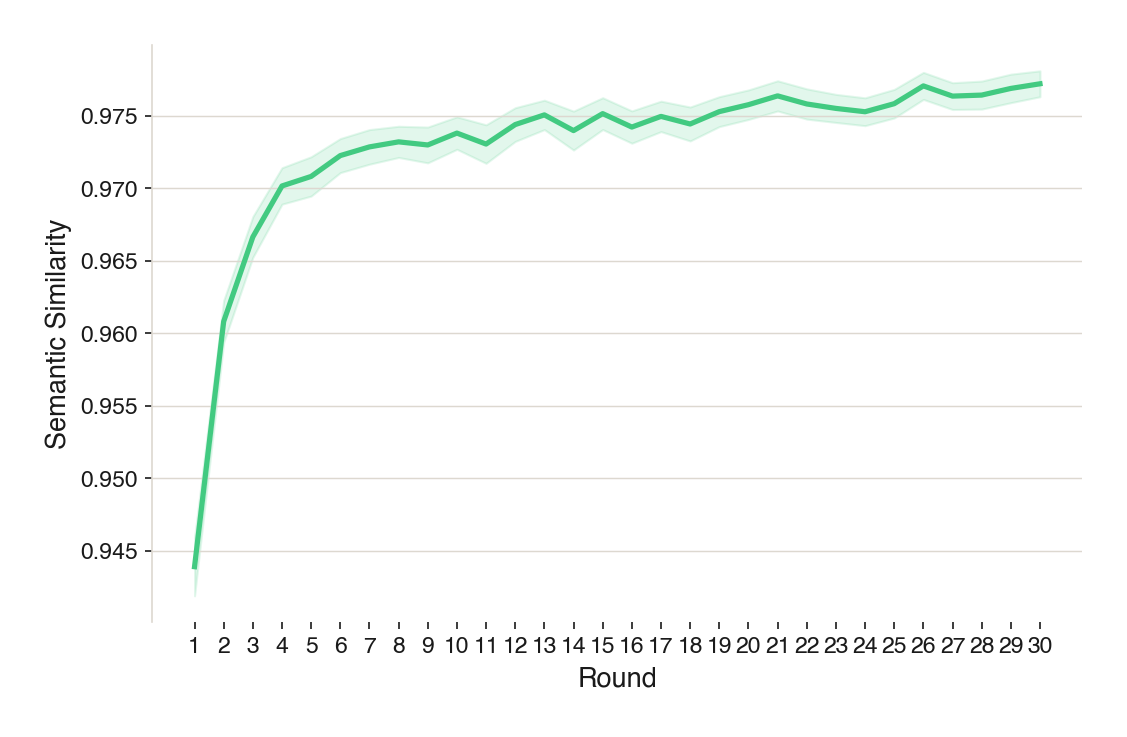}{Semantic similarity.}{fig:image38}

\medskip
\plotpanel{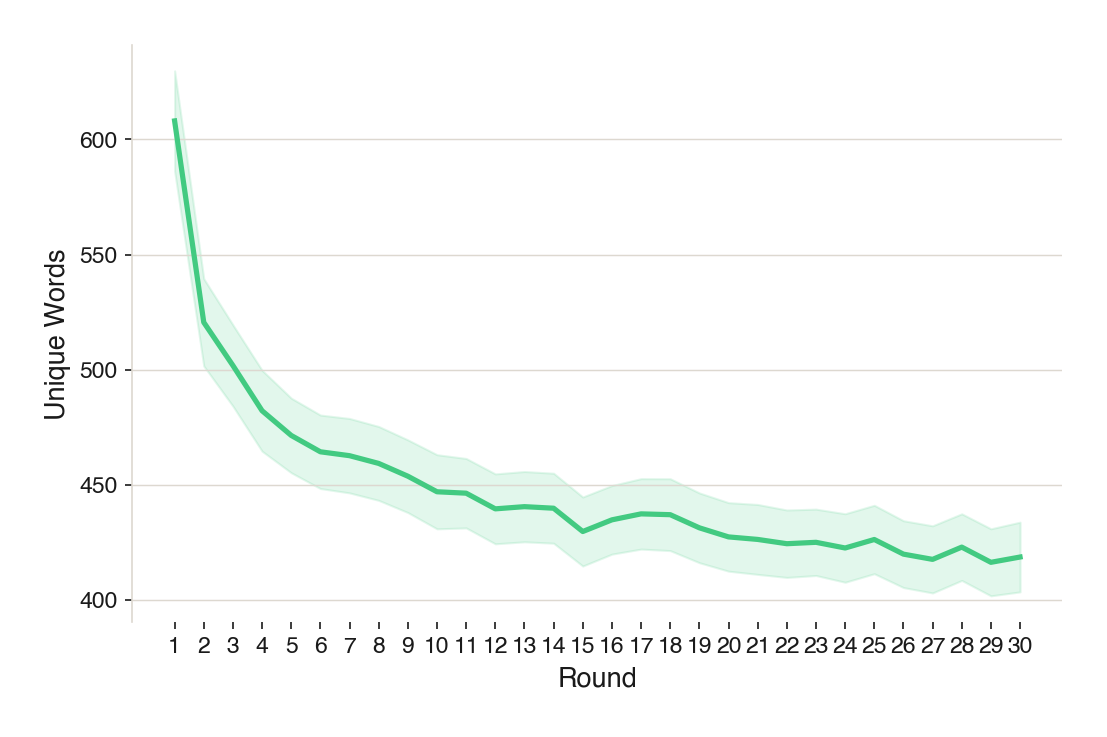}{Unique words.}{fig:image29}
\hfill
\plotpanel{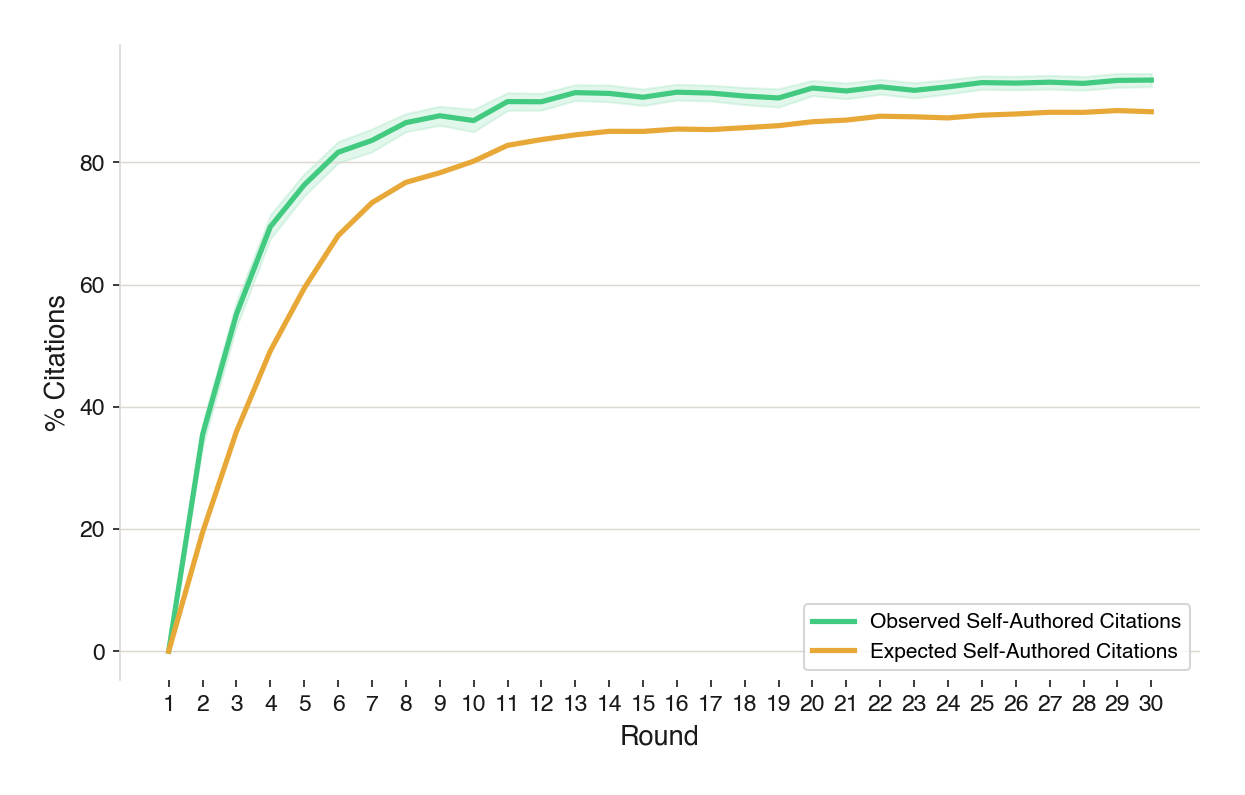}{Observed and expected self-authored citations.}{fig:image69}
\caption{Large-scale editorial-question results for Search using GPT-5.2.}
\label{fig:large-scale-editorial-grid}
\end{plotblock}

\begin{plotblock}
\plotpanel{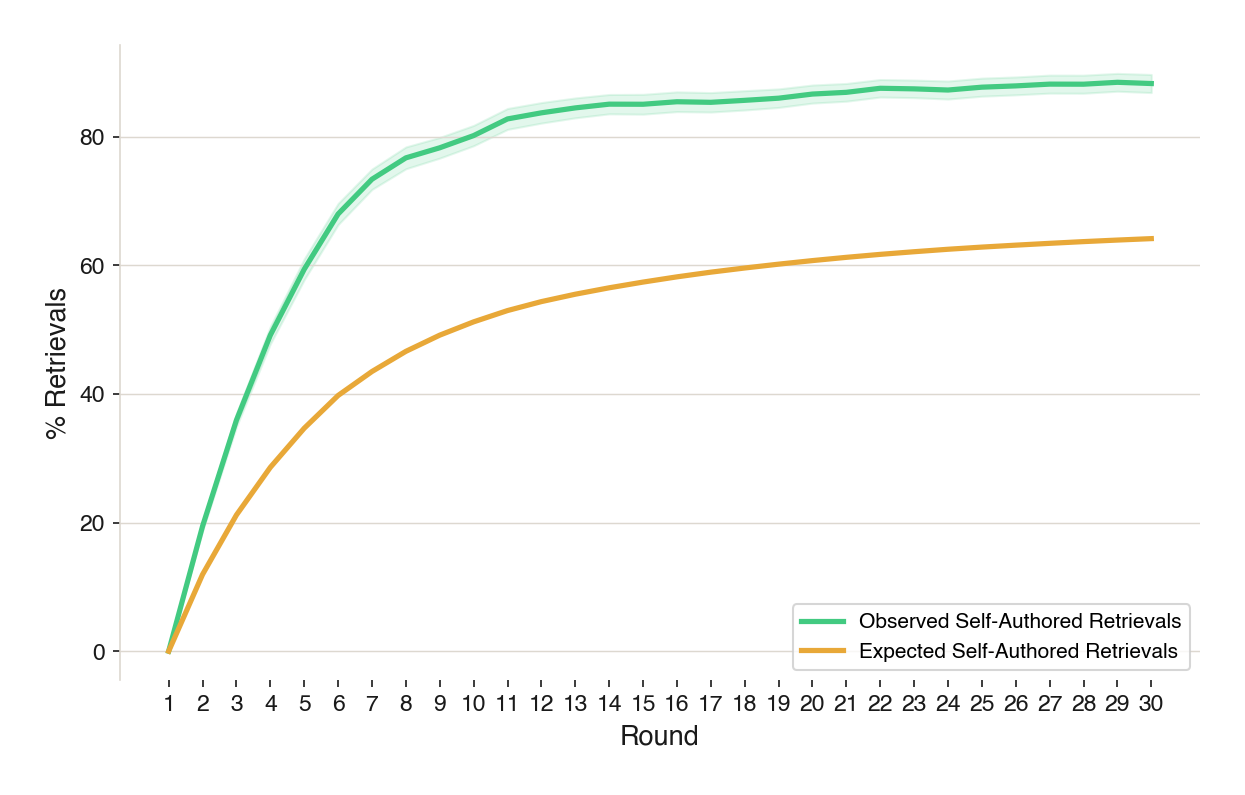}{Observed and expected self-authored retrievals.}{fig:image8}
\hfill
\plotpanel{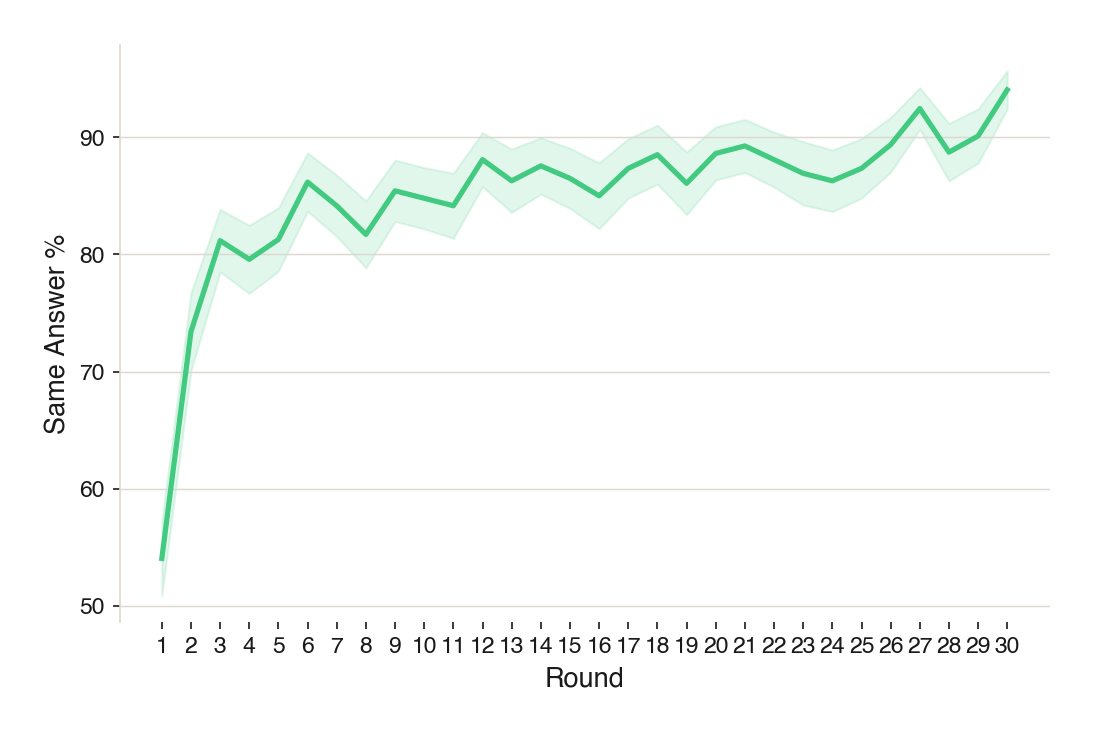}{Same-answer rate by round.}{fig:image27}
\caption{Retrieval and same-answer metrics for the large-scale editorial-question Search experiment.}
\label{fig:large-scale-editorial-retrieval-answer-row}
\end{plotblock}

\section{Why Do Self-Authored References Have Disproportionate Influence?}
\label{sec:self-authored-influence}
It is not surprising that the Replace All simulation often results in collapse, as it most closely mirrors the original model collapse paper~\citep{shumailov2024collapse}. However, in the Replace One simulation, we find that collapse begins when self-authored references constitute only a small proportion of the total reference set, even a single self-authored reference, or 10--20\% of the total references. We find that these references are disproportionately cited, suggesting they exert greater influence on the response than the original references.

Is this because the references are AI-generated, or because they are self-authored? In this section, we analyze each candidate reference in round 2 of the Replace One simulation on Entity ChatGPT, in which complete (unchunked) self-authored and original references compete directly for citations.

We find that 39.7\% of the original references in the Entity ChatGPT dataset used in the Replace One simulation experiments are AI-generated. However, only about 3\% of entity questions are collapsed at the start, before any self-authored references are added. We only observe widespread collapse when self-authored references are introduced. This suggests self-authorship, not AI generation, is the driver of collapse in our simulations. We test this more thoroughly below.

Self-authored references are cited far more often than original references, both AI-generated and human-written. We saw evidence of this in the self-authored citations over expected metric in our experiments, and we investigate it in more detail here. We use GPTZero to classify original references as human-written or AI-generated. AI-generated originals have a citation rate of 9.4\%, only modestly higher than that of human-written originals (7.4\%), while self-authored references have a citation rate of 38.9\%. We verified the same pattern holds in the Search simulation and on editorial questions.

\begin{plotblock}
\includegraphics[width=\plotwidth,height=\plotheight,keepaspectratio]{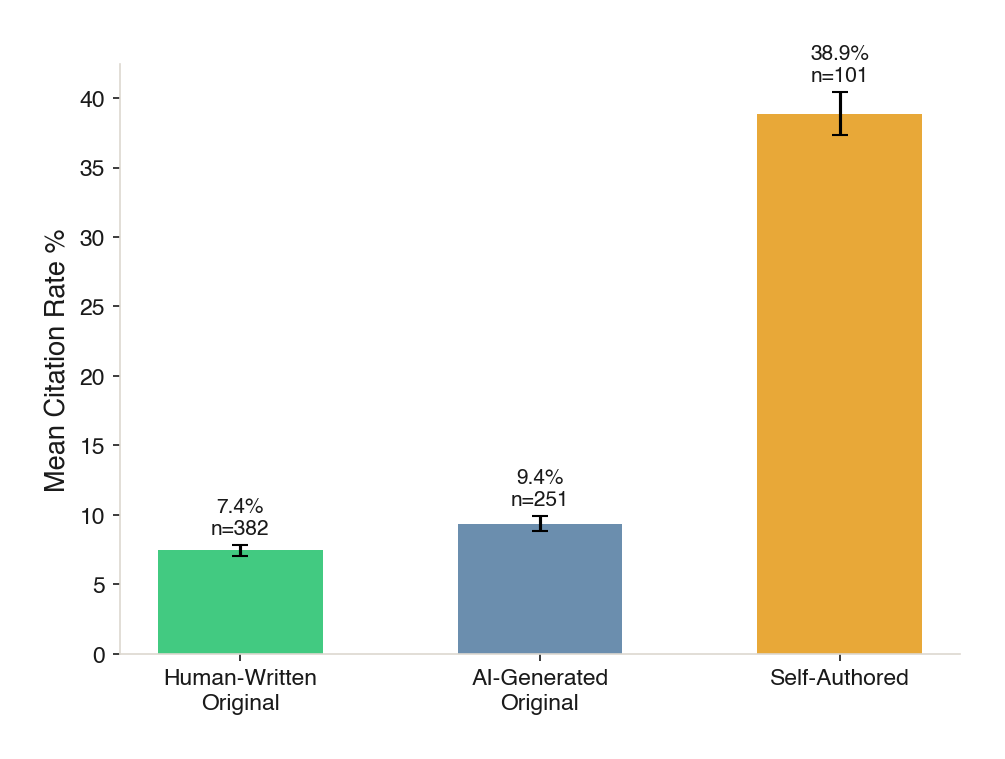}
\caption{Citation rate by reference type. AI-generated originals are cited only slightly more often than human-written originals (9.4\% vs. 7.4\%), whereas self-authored references are cited at 38.9\%.}
\label{fig:image60}
\end{plotblock}

One explanation is that self-authored references are simply higher quality. To test this, we use GPT-5.2 to independently score each reference on eight dimensions, each on a 1--5 scale: overall, relevance, accuracy, thoroughness, specificity, up to date, organization, and direct answer. Note that although LLM-as-judge is subject to bias, here we explicitly want to know what the LLM thinks, since it is the LLM choosing which references to cite.

Self-authored references do score higher on most dimensions, with the largest gaps on direct answer (+1.23), relevance (+0.78), thoroughness (+0.74), organization (+0.65), and overall (+0.49). The two types are tied in accuracy, self-authored references score marginally higher on the up-to-date dimension, and original references have a slight edge in specificity.

\begin{plotblock}
\includegraphics[width=\plotwidth,height=\plotheight,keepaspectratio]{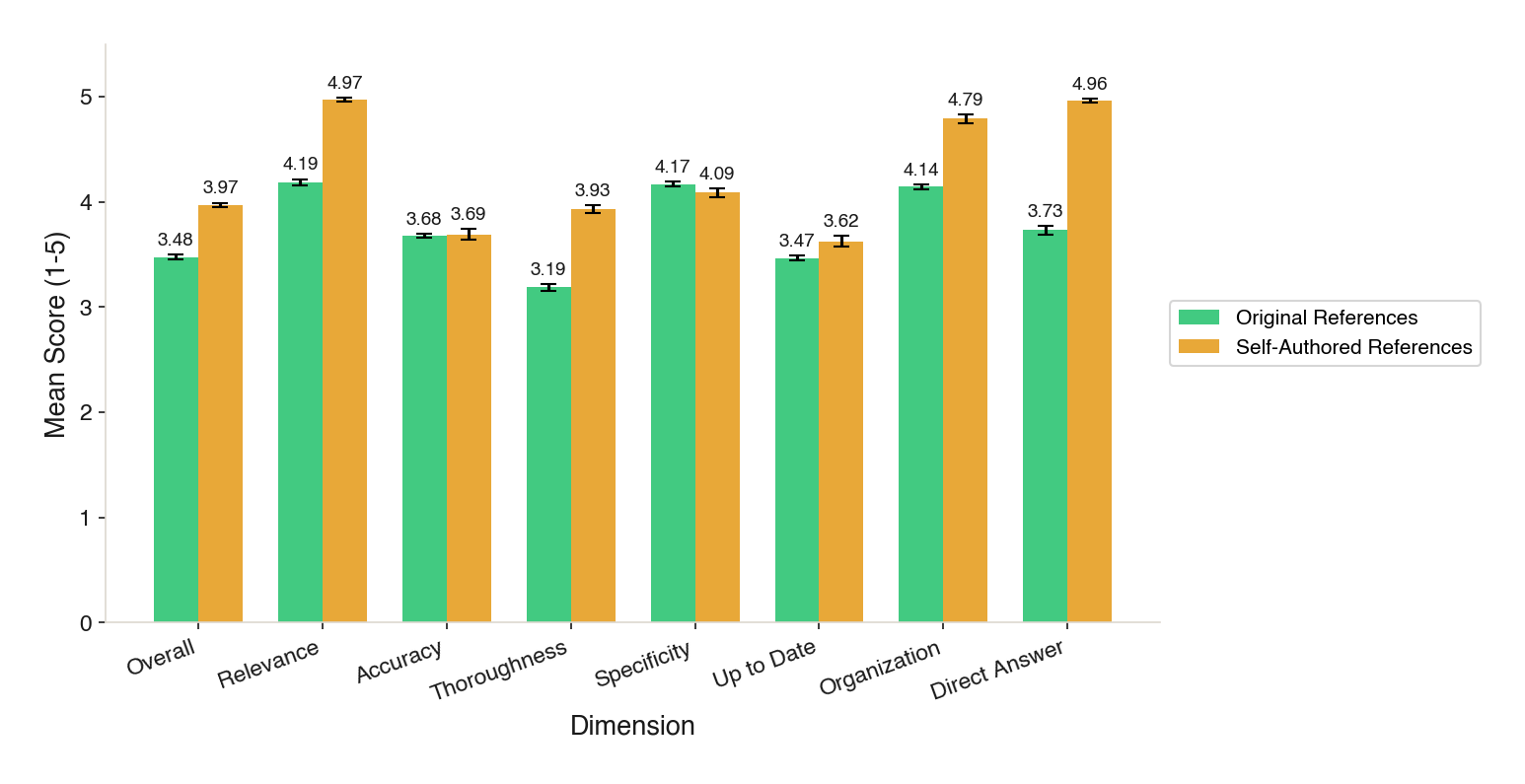}
\caption{Mean reference-quality score by dimension. Self-authored references score higher on most dimensions, while the two reference types are tied in accuracy.}
\label{fig:image67}
\end{plotblock}

To control for this quality advantage, we restrict to references the judge scored 5/5 on the three dimensions where self-authored references most consistently scored highest: direct answer, organization, and relevance. Even within this subset ($n=193$: 75 self-authored and 118 original), self-authored references are cited at 38.2\% compared to 13.3\% for original references, a 2.9$\times$ gap.

\begin{plotblock}
\includegraphics[width=\plotwidth,height=\plotheight,keepaspectratio]{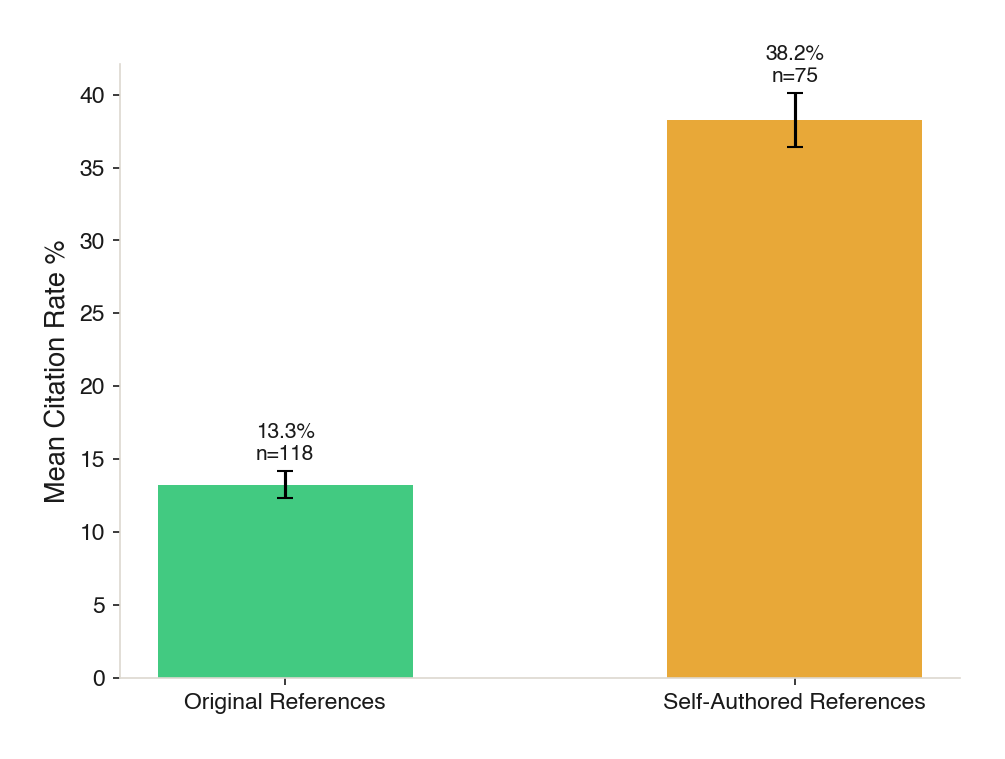}
\caption{Citation rates in the quality-controlled subset. Self-authored references retain a 2.9$\times$ citation-rate advantage.}
\label{fig:image21}
\end{plotblock}

As a stronger test, we fit a linear regression model to predict the citation rate using indicators of whether a reference is self-authored or an original AI-generated reference, as well as the eight quality scores. Being self-authored increases the citation rate by 0.26 (95\% bootstrap CI: [+0.23, +0.29], $p < 0.001$), whereas being an AI-generated original has no effect ($-$0.01, not significant). After controlling for all eight quality dimensions, self-authorship alone adds about 26 percentage points to a reference's citation rate, while AI generation adds none.

Self-authored references are not cited more often only because they are higher quality or AI-generated. While it is possible there is another important attribute of the self-authored references we did not control for that explains the gap, the results suggest self-bias. \citet{panickssery2024own} showed that LLMs prefer their own generations. From these experiments, we cannot tell whether the model prefers these references because their content matches its reasoning (how it would answer given similar references) or because of their style. In future work, we could add a reference that states a different answer, or test whether one model over-cites another's generated references.

This does not mean that the AI-generated references already being cited by LLMs are harmless. The collapse we observe in our simulations is driven by the self-authored references we introduce, but we do not test iteratively adding other types of AI-generated content, and it is possible that doing so could also lead to collapse.

In the Search simulation experiments, we also find that self-authored references are disproportionately retrieved. The retrieval mechanism itself may favor them due to higher semantic similarity with the query. We leave a deeper investigation of these mechanisms to future work.

\section{What Types of Prompts Collapse?}
We aggregate all the Search simulation experiments using the Entity ChatGPT dataset and examine several factors to see which are correlated with collapse. Specifically, we examine statistics of the generations in the initial round, using only the original references. The factors most correlated with collapse are the number of unique entities and the mean length of responses.

\begin{table}[htbp]
\centering
\caption{Correlations between initial-response statistics and eventual collapse in entity-question Search simulations.}
\label{tab:4}
\small
\begin{tabularx}{\linewidth}{YZZ}
\toprule
\textbf{Name} & \textbf{Pearson Correlation} & \textbf{$p$} \\
\midrule
unique entities & $-$0.342 & $1.32 \times 10^{-24}$ \\
length of responses & $-$0.245 & $5.31 \times 10^{-13}$ \\
unique words & $-$0.223 & $6.21 \times 10^{-11}$ \\
entity ranking similarity & 0.164 & $1.61 \times 10^{-6}$ \\
same-answer \% & 0.151 & $1.01 \times 10^{-5}$ \\
\bottomrule
\end{tabularx}
\end{table}

These correlations are statistically significant but modest, suggesting other factors also contribute to collapse. However, we see that prompts with responses that are longer and contain more unique entities, that is, more varied initially, are more resistant to collapse.

\section{Which Entities Drop Out?}
Next, we examine which entities tend to drop out of responses by the end of the simulation.

As expected, the probability of an entity dropping out decreases as its initial visibility increases. Low-visibility entities that appear in few responses initially drop out very frequently, whereas high-visibility entities rarely do, though even 100\%-visibility entities drop out about 7--10\% of the time.

\begin{plotblock}
\plotpanel{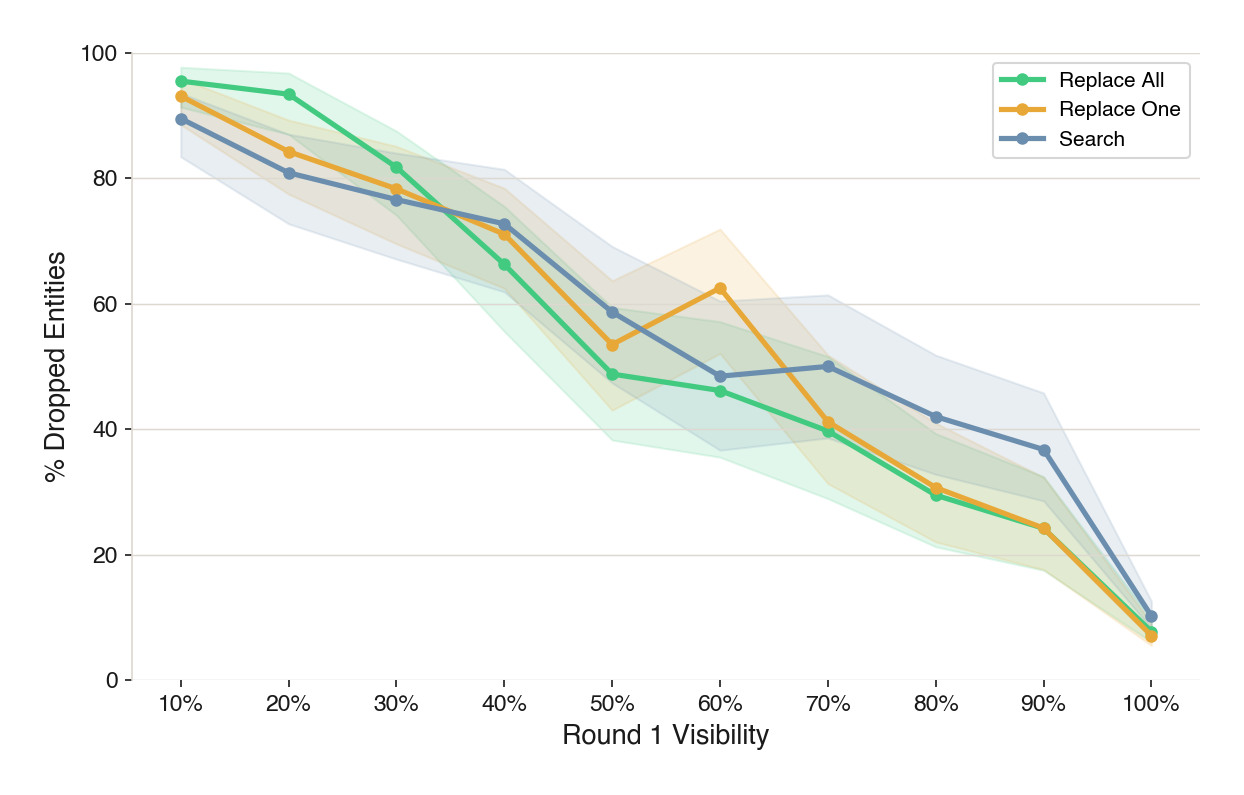}{Dropout by initial visibility.}{fig:image52}
\hfill
\plotpanel{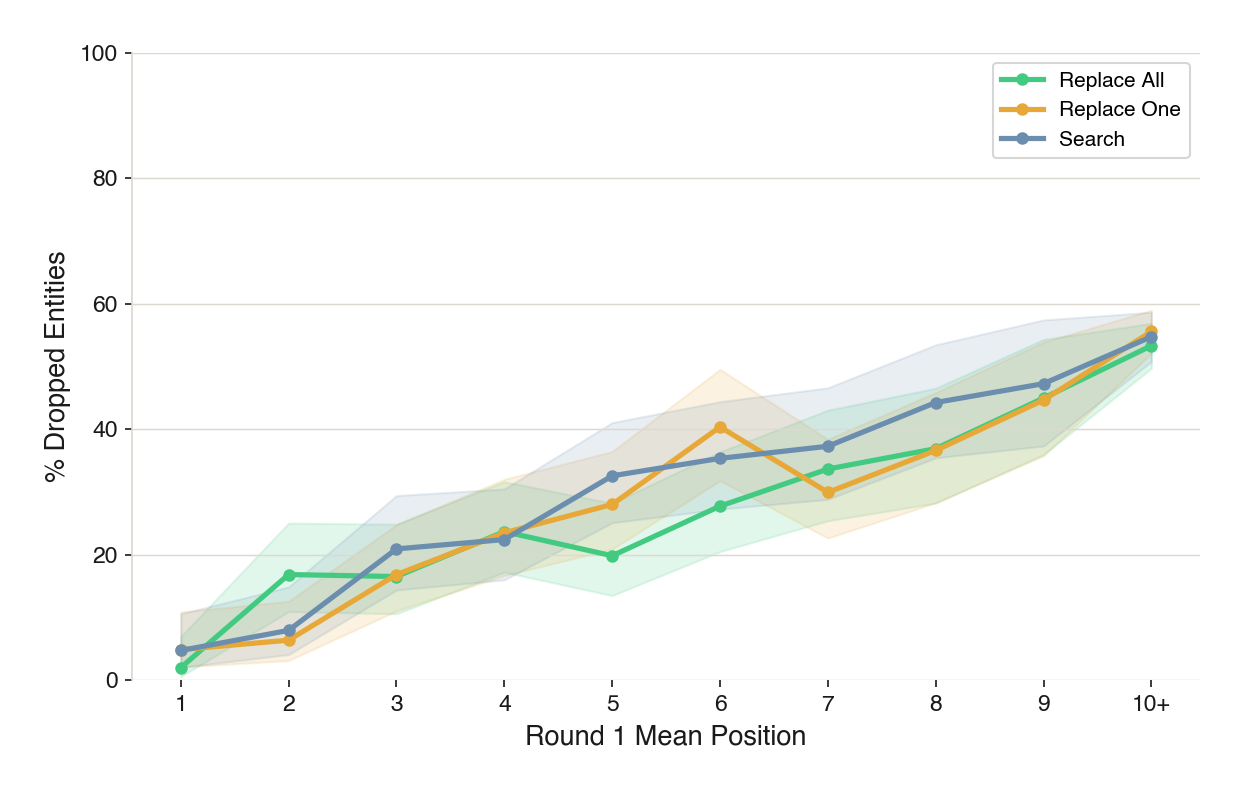}{Dropout by initial position.}{fig:image73}
\caption{Entity dropout rates by initial visibility and position. Dropout decreases as initial visibility increases and as position improves.}
\label{fig:entity-dropout-row}
\end{plotblock}

We also find a correlation between visibility and position. High-visibility entities tend to appear near the beginning of the response. We provide supporting data in prior work~\citep{druck2026entity}. Better-positioned entities tend to drop out less frequently.

\section{Does RAG Mirror the Input Entity Distribution?}
We might expect RAG to mirror the distribution of answers in the references, but our earlier results suggested otherwise (Section~\ref{sec:rag-collapse}). In this section, we provide direct evidence that the distribution of entity mentions in responses does not necessarily reflect the distribution of entities in the references. For each entity, we plot the number of responses that mention it (y-axis) versus the number of references that mention it (x-axis). We restrict to questions with ten references.

Across all models, we generally see more responses mentioning an entity than references mentioning the entity, saturating around five input references. Gemini 3 Pro includes low-mention-count entities in the responses more often.

\begin{plotblock}
\includegraphics[width=\plotwidth,height=\plotheight,keepaspectratio]{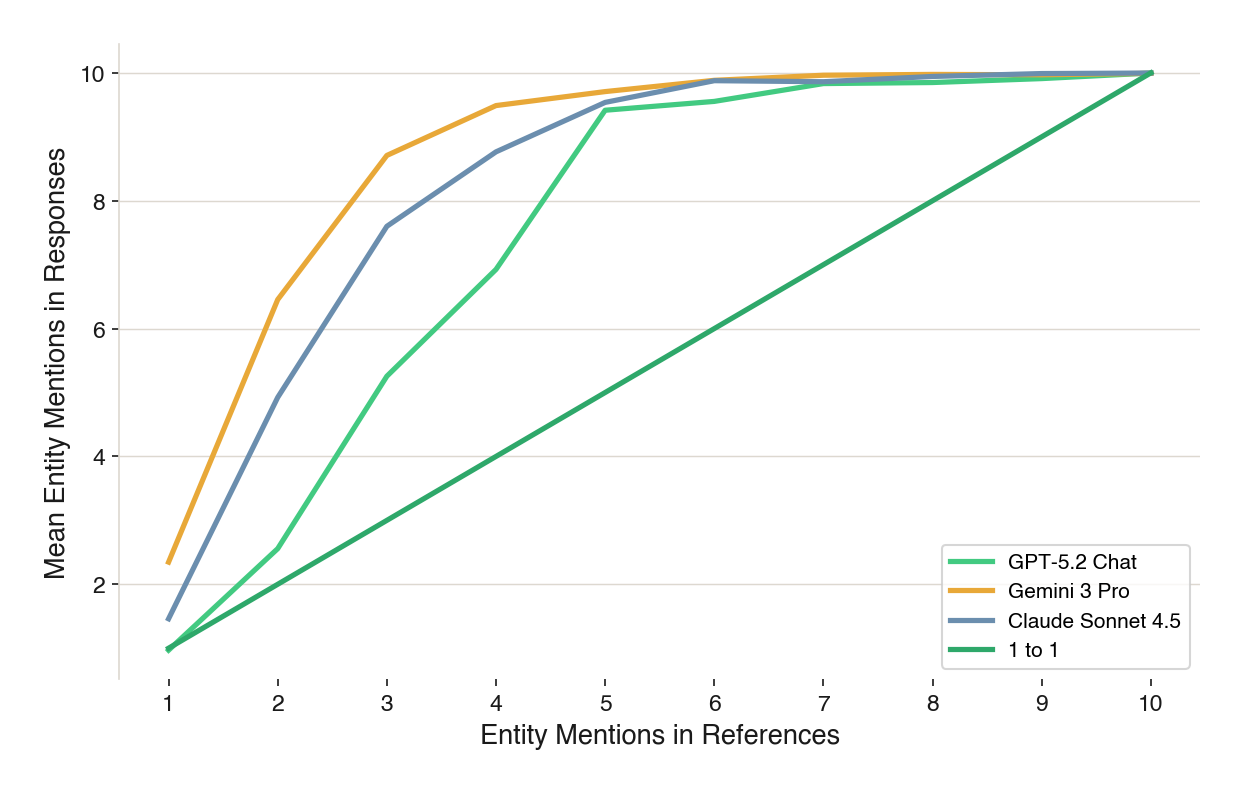}
\caption{Mean entity mentions in responses versus references, showing that response frequencies do not simply mirror reference frequencies.}
\label{fig:image45}
\end{plotblock}

\section{Simulation Design Choices}
The simulations require configuration. We conducted 68 experiments while developing the approach to investigate RAG collapse for this paper, to ensure the results were fair and not dependent on a single simulation detail. Below, we describe several factors that accelerate collapse but that we chose not to include in the final experiments.

In early experiments, GPT-5.1 returned very short responses, resulting in short articles and quick collapse. We changed the prompt to elicit longer responses, so the average length of the original and self-authored references is similar.

We generally observed collapse across all prompts we tried, though we did not experiment with prompts explicitly designed to prevent collapse. We plan to study this in future work.

In the Search simulation, larger chunk sizes increase the rate and probability of collapse. The fact that smaller chunks lead to less collapse supports the hypothesis that self-authored references have disproportionate influence because they provide a direct and complete answer. However, the results we report here use the default chunk size.

Initially, we used raw responses as references, which led to more rapid and frequent collapse. However, we were concerned that this was unfair because the self-authored and original references looked very different, so we decided to convert them into articles.

In early experiments, we used smaller models to reduce costs, and in some cases, their responses collapsed more quickly. We chose to use larger frontier models that power commercial AI systems for the experiments in this paper.

\section{Can We Mitigate RAG Collapse?}
In practice, there may be ways to mitigate collapse:

\begin{itemize}
  \item Filter out AI-generated references during retrieval
  \item Diversify search results to filter out near-duplicates
  \item Encourage the model to favor diversity or fall back to its parametric knowledge via the system prompt or fine-tuning
\end{itemize}

We do not know precisely how commercial AI systems work, so these mechanisms may already be in place. However, whether these mitigations will be effective remains an open question.

Diversifying search results would only help if there is diverse, relevant information online. In some cases, a lack of diversity indicates real consensus, which can be challenging to distinguish.

\subsection{AI Detection}
Filtering out AI-generated references could mitigate the risk of collapse, since self-authored articles are AI-generated. Are AI companies already doing it?

As discussed in Section~\ref{sec:datasets}, we find that as of June 2026, 42.7\% of the references for prompts in our dataset are AI-generated. While it does not appear that ChatGPT is filtering out AI-generated references now, would this be an effective strategy generally?

While AI detection algorithms can be accurate at distinguishing purely AI-generated content from human-written content, they are easily fooled by adversarial attacks and unseen generative models~\citep{dugan2024raid}, and their performance on AI-generated, human-edited content is less well studied. Additionally, identifying AI-generated content is likely to become more challenging as models improve.

We may be able to predict self-authorship directly by comparing a reference to an AI's answer, but this is more specific and challenging than standard AI detection. Note also that we find that simply being AI-generated may not drive collapse, as self-authored references have more influence than AI-generated original references.

As we have seen, even a single self-authored reference can trigger collapse, so even a small number of false negatives could be problematic.

\subsection{Knowledge Collapse}
The self-authored references in our simulations are also AI-generated: the text was written by AI, and the substance is its own answer.

There is also potential for a different type of collapse if many writers use AI as a thought partner while writing, and the substance of their article comes from AI, even if they write or edit the final text. For example, a writer covering ``the best movie of all time'' may use ChatGPT to compile a list of movies, even if they do not fully AI-generate the article. The article is not self-authored by our definition, yet the substance is still the model's answer. This is a form of what others have called knowledge collapse~\citep{peterson2024knowledge}.

AI detection would not help here. A human-written article based on the model's answer may drive collapse but would not be flagged.

A similar dynamic occurs without AI when humans read the same sources or rely on Google's top results while researching an article. But the cost of creating that content is much lower with AI, making it easier to flood the internet with derivatives.

\section{Limitations}
Our simulations seed each question with original references retrieved from commercial AI systems in January 2026. We find that 38.9\% are already AI-generated, and we cannot directly identify which are self-authored, the subset that seems to drive collapse. Any self-authored content already in the seed pool would only reduce the contrast we measure, so our collapse and disproportionate-influence results are conservative.

We do not test iterative addition of other types of AI-generated content, and doing so could also lead to collapse.

Our simulations suggest that retrieving self-authored articles leads to collapse, but we do not definitively prove that RAG collapse is already happening. As discussed, there may already be mechanisms in place intended to mitigate collapse, though it is unclear whether they will be successful.

We focus on information-seeking prompts, especially those involving comparisons of named entities. It is possible that collapse behaves differently on other types of prompts.

\section{Conclusion and Future Work}
AI-generated content is already prevalent online, and AI systems are retrieving it. Our simulations show that responses collapse when those systems retrieve self-authored articles. This result is consistent across models, question types, and simulation designs we explore. Surprisingly, even a single self-authored reference can trigger collapse, because such references disproportionately influence the response. Whether current or future mitigations can prevent RAG collapse in practice remains an open question.

Future work, some of which is already in progress, includes studying collapse under agentic search, the effectiveness of mitigations, experimenting with one model retrieving another's authored references, and how collapse may affect other types of prompts.

\bibliographystyle{plainnat}
\bibliography{references}

\clearpage
\appendix
\section{Complete Replace One Results}
\label{sec:appendix}
In Section~\ref{sec:comparison-models}, we presented experiments comparing models on questions common to both the ChatGPT and AI Overview datasets. Here we provide complete results on those datasets, including questions that are not in the intersection.

\subsection{Replace One, GPT-5.2 Chat}
\begin{plotblock}
\plotpanel{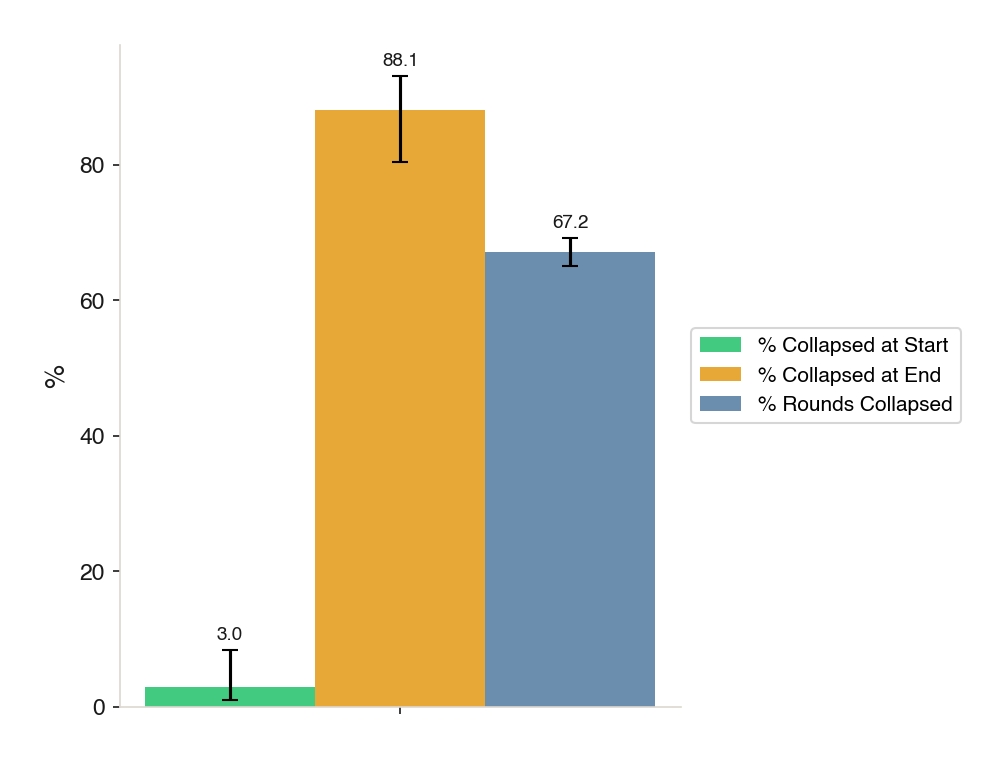}{Entity collapse rates.}{fig:image47}
\hfill
\plotpanel{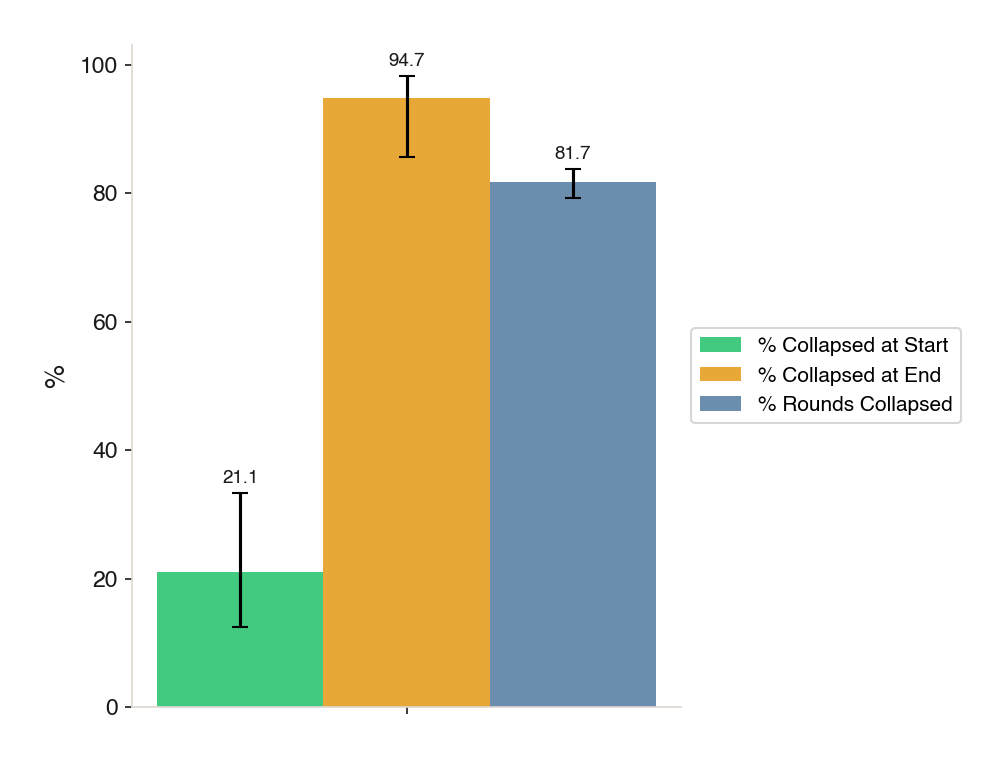}{Editorial collapse rates.}{fig:image6}

\medskip
\plotpanel{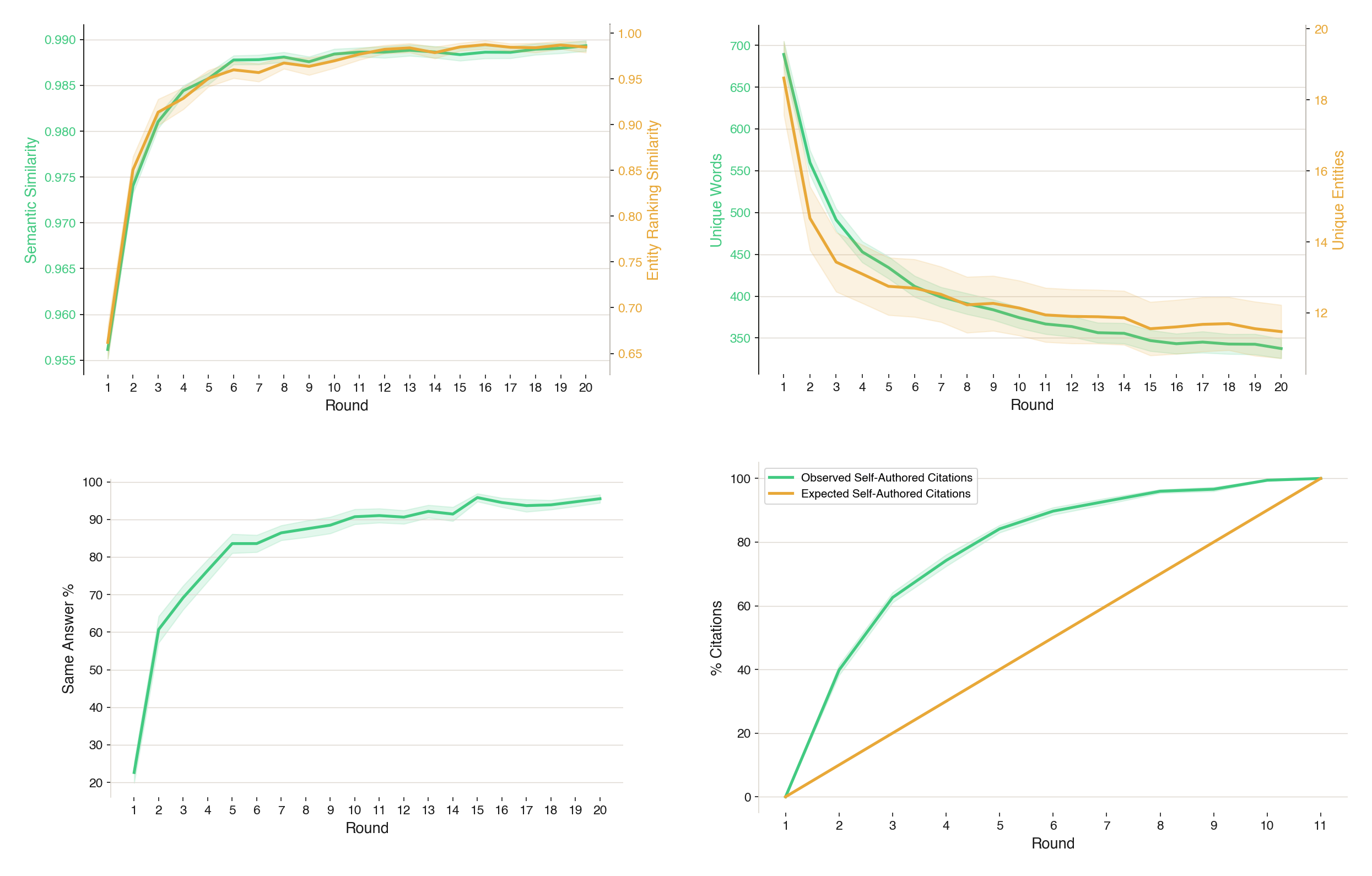}{Entity per-round metrics.}{fig:image53}
\hfill
\plotpanel{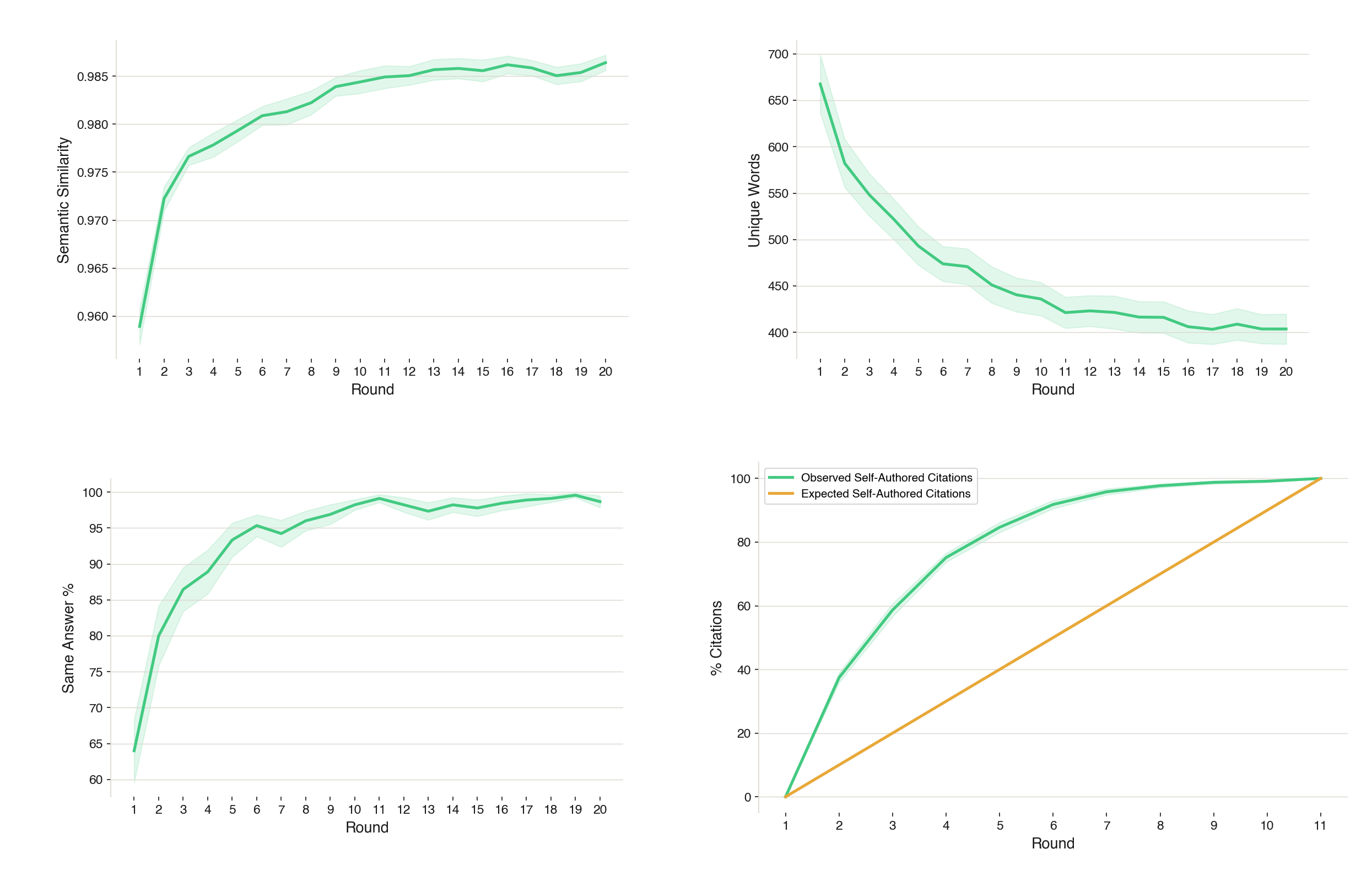}{Editorial per-round metrics.}{fig:image1}
\caption{Complete Replace One results for GPT-5.2 Chat.}
\label{fig:appendix-gpt-grid}
\end{plotblock}

\clearpage
\subsection{Replace One, Gemini 3 Pro}
\begin{plotblock}
\plotpanel{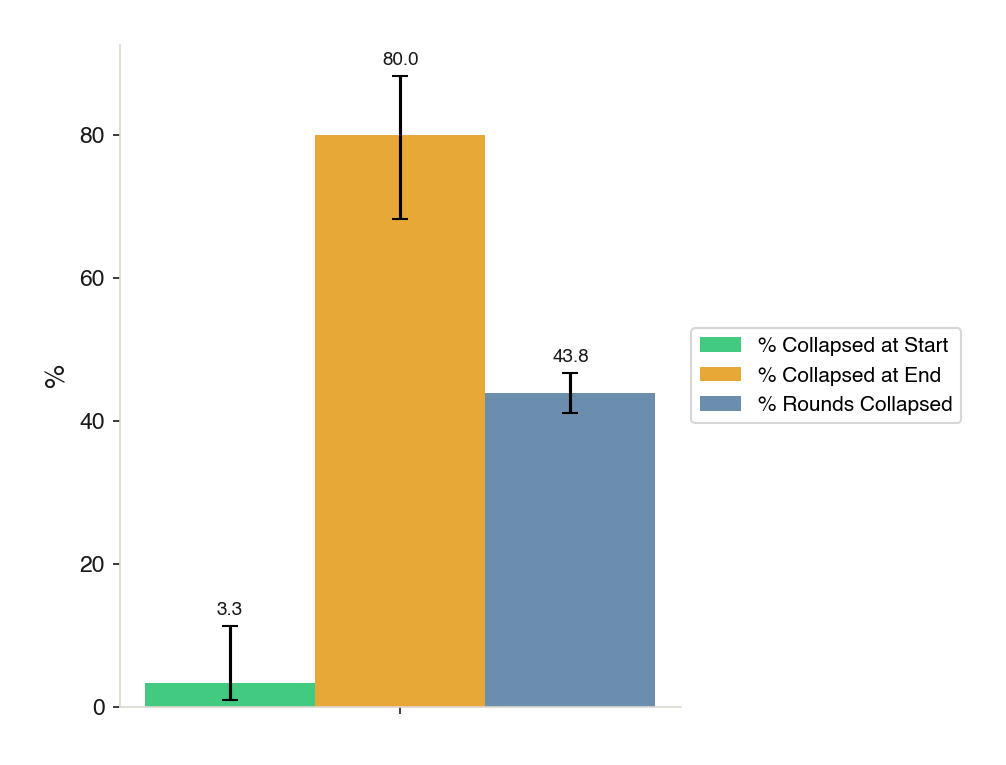}{Entity collapse rates.}{fig:image51}
\hfill
\plotpanel{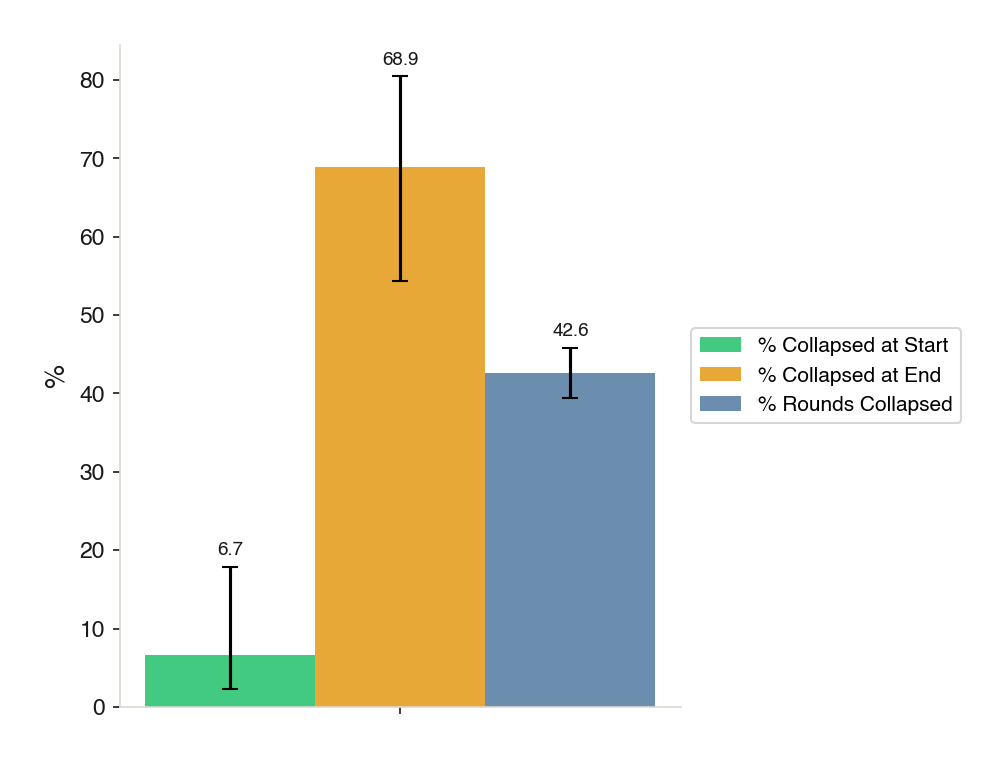}{Editorial collapse rates.}{fig:image43}

\medskip
\plotpanel{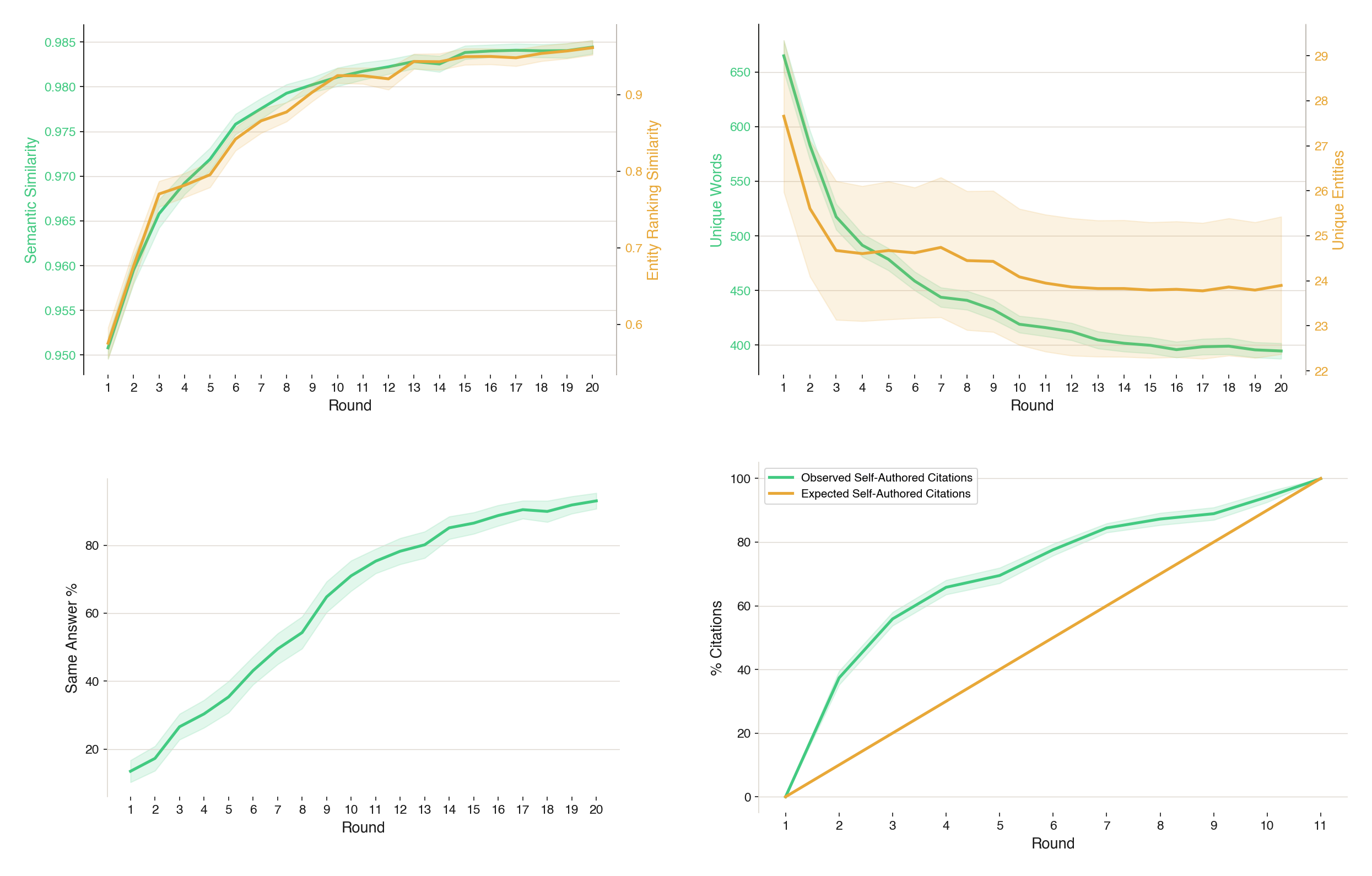}{Entity per-round metrics.}{fig:image24}
\hfill
\plotpanel{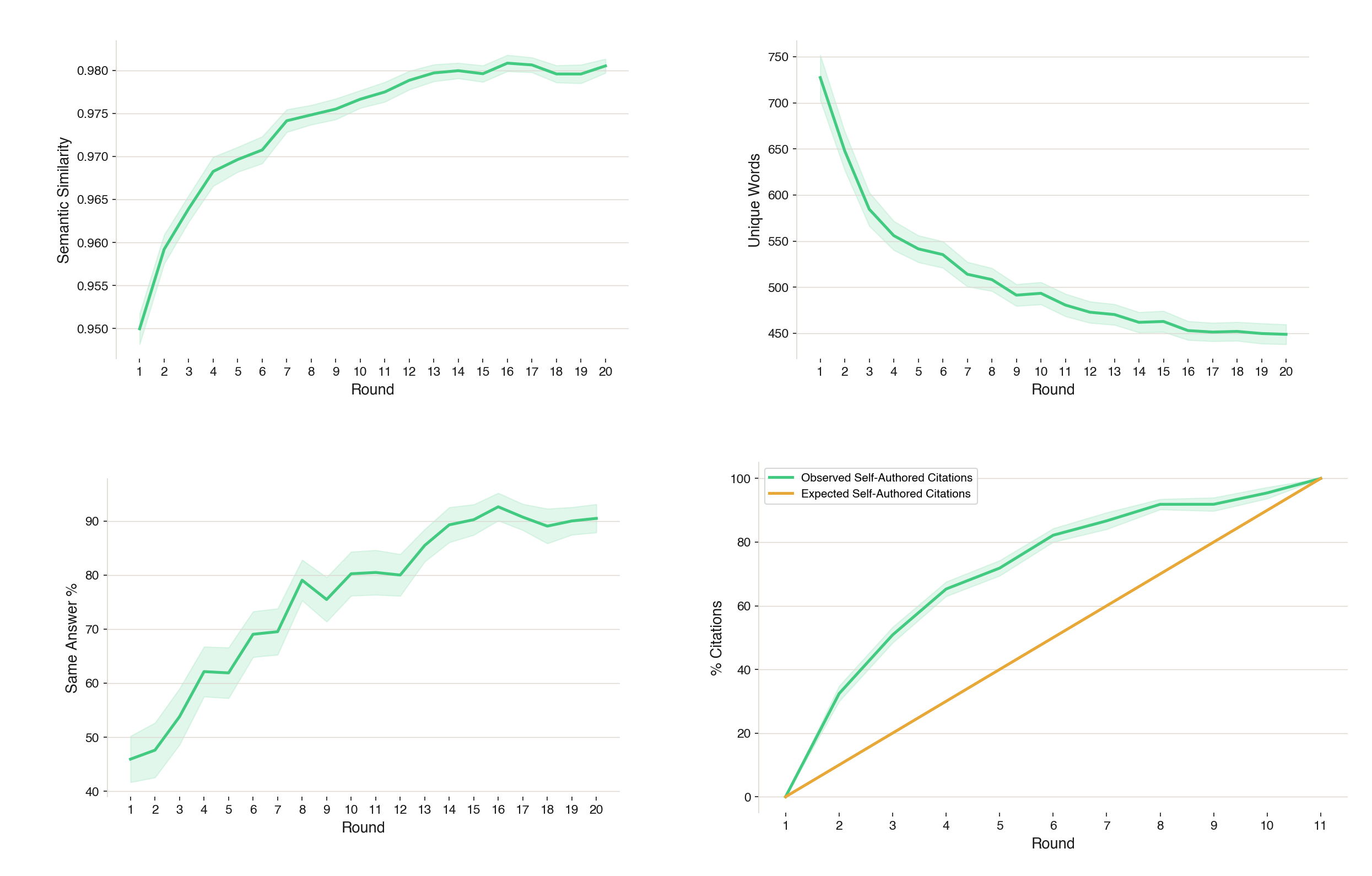}{Editorial per-round metrics.}{fig:image31}
\caption{Complete Replace One results for Gemini 3 Pro.}
\label{fig:appendix-gemini-grid}
\end{plotblock}

\clearpage
\subsection{Replace One, Claude Sonnet 4.5}
\begin{plotblock}
\plotpanel{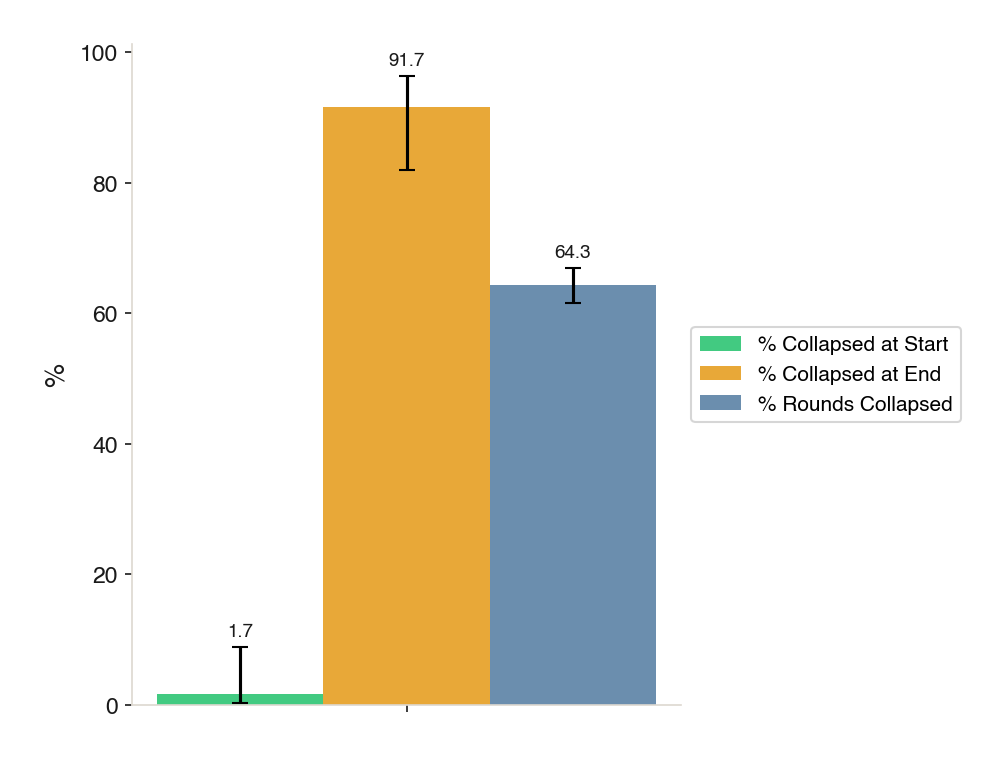}{Entity collapse rates.}{fig:image68}
\hfill
\plotpanel{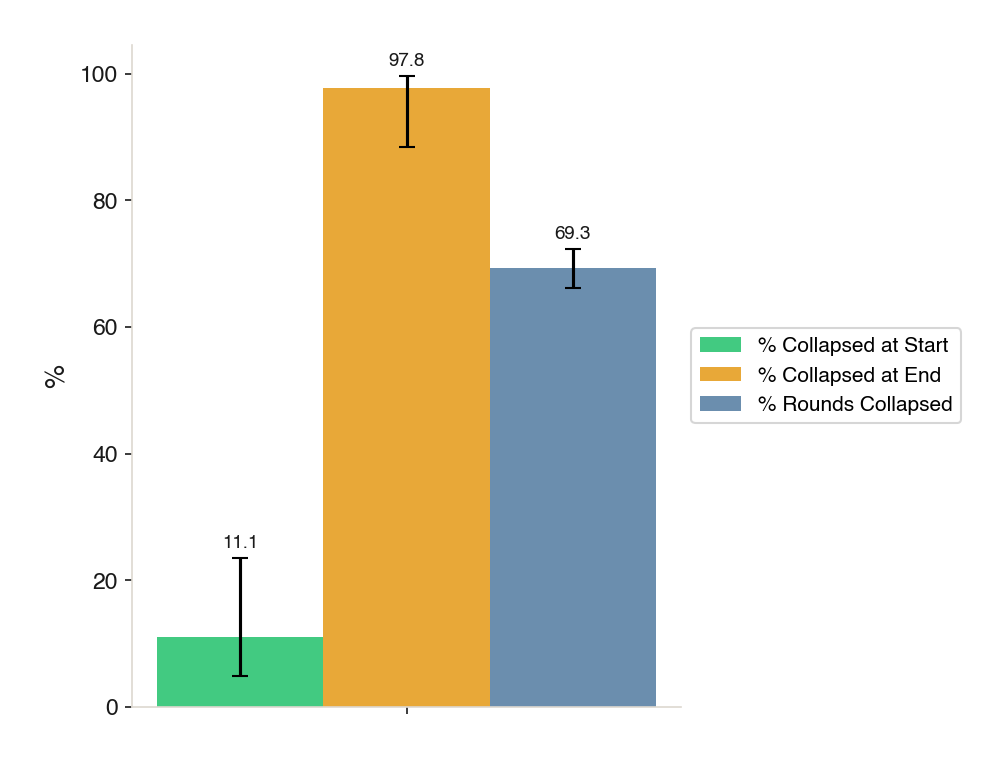}{Editorial collapse rates.}{fig:image64}

\medskip
\plotpanel{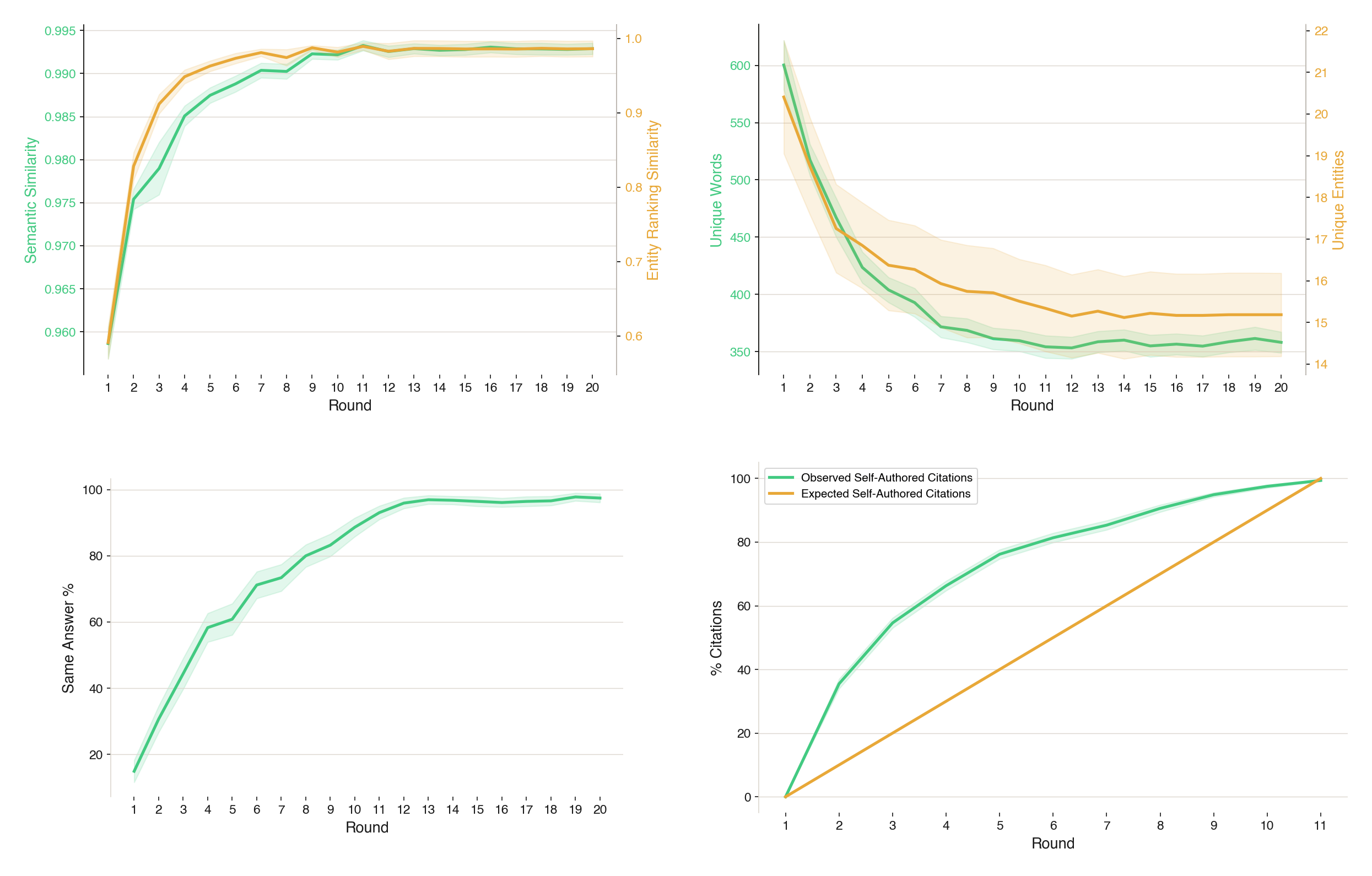}{Entity per-round metrics.}{fig:image65}
\hfill
\plotpanel{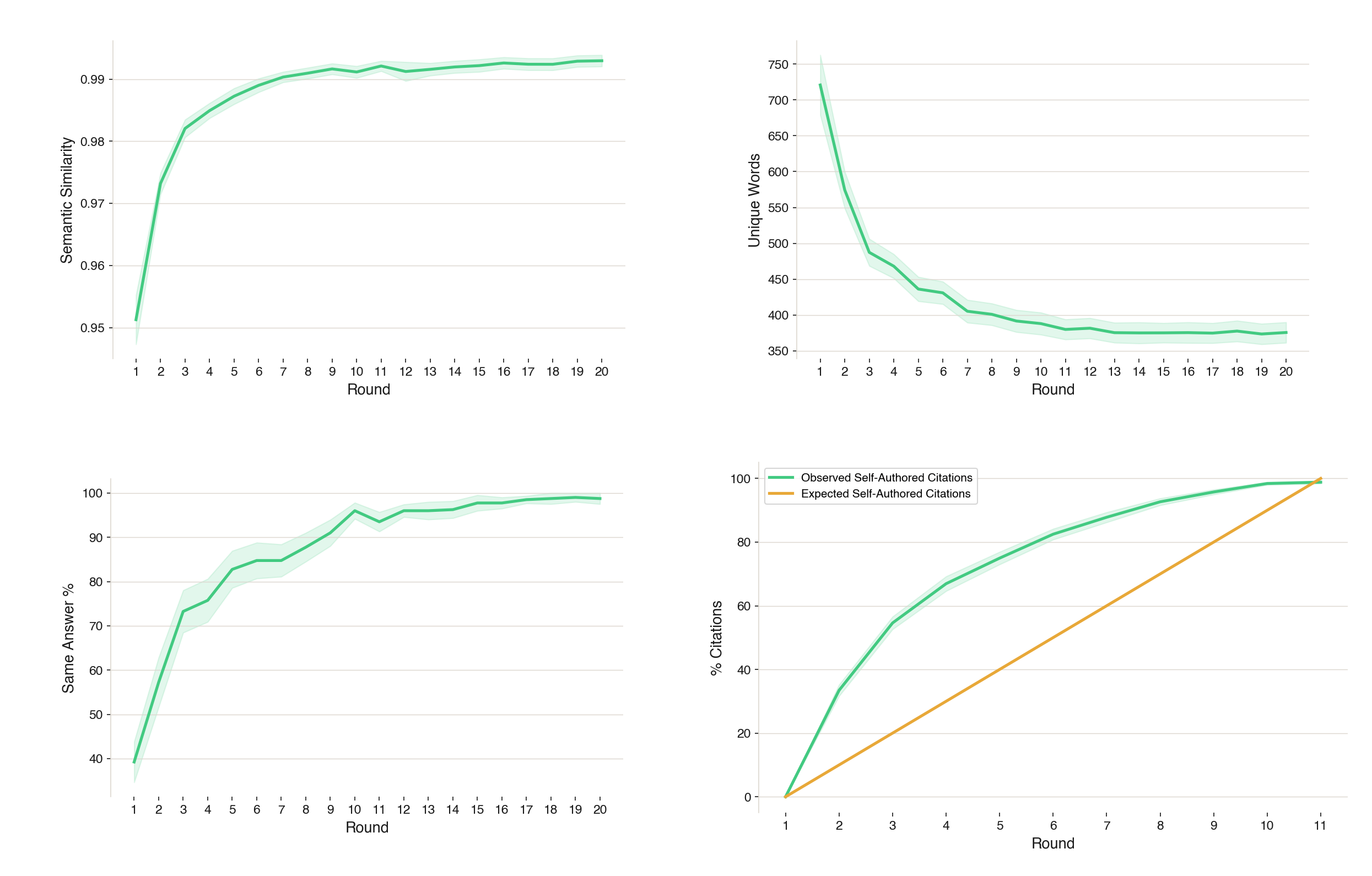}{Editorial per-round metrics.}{fig:image4}
\caption{Complete Replace One results for Claude Sonnet 4.5.}
\label{fig:appendix-claude-grid}
\end{plotblock}
\FloatBarrier

\end{document}